\documentclass{article}

\PassOptionsToPackage{numbers, compress}{natbib}

\usepackage[final]{neurips_2025}

\usepackage[utf8]{inputenc} 
\usepackage[T1]{fontenc}    
\usepackage{hyperref}       
\usepackage{url}            
\usepackage{booktabs}       
\usepackage{amsfonts}       
\usepackage{nicefrac}       
\usepackage{microtype}      
\usepackage{xcolor}         

\usepackage{algorithm}
\usepackage{algorithmic}

\usepackage{comment}   
\usepackage{amsmath, pifont}    
\usepackage{amssymb}
\usepackage{dsfont}    
\usepackage{graphicx}  
\usepackage{colortbl}  
\usepackage{multirow}  
\usepackage{multicol}  
\usepackage{wrapfig}  
\usepackage{rotating}   
\usepackage{tablefootnote} 
\usepackage{caption}     
\usepackage{subcaption}  
\usepackage{titletoc}  
\usepackage[title]{appendix}
\setcitestyle{square}  

\newcommand{\X}{\textbf{X}}  
\newcommand{\D}{\textbf{D}}  
\newcommand{\B}{\textbf{B}}  
\newcommand{\A}{\textbf{A}}  
\newcommand{\C}{\textbf{C}}  
\newcommand{\I}{\textbf{I}}  
\newcommand{\cmark}{\ding{51}} 
\newcommand{\xmark}{\ding{55}} 
\newcommand{\NA}{--}

\def\1{\mathbbm{1}}
\newcommand{\RR}{\mathbb{R}}
\newcommand{\ZZ}{\mathbb{Z}}

\newcommand{\NN}{\mathbb{N}}

\newcommand{\gray}[1]{\textcolor{gray}{#1}}

\definecolor{colorw}{RGB}{0,102,102}

\definecolor{colorh}{RGB}{102,0,102}
\newcommand{\colorh}[1]{\textcolor{colorh}{#1}}
\definecolor{colory}{RGB}{255,0,255}

\definecolor{colorx}{RGB}{0,0,128}

\definecolor{colorhres}{RGB}{0,64,64}

\definecolor{shapecolor}{RGB}{102,0,102}

\definecolor{colorm}{RGB}{110,84,121}
\definecolor{colorf}{RGB}{80,22,74}
\newcommand{\colorm}[1]{\textcolor{colorm}{#1}}
\newcommand{\colorf}[1]{\textcolor{colorf}{#1}}
\arrayrulecolor{colorf} 

\newcommand*{\bestc}{\cellcolor{violet!5}}

\newcommand{\XERA}{\textbf{X}_{ERA\textit{5}}}  
\newcommand{\XCPC}{\textbf{X}_{CPC}}  
\newcommand{\XGloFAS}{\textbf{X}_{GloFAS}}  
\newcommand{\XHRES}{\textbf{X}_{HRES}}  
\newcommand{\XSTATIC}{\textbf{X}_{static}}  

\newcommand{\draft}{false}

\def\mambalogo{\makebox[44pt][c]{\raisebox{-3.0ex}{\includegraphics[draft=false, height=56pt]{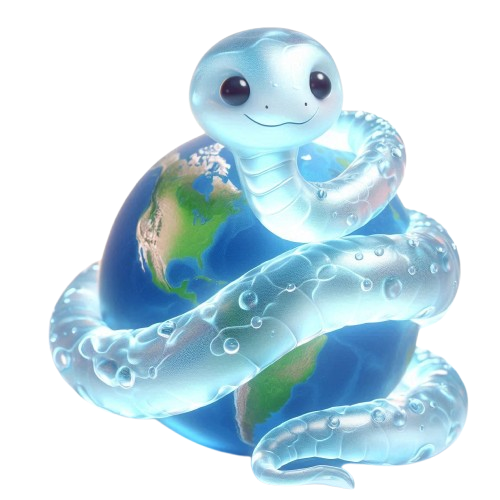}}}\hspace{20pt}}

\title{
RiverMamba: A State Space Model for Global River Discharge and Flood Forecasting}

\author{
  Mohamad Hakam Shams Eddin \qquad Juergen Gall\\\\
  Institute of Computer Science, University of Bonn\\
  Lamarr Institute for Machine Learning and Artificial Intelligence \\
  \texttt{\{shams, gall\}@iai.uni-bonn.de} \\
  \And
  Yikui Zhang \qquad Stefan Kollet\\\\
  Institute of Bio- and Geosciences Agrosphere (IBG-3), Research Centre J\"ulich\\
  Centre for High-Performance Scientific Computing in Terrestrial Systems, \\
  Geoverbund ABC/J, J\"ulich \\
  \texttt{\{yik.zhang, s.kollet\}@fz-juelich.de} \\
}

\begin{document}

\maketitle

\begin{abstract}
Recent deep learning approaches for river discharge forecasting have improved the accuracy and efficiency in flood forecasting, enabling more reliable early warning systems for risk management.
Nevertheless, existing deep learning approaches in hydrology remain largely confined to local-scale applications and do not leverage the inherent spatial connections of bodies of water. 
Thus, there is a strong need for new deep learning methodologies that are capable of modeling spatio-temporal relations to improve river discharge and flood forecasting for scientific and operational applications.
To address this, we present RiverMamba, a novel deep learning model that is pretrained with long-term reanalysis data and that can forecast global river discharge and floods on a $0.05^\circ$ grid up to 7 days lead time, which is of high relevance in early warning. To achieve this, RiverMamba leverages efficient Mamba blocks that enable the model to capture spatio-temporal relations in very large river networks and enhance its forecast capability for longer lead times. The forecast blocks integrate ECMWF HRES meteorological forecasts, while accounting for their inaccuracies through spatio-temporal modeling. Our analysis demonstrates that RiverMamba provides reliable predictions of river discharge across various flood return periods, including extreme floods, and lead times, surpassing both AI- and physics-based models.
The source code and datasets are publicly available at the project page \colorh{\url{https://hakamshams.github.io/RiverMamba}}.

\end{abstract}

\section{Introduction}
\label{sec:intro}

Riverine floods are one of the most destructive natural disasters, with their risk anticipated to rise in the future as a result of climate change and socioeconomic developments \cite{dottori2018increased, merz2021causes, kreibich2022challenge, rentschler2022flood, Myanna_2022}. 
They arise from compound effects, including atmospheric conditions like heavy precipitation caused by circulation patterns and snowmelt succeeding high temperature, all shaped by the specific characteristics of the river drainage area \cite{Shijie_2022}. The interaction of these elements influences flood timing, scale, and severity \cite{Shijie_2022}. This complexity makes future flood risk assessment challenging, as a changing climate may alter these drivers in unpredictable ways \cite{Shijie_2024}. 
Therefore, early prediction of flood risk, especially for extreme floods, is a key measure for effective flood risk mitigation \cite{PAPPENBERGER2015278, camps2025artificial}. 

To support national forecasting initiatives, current operational flood early warning systems can forecast river discharge in real-time and provide flood forecasts at different scales \cite{Emerton_2016, HyFS, EFAS}.
The discharge forecasts derived from these systems can be further processed using inundation models to create anticipated flooded areas \cite{Nevo_2022, najafi2024high}.
The Global Flood Awareness System (GloFAS) \cite{GloFAS, GloFAS_Forecast}, developed under the Copernicus Emergency Management Service (CEMS) and operated by the European Centre for Medium-Range Weather Forecasts (ECMWF), represents the cutting-edge physics-based model for real-time and worldwide hydrological forecasting. 
However, physics-based hydrological models are expensive to run and require extensive calibration to handle complex catchment characteristics. 

\begin{figure}[t]
  \centering
  \includegraphics[draft=\draft, width=.85\textwidth]{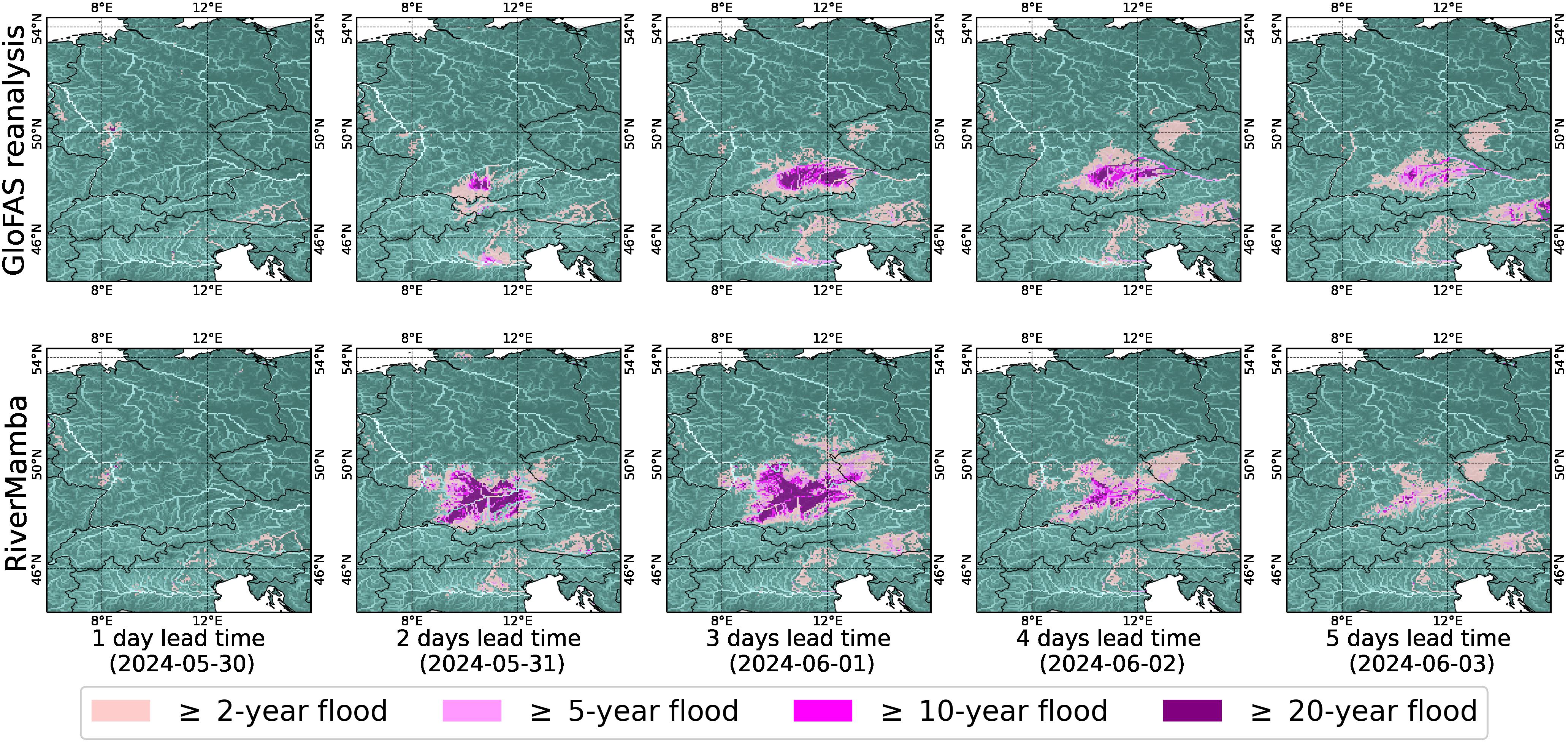}
  \caption{Example of a 5-day forecast of river discharge and flood events. In early June 2024, a significant flood affected Southern Germany. While the top row shows the floods obtained from the GloFAS reanalysis, the bottom row shows the river discharge forecast by our approach. The severity of floods is categorized by the statistical flood return period, i.e., occurring every 10 years.}\label{fig:teaser}
  \vspace{-4mm}
\end{figure}

AI-based early warning systems are thus considered as vital tools to enhance climate risk resilience \cite{jones2023ai, reichstein2025early} and to enable flood forecasting without requiring full physical process understanding \cite{Nearing_2021, Frame_2022}.
While deep-learning approaches for weather forecasting~\cite{pathak2022fourcastnet,GenCast,PanguWeather} have been investigated in recent years, very little work has been done for forecasting river discharge at large spatial regions since it is very challenging. It requires the combination of sparse gauged river observations with high-resolution land surface, re-analysis, and weather forecast data. 
Furthermore, floods occur rarely and the goal is to forecast floods of different severity as shown in Fig.~\ref{fig:teaser}. Recently, an LSTM-based model has been proposed \cite{Google_nature}. 
While it achieves promising results, it forecasts floods only locally at sparse river basins and does not consider routing. 
Modeling spatio-temporal relations, however, is very important and required to generate consistent dense maps as in Fig.~\ref{fig:teaser}, since river discharge at points near connected bodies of water is highly correlated. 
 

In this work, we propose the first deep learning approach for global river discharge and flood forecasting that is not only capable of forecasting at sparse gauged observation points, but also of forecasting accurate, high-resolution ($0.05^\circ$) global river discharge maps. In order to deal with the sparseness of gauged river points and the computational complexity of modeling spatio-temporal relations at the global scale, our proposed RiverMamba leverages Mamba blocks, which are bidirectional state space models \cite{Linear_SSM,S4,S5,mamba}, and spatio-temporal forecast blocks. 
Using a specialized procedure to convert sampled points into 1D sequences, RiverMamba maintains a very large spatio-temporal receptive field, connecting the routing of the river channel networks and the teleconnection of meteorological data across space and time. RiverMamba has thus the possibility to consider a spatio-temporal context that covers very large river networks like the Amazon River.  
The forecast layers are further forced by high-resolution meteorological data (HRES) to generate medium-range river discharge forecasts up to 7 days lead time.
To address uncertainty in the meteorological forcing, we built the forecast layers so that they can, for each catchment point, incorporate information about meteorological forcing from the neighboring points and throughout the temporal dimension. Thus, RiverMamba ensures a consistent forecast through space and time.
Our contributions can be summarized as follows:
\begin{itemize}
    \item We introduce a novel Mamba-based approach, called RiverMamba, for global river discharge and flood forecasting. It is the first deep learning approach that is capable of providing maps of global river discharge forecasting at $0.05^\circ$, and it introduces a novel methodology to hydrology. It is able to integrate sparse gauged observations, river attributes, high-resolution reanalysis data, and weather forecast. The efficient structure allows to model spatio-temporal relations covering entire river networks.
    
    \item We evaluate RiverMamba on both long-term real-world reanalysis and observational data where it outperforms state-of-the-art AI- and physics-based operational systems for global flood forecasting. 
\end{itemize}

\section{Related works}
\label{sec:related_works}

\textbf{Flood forecasting.} Floods can be categorized into three common types. The first type is the \textit{fluvial} or \textit{riverine flood} \cite{Bomers_2023}. It occurs when the water level in a stream rises and overflows onto the adjacent land. The second type is the \textit{coastal flood}, also known as storm surges \cite{nhess-25-747-2025}. The third type is the \textit{pluvial flood}, often referred to as \textit{flash flood} \cite{hess-28-5443-2024, w13162255, busker2025value} that can occur with extreme rainfalls. 
Machine learning (ML) has become an essential element for the development of hydrological simulation and flood models \cite{mosavi2018flood, Bentivoglio_2022}.
Each type of flood has unique drivers and impacts. 
Consequently, ML methods require different strategies to forecast them.
Related tasks to flood forecasting are urban flood modeling \cite{Gau_2021, 10351020, Seleem31122022, w15091760}, flood inundation \cite{CHU2020104587, w15030566, Floodsformer_2025}, and flood extension and susceptibility mapping \cite{10758300, JANIZADEH2021113551, Dasgupta_2021, NEURIPS2024_43612b06}. In this work, we are interested in forecasting riverine floods (fluvial) based on river discharge. 

\textbf{River discharge forecasting.} 
River discharge can be used to detect fluvial flood signals when the magnitude of the flow exceeds certain thresholds. 
Current deep learning methods for forecasting river discharge are primarily based on locally lumped models \cite{palash2024data, ZHONG2024132165}, hypothesizing that a single model can generalize across many catchments without considering the spatial-temporal information over grids \cite{hess-28-4187-2024}. 
The dominating backbone is the LSTM model \cite{LSTM} which is used in most recent studies such as EA-LSTM \cite{EA-LSTM, Heudorder_2025}, ED-LSTM \cite{ZHANG2022128577, ZHANG2024100617}, Hydra-LSTM \cite{ruparell2024hydra}, MC-LSTM \cite{MC-LSTM}, MF-LSTM \cite{MF-LSTM}, and DRUM \cite{DRUM_2025}.
These models learn features specific to individual rivers or entities and lack spatial and topological information. However, river networks have spatio-temporal causal relations \cite{stein2025causalrivers}. 
Only a few studies deviate from this conventional modeling and propose to model the network topology with Graph Neural Networks \cite{kirschstein2024merit, roudbari2024data, FloodGNN}. They are still limited to small scales and  the graph models fail in most cases to capture topological information \cite{kirschstein2024merit}. 
Others applied an LSTM model on a coarse grid to estimate runoff and then coupled it with a river routing model to produce daily discharge at coarse resolution \cite{yang2025global}. 
In \cite{Vischer_2025}, LSTM resolves local runoff spatially on a regular grid in central Europe. Then, routing the runoff along the entire river networks is implemented as 1D-convolutions and fully connected layers.
The impact of defining routing explicitly with physics-informed neural networks has also showed an advantage in recent studies, especially, in improving streamflow in large continental river networks compared to models that do not consider routing \cite{Bindas_2024, Song_2025}. In a hybrid modeling framework, physical equations including river routing are parametrized using 3D-convolutions and fully connected layers for distributed hydrological modeling \cite{Wang_2024}.
The most relevant work is the Encoder-Decoder LSTM \cite{EA-LSTM} developed for the Google global operational forecasting system \cite{Google_nature}, which is a locally lumped model. 
In this work, distinct from previous works, we propose an approach that is capable of modeling a large spatio-temporal context and forecasting medium-range river discharge at grid-scale.

\textbf{State space model (SSMs) and the Mamba family.}
Linear SSMs \cite{Linear_SSM} and structured SSMs like S4 \cite{S4} and S5 \cite{S5} were primarily introduced for long-sequence modeling in NLP. 
Recently, Mamba \cite{mamba} introduced the selective scan mechanism, enabling efficient training and linear-time inference.
Built upon Mamba, VMamba \cite{vmamba} and Vim \cite{vimmamba} in the vision domain were introduced as appealing alternatives to the quadratic complexity of vision transformers \cite{ViT} while improving scaling efficiency on long token sequences. A series of works have adapted Mamba to tasks like image generation \cite{DiMSUM, zigmamba}, image classification \cite{NEURIPS2024_89b89c04, GroupMamba}, video understanding \cite{videomamba, chen2024video}, motion generation \cite{motionmamba}, dense action anticipation \cite{zatsarynna2025manta}, and point cloud processing \cite{pointmamba, NEURIPS2024_947b6383}. 
In this work, we propose a Mamba-based approach for global river discharge and flood forecasting.

\section{RiverMamba}
\label{sec:method}

\begin{figure}[!t]
  \centering
  \includegraphics[draft=\draft, width=.99\textwidth]{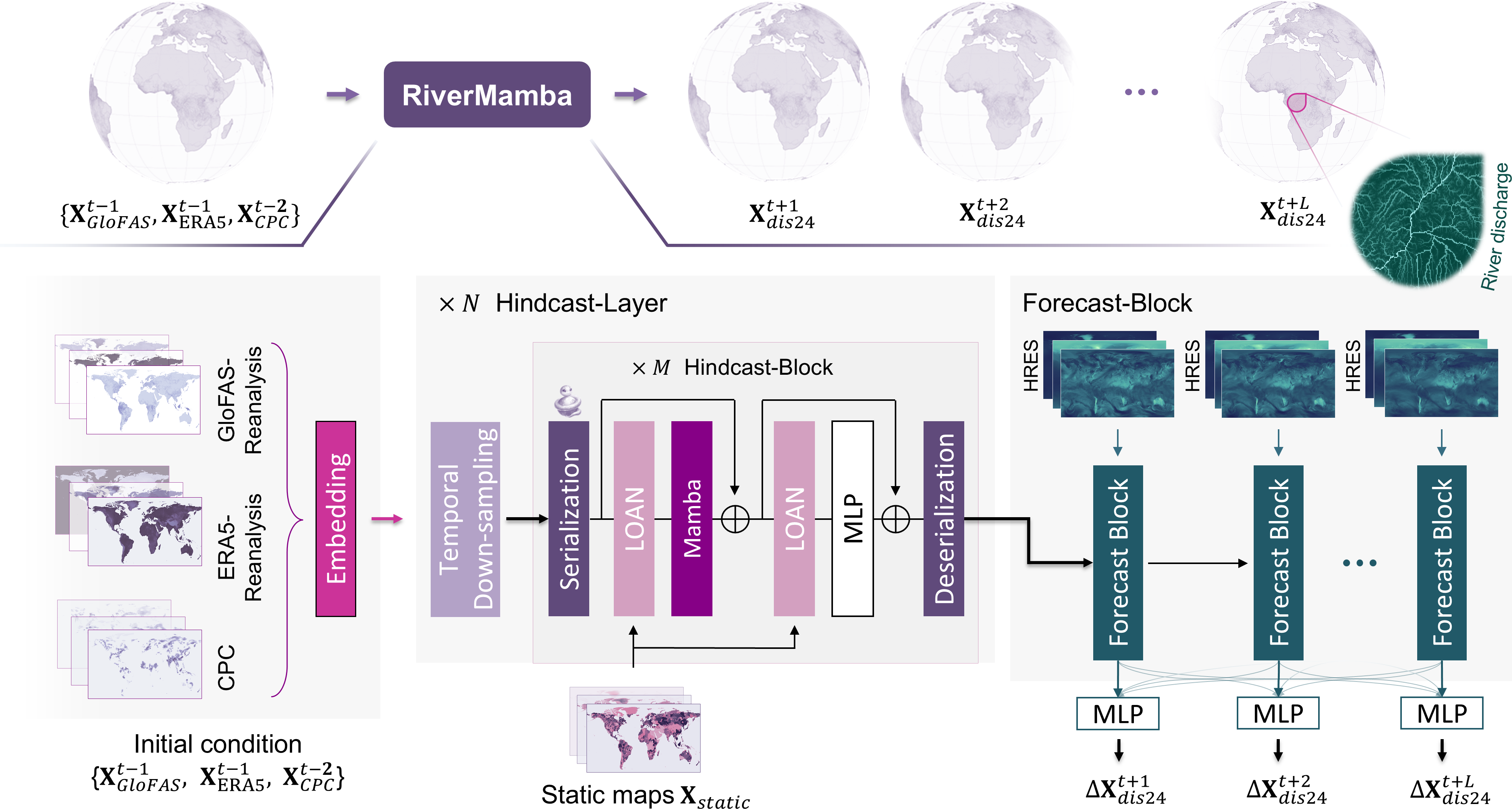}
  \caption{An overview of the proposed RiverMamba model for river discharge forecasting. The model forecasts at time $t$, high-resolution river discharge maps $\X_{dis24}^{t+1:t+L}$ from initial conditions ($\XERA^{t-T:t-1}$, $\XGloFAS^{t-T:t-1}$, $\XCPC^{t-T-1:t-2}$), static river attributes ($\XSTATIC$), and meteorological forecasts ($\XHRES^{t+1:t+L}$). 
  }\label{fig:Model}
  \vspace{-3mm}
\end{figure}

In this work, we present the first deep learning approach that not only forecasts flood events at sparse gauged river observations, but that is capable of forecasting accurate, high-resolution (i.e., at $0.05^\circ$) maps of river discharge up to few days at global scale, as shown in Figs.~\ref{fig:teaser} and \ref{fig:Model}. These maps are essential to forecast flood events of various severity like a flood that re-occurs statistically within a 1.5-year return period or a `flood of the century'. This is very challenging since it requires a model that models spatial-temporal relations in an efficient way and integrates different sources of data (Fig.~\ref{fig:Model}).

As input, we use the initial condition of the forecasts from ERA5-Land reanalysis \cite{ERA5-Land}, denoted by $\XERA^{t-T:t-1} = \{\XERA^{t-T}, \ldots, \XERA^{t-2}, \XERA^{t-1}\}$, the initial condition $\XGloFAS^{t-T:t-1}$ from the GloFAS reanaylsis data \cite{GloFAS_Reanalysis}, and the initial condition $\XCPC^{t-T-1:t-2}$ from the operational global unified gauge-based analysis of daily precipitation \cite{xie2010cpc, xie2007gauge, chen2008assessing}. We also include data from weather forecasts, where we use the high-resolution meteorological forcing forecasts $\XHRES^{t+1:t+L}$ from the ECMWF Integrated Forecast System (IFS), where $L$ is the lead time for the forecast. We generate the river discharge forecast at $t$, using 00:00 UTC as reference time, for $t{+}1$ until $t{+}L$. This means that we do not address nowcasting but only forecasting as it is more relevant. 
We also do not include any nowcasts ($\XHRES^{t}$) as input.
The rationale behind this is to ensure broader applicability, since many weather forecast systems especially ML models provide forecasts at $t > 0$.
However, adding nowcasts to the model is straightforward if they are available.
To make the setup as realistic as possible, we do not include any data after 00:00 UTC and we consider $\XGloFAS$ and $\XERA$ at day $t-1$ and $\XCPC$ at day $t-2$. 
Additionally, we include river attributes $\XSTATIC$ like catchment morphology from LISFLOOD \cite{LISFLOOD_Static}. The input variables are described in details in the suppl.\ material. 
Given these inputs, RiverMamba forecasts changes of the daily mean river discharge $\Delta \X_{dis24}^{t+1:t+L}$ relative to the daily mean river discharge at $t{-1}$, i.e., $\X_{dis24}^{t-1}$. The forecast daily mean river discharge  is thus given by $\X_{dis24}^{t+l} = \X_{dis24}^{t-1} + \Delta \X_{dis24}^{t+l}$.      

An overview of RiverMamba is shown in Fig.~\ref{fig:Model}. For training, we sample $P$ points that are on the land surface and near water bodies. The details are described in the suppl.\ material. For each point $p$, we obtain a temporal sequence of embedding vectors $\X_{embed}^{t-T:t-1}(p)$:
\begin{align}
    \X_{embed}^{t}(p) = \text{LN}\Bigl(\text{Tanh}\Bigl(\text{Concat}\bigl(\text{Linear}(\XERA^t(p)), \text{Linear}(\XGloFAS^t(p)), \text{Linear}(\XCPC^{t-1}(p))\bigr)\Bigr)\Bigr)\,, \label{eq:05}
\end{align}
where LN is the layer norm and Linear is the projection layer. The dimensions of the input are $\XERA \in \RR^{B \times T \times P \times V_{e}}$, $\XGloFAS \in \RR^{B \times T \times P \times V_{g}}$, and $\XCPC \in \RR^{B \times T \times P \times 1}$, where $B$ is the batch size, $V_{e}$ is the number of variables from ERA5, and $V_{g}$ is the number of variables from GloFAS. The embedding $\X_{embed} \in \RR^{B \times T \times P \times K}$, where $K=192$ is the dimensionality of the embedding, is then the input to the encoder defined by the hindcast layers.  

\textbf{Hindcast layer.} The hindcast layers model spatio-temporal relations and aggregate the observations over time. 
Except for the $1^{st}$ layer which processes the full temporal resolution, the temporal resolution is down-sampled by a factor of $2$ with a linear layer at the beginning of each hindcast layer, 
such that the output of the last hindcast layer $\X_{hindcast} \in \RR^{B \times 1 \times P \times K}$ has a temporal resolution of $T=1$.
In our implementation, we chose $T=4$ as for the GloFAS operational system and a temporal down-sampling of $2$. Consequently, we defined $3$ layers to encode the input.

The hindcast layers further integrate the static river attributes $\XSTATIC$ that contain additional information like catchment morphology, which is relevant for flood forecasting. While we analyze the impact of the different inputs, in particular the river attributes, in the suppl.\ material, another key aspect of the hindcast blocks is the specialized serialization of the spatio-temporal points and the Mamba blocks~\cite{mamba, vmamba, vimmamba}. The serialization defines the way the sampled points are connected, and the Mamba block efficiently updates the features of each point based on the spatio-temporal structure. This is a very important design choice since transformer blocks are computationally infeasible for global flood forecasting, whereas \cite{Google_nature} does not consider spatial relations at all. In the suppl.\ material, we also show that an alternative using Flash-Attention \cite{dao2022flashattention, dao2023flashattention2} is inferior in terms of inference time and accuracy compared to our approach.    

The output of the last hindcast layer is then processed along with the HRES meteorological forcing by forecast blocks, and MLP-based regression heads predict for each lead time $l$ the difference of daily mean river discharge $\Delta \X_{dis24}^{t+l}$ with respect to the daily mean river discharge at $t{-}1$. In the following, we describe the components of RiverMamba in details.


\textbf{Hindcast block.} As shown in Fig.~\ref{fig:Model}, the hindcast block has three main components: serialization and deserialization, location-aware adaptive normalization layers (LOAN) to integrate static river attributes, and the Mamba block. 

\begin{wrapfigure}{r}{0.5\textwidth}
\includegraphics[draft=\draft, width=.99\linewidth]{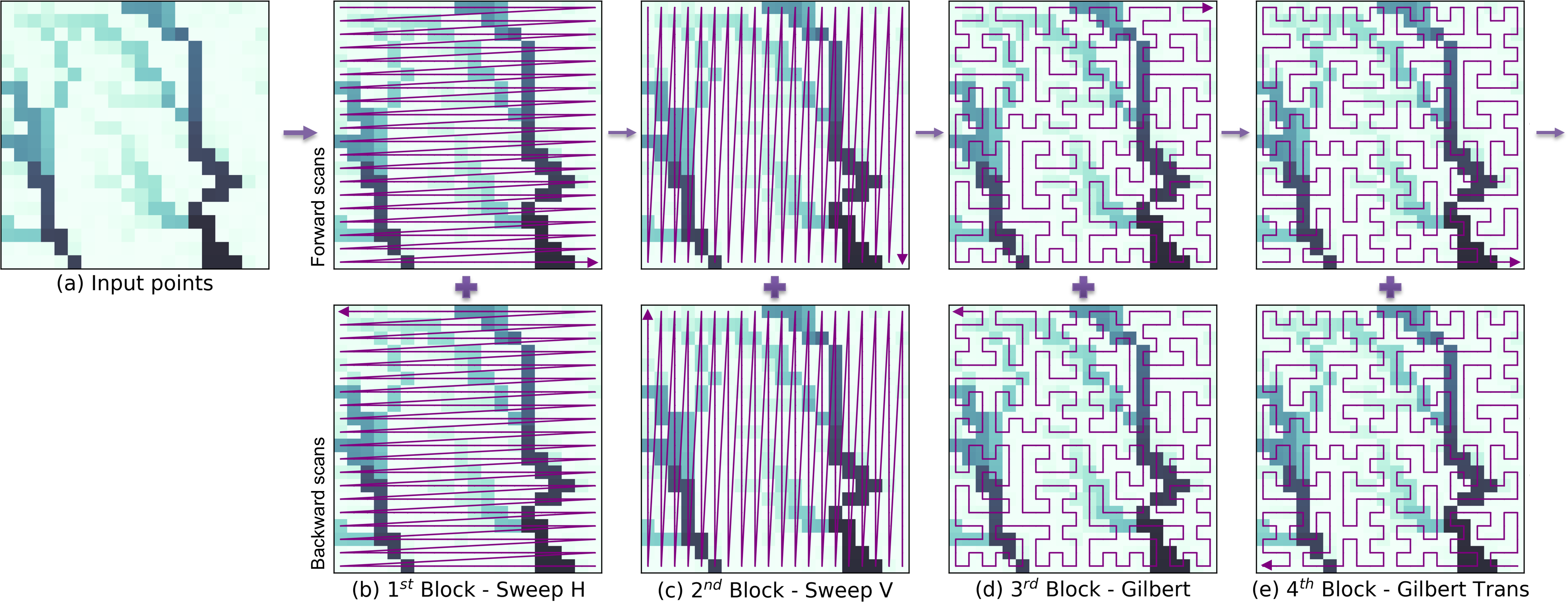}
\caption{Illustration of the spatial scans in RiverMamba. Larger images are in the supp.\ material.}\label{fig:curves_blocks}
\end{wrapfigure}

\textbf{Serialization.}  
The serialization defines the spatio-temporal scanning path over all sampled points for the following Mamba block.
For this, we propose space-filling curves that sequentially traverse through all points.  
The concept was introduced in \cite{peano1990courbe} and the space-filling can be defined as a bijective function $\Phi: \ZZ^3 \rightarrow \NN$, where every point in the discrete space corresponds to a unique index within the sequence. We call this mapping the serialized encoding. The serialized decoding is done as $\Phi^{-1}: \NN \rightarrow \ZZ^3$, where every index is mapped back into its corresponding position. We call this deserialization. We investigated three curves: the Generalized Hilbert (Gilbert) curve, which is a generalized version of the Hilbert curve \cite{hilbert1935stetige}, as well as the Sweep and Zigzag curves in vertical and horizontal directions. Examples of space-filling curves in 2D are illustrated in Fig.~\ref{fig:curves_blocks}. 
As shown in the suppl.\ material, a combination of Sweep and Gilbert curves performs best. To this end, each hindcast block has its own curve. As shown in Fig.~\ref{fig:curves_blocks}, we sweep in the first block over the horizontal direction. The spatial curves are connected over time by continuing the last point of the curve at $t$ with the first point of the curve at $t{+}1$. The second block then sweeps over the vertical direction and we continue with the Gilbert curve and its transposed. These four space-filling curves are iterated. By altering the curves sequentially through the hindcast blocks, the sampled points will be connected and scanned from diverse spatial perspectives, enabling RiverMamba to capture different contextual features.

\textbf{Location-aware adaptive normalization layer.}
In order to condition the model on static river attributes $\XSTATIC$, the location-aware adaptive normalization layer (LOAN) \cite{LOAN} modulates the features $\X$ within the hindcast block: 
\begin{align}
    \text{LOAN}(\X) = \left(\frac{\X - \mu}{\sigma}\right) ~+~ \text{GELU}(\text{Linear}(\XSTATIC)) \,, \label{eq:06}
\end{align}
where a linear layer projects $\XSTATIC \in \RR^{B, 1, P, V_{s}}$, with $V_{s}$ being the number of static variables, to $\RR^{B, 1, P, K}$. The output is then duplicated along the temporal dimension so that the output has the dimension $\RR^{B, T, P, K}$. $\mu \in \RR^{B, T, P, 1}$ and $\sigma \in \RR^{B, T, P, 1}$ are the mean and standard deviation of $\X$ along the channel dimension, respectively, and $\X \in \RR^{B, T, P, K}$ is the input to the LOAN layer. Both $\mu$ and $\sigma$ are duplicated $K$ times along the last dimension to match $\X$.
The layer normalizes the features and adds a systematic bias based on the attributes. For instance, the features are normalized and biased based on location attributes that have an impact on drainage and floods.

\begin{figure}[!t]
  \centering
  \includegraphics[draft=\draft, width=.99\textwidth]{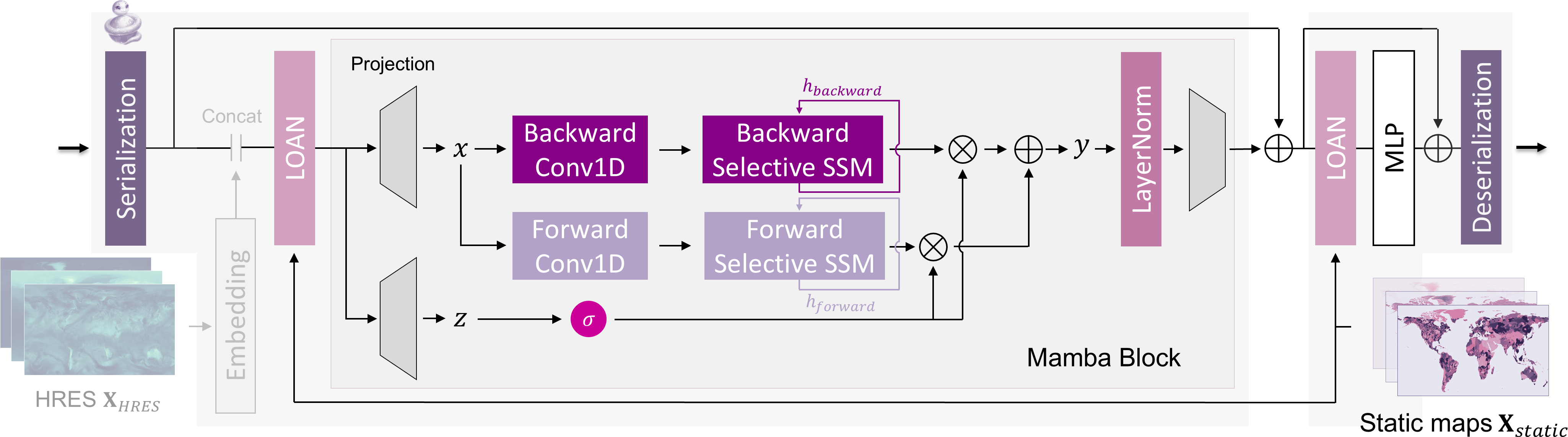}
  \caption{The structure of the hindcast block and forecast block. Both use a bidirectional Mamba block and the forecast block has the same structure as the hindcast block, but it additionally incorporates meteorological forecasts (HRES) by concatenation. The forecast block also includes LOAN layers although it is not shown in Fig.~\ref{fig:Model}}\label{fig:Mamba_block}
  \vspace{-2mm}
\end{figure}

\textbf{Mamba block.} Fig.~\ref{fig:Mamba_block} shows a more detailed structure of the hindcast block with the elements of the Mamba block. After the input is serialized into a 1D sequence based on the block-specific space-filling curve and the features are normalized by the LOAN layer, the Mamba block processes the features of the sampled points along the sequence. 

The Mamba block is based on a state-space model that transforms a 1D sequence of states $x(t)$ into another representation $y(t)$ through an implicit hidden latent state $h(t)$ and a first-order ordinary differential equation:
\begin{align}
    h^{\prime}(t) = \A h(t-1) + \B x(t)\,, ~~~~ y(t) = \C h(t) + \D x(t)\,. \label{eq:01}
\end{align}
To integrate Eq.~\eqref{eq:01} into a deep learning framework, S4 \cite{S4} parametrized the system with the matrices ($\A, \B, \C, \D$) and discretized it with a timescale parameter $\Delta$:
\begin{align}
    h_t = \bar{\A}h_{t-1} + \bar{\B}x_t\,, ~~~~ y_t = \C h_{t} + \D x_t\,, \label{eq:02}\\
    \bar{\A} = e^{(\Delta\textbf{A})}\,, ~~~~ \bar{\B} = (\Delta \A)^{-1} (e^{(\Delta\textbf{A})}-\I)\Delta \B \,, \label{eq:03}
\end{align}
where $\bar{\A}$ and $\bar{\B}$ are the discretized versions of the system. Recently, S6 \cite{mamba} proposed to make Eqs.~\eqref{eq:02} and \eqref{eq:03} time-variant. To this end, the parameters $\B(x)$, $ \C(x)$, and $\Delta(x)$ become dependent on the input state $x$. 
This representation of a state-space model is called Mamba, which is an efficient alternative to transformers \cite{Transformer}, particularly when processing many points as in our case. 

Fig.~\ref{fig:Mamba_block} illustrates the steps of the Mamba block. The normalized sequence $\X \in \RR^{B \times (T \times P) \times K}$ is projected into $\mathbf{x} \in \RR^{B \times (T \times P) \times K}$ and $\mathbf{z} \in \RR^{B \times (T \times P) \times K}$, where $T{\times}P$ is the length of the sequence. Note that the order of the elements in the sequence depends on the serialization, which differs between the hindcast blocks. We use a bi-directional approach that converts $\mathbf{x}$ into $\mathbf{x}_o^{\prime}$ using a forward and a backward 1-D causal convolution, where $o\in\{f,b\}$ denotes the forward or backward pass. For each direction, $\B_o$, $\C_o$, and $\Delta_o$ are obtained by projection layers from $\mathbf{x}_o^{\prime}$, and $\bar{\A}_o$ and $\bar{\B}_o$ are computed using Eq.~\eqref{eq:03}. The selective SSM then uses Eq.~\eqref{eq:02} to obtain  $\mathbf{y}_{forward}$ and $\mathbf{y}_{backward}$ for the forward and backward pass, respectively. The final output $\mathbf{y}$ is obtained by gating $\mathbf{y}_{forward}$ and $\mathbf{y}_{backward}$ via SiLU($\mathbf{z}$) and adding them up. Finally, $\mathbf{y}$ is normalized and projected back linearly to $\RR^{B \times (T \times P) \times K}$. The complete algorithm for the Mamba block is described in the suppl.\ material. 
After the Mamba block, the hindcast block includes another LOAN layer followed by an MLP. The final output $\X$ is then deserialized at the end since the next hindcast block uses a different serialization.  


\textbf{Forecasting layer.} While the hindcast layers encode the sequence of past input variables into a $K$-dimensional vector per sampled point, i.e., $\X_{hindcast} \in \RR^{B \times 1 \times P \times K}$, the forecasting layers forecast the difference of daily mean river discharge $\Delta \X_{dis24}^{t+l}$ for each lead time $l$, using $\X_{hindcast}$ and meteorological forecasts $\XHRES^{t+1:t+L}$ as input, as shown in Fig.~\ref{fig:Model}. The forecast blocks have the same structure as the hindcast blocks except that the forecast block incorporates the meteorological forcing (HRES). This is done by projecting $\XHRES^{t+l}$ with a linear layer to $64$ dimensions, serializing it, and concatenating it with the input $\X$ as illustrated in Fig.~\ref{fig:Mamba_block}. The processing of HRES is done sequentially, i.e., we have $L$ forecast blocks and the $l$-th forecast block processes $\XHRES^{t+l}$. We argue that this design is crucial to ensure that the temporal relationships between the meteorological forcing and the initial conditions are maintained. 

The output of all forecast blocks is processed by $L$ regression heads implemented as multi-layer perceptrons (MLP) where the output for the lead time $t+l$ is obtained as: 
\begin{align}
    \Delta\X_{dis24}^{t+l} ={}& \text{Linear}\Bigl(\text{ReLU}\Bigl(\text{Concat}\bigl(\text{Linear}(\X_{forecast}^{t+l}), \text{Linear}(\X_{forecast}^{t+1:t+L \textbackslash t+l})\bigr)\Bigr)\Bigr)  \,,\label{eq:07}
\end{align}
where $\X_{forecast}^{t+l}$ are the features from the $l$-th forecast block and $\X_{forecast}^{t+1:t+L \textbackslash t+l}$ are the concatenated features from all forecast blocks except of the $l$-th block. The linear layers project the input $\X_{forecast}^{t+l}$ or $\X_{forecast}^{t+1:t+L \textbackslash t+l}$ to 32 dimensions and the last linear projection estimates finally $\Delta\X_{dis24}^{t+l} \in \RR^{B \times 1 \times P \times 1}$.

\textbf{Training.} As already mentioned, we sample $P$ points around the globe for training. As a target value for training, we first use the river discharge data from the GloFAS reanalysis as ground truth and then fine-tune on sparse observations using data from the Global Runoff Data Centre (GRDC).
For GRDC fine-tuning, we take $P$ as the number of input points per sample and compute the loss only on points where GRDC observations are available without considering reanalysis data from GloFAS.
We obtain the target values by $\Delta \hat{\X}_{dis24}^{t+l}(p) = \hat{\X}_{dis24}^{t+l}(p) - \hat{\X}_{dis24}^{t-1}(p)$, where $\hat{\X}$ are the values from GloFAS or GRDC. For the training loss, we propose a weighted version of the mean-squared error (MSE) loss:
\begin{align}
    \mathcal{L} = \frac{1}{B \times P \times L}\sum_{b=1}^{B} \sum_{p=1}^{P} \sum_{l=1}^{L} {w}^{b,t+l}(p) \lVert \Delta \hat{\X}_{dis24}^{b,t+l}(p) - \Delta \X_{dis24}^{b,t+l}(p) \rVert_{2}^{2}  \,, \label{eq:09} 
\end{align}
where $B$ is the batch size. Since the severity of a flood is highly important for flood forecasting and severe floods occur rarely, the weighting factor ${w}^{b,t+l}(p)$ takes this into account. The severity of a flood is ranked by the statistical flood return period in years, which we denote by $r$ and ranges from $1.5$ to $500$. These ranges are also used in GloFAS. 
We note that a high return period event simply reflects statistical rarity in streamflow magnitude, and should not be equated with a flood event without additional context, e.g., thresholds or inundation. The return period is used here as a proxy indicator of hydrological extremity, which we call flood.
The severity of a flood is thus given by $\hat{r}^{t+l}(p)=\max_r \left\{r:\hat{\X}_{dis24}^{t+l}(p) \geq \theta_r(p) \right\}$, where $\theta_r$ is the statistical threshold for a given flood return period $r$. We also include the case $r{=}0$ with $\theta_r{=}0$ for defining events that are not floods. Using this notation, the weighting is thus given by   
\begin{align}
    \hat{w}^{b,t+l}(p) =
    \begin{cases}
    \hat{r}^{b,t+l}(p) & \text{if}\quad \hat{r}^{b,t+l}(p) > 1 \\
      1 & \text{otherwise.}\enspace
    \end{cases}\label{eq:10}
\end{align}
We thus weight the loss based on the flood return period if a flood occurred at location $p$ and time $t+l$, and we use $1$ if there has been no flood. 
We further weight the loss with $\hat{u}^{b,t+l} = e^{\alpha( L - l + 1)}$, where we give a higher weight to a shorter lead time $l$ and use $\alpha{=}0.25$. This compensates for the sequential structure of the forecast blocks where each forecast block takes the features of the previous block as input.
The final weight is thus given by ${w}^{b,t+l}(p) = \hat{u}^{b,t+l} \hat{w}^{b,t+l}(p)$. 
Since river discharge exhibits a very large dynamic with varying orders of magnitude, we transform the discharge values by $\text{sign}(\Delta \hat{x}) \text{log}(1+\lvert \Delta \hat{x} \rvert)$. We evaluate the impact of the weighting in Table \ref{table:ablation} (a) and provide more details in the suppl.\ material. For inference, we can forecast floods for any set of points or densely as in Fig.~\ref{fig:teaser}.

\section{Experimental results}
\label{sec:experimental_results}

\paragraph{Dataset.} We obtain data for river discharge from the ECMWF GloFAS reanalysis \cite{GloFAS_Reanalysis}.
It is generated by forcing the LISFLOOD hydrological model \cite{LISFLOOD} using meteorological data from ERA5 \cite{ERA5}.
GloFAS reanalysis combines physics-based simulation with observations to generate a consistent reconstruction of the past.
The dataset is provided as a daily averaged discharge on a global coverage at $3$ arcmin grid ($0.05^{\circ}$).
We use the GloFAS reanalysis as a target discharge for training and testing the model in Sec.~\ref{sec:results_reanalysis}. The ablation studies are done using GloFAS reanalysis over Europe. In addition, we fine-tune and test the model on observational GRDC river discharge data in Sec.~\ref{sec:results_obs}.
Flood thresholds are determined using return periods for individual points and are calculated from the long-term data. The thresholds allow for the identification of a flood when the threshold is surpassed.

\paragraph{Evaluation metrics.} We evaluate the performance of RiverMamba on both GloFAS reanalysis and GRDC, where diagnostic GRDC stations are available (3366 stations). For evaluation, we use common metrics like the coefficient of determination (R2), Kling–Gupta efficiency (KGE), and the averaged F1-score for floods with return periods of $1.5$ to $20$ years.
Details about these metrics can be found in the suppl.\ material. 
We train on the years 1979-2018, validate on 2019-2020, and test on 2021-2024.
All evaluation points are gauged stations and temporally out-of-sample. Results on ungauged stations are also available in the suppl.\ material. The metrics are calculated on the time series at single grid points and then averaged over all points.

\paragraph{Baselines.} We compare RiverMamba to persistence, climatology, and the state-of-the-art deep learning Encoder-Decoder LSTM of Google's operational flood forecasting system \cite{Google_nature}.
For the LSTM model, we followed the same protocol as originally proposed in \cite{Google_nature}, which considers only temporal context but does not include any spatial connections. The space filling curves are thus not used in combination with the LSTM baseline. To ensure a consistent evaluation, we train LSTM on the same input data as RiverMamba. All results in the paper are obtained with our trained LSTM.
A comparison with the published reforecasts by Google's LSTM \cite{Google_nature} is also available in the suppl.\ material.
For evaluation on GRDC observations, we additionally compare our approach to the reforecast version of the state-of-the-art operational GloFAS forecasting system operated by ECMWF \cite{GloFAS, GloFAS_Forecast}.
More details about dataset, evaluation metrics and baselines are provided in the suppl.\ material.

\subsection{Experiments on GloFAS river discharge reanalysis}
\label{sec:results_reanalysis}

\begin{figure}[!b]
  \centering
  \includegraphics[draft=\draft, width=.99\textwidth]{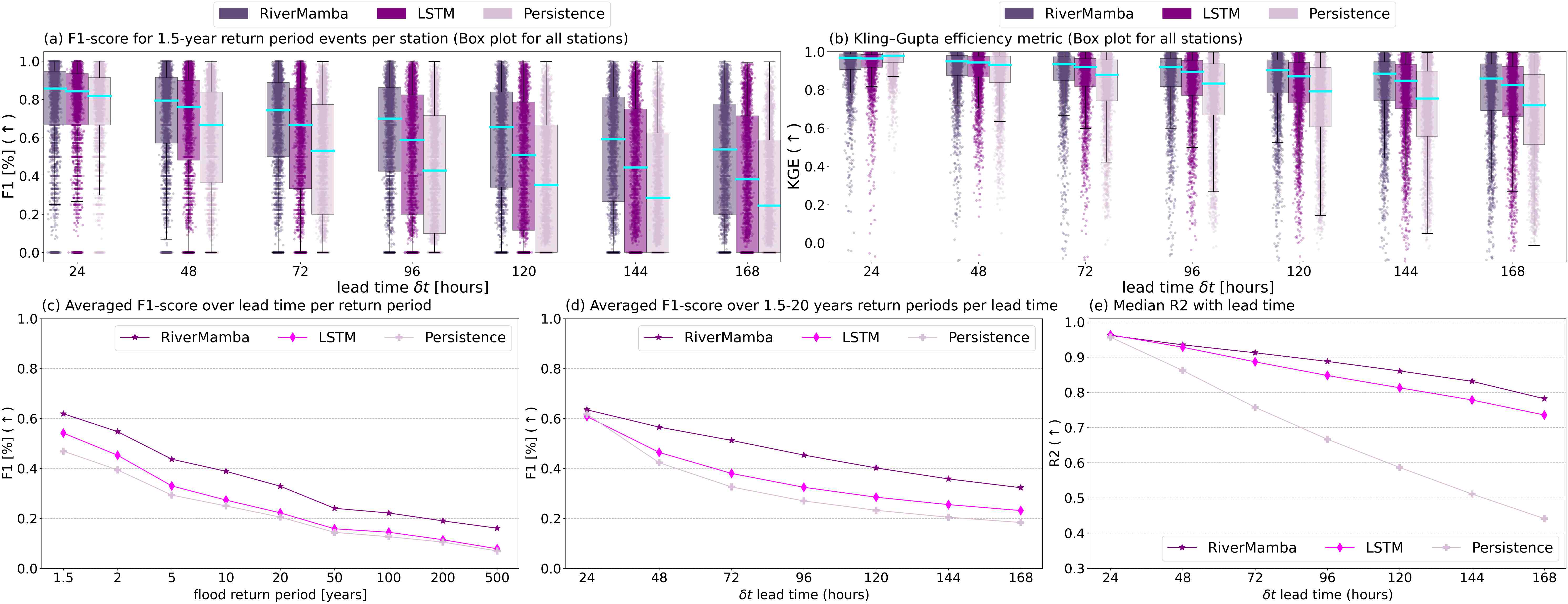}
  \caption{Results on GloFAS reanalysis across lead times and flood return periods.}\label{fig:reanalysis_baselines}
\end{figure}

The quantitative results are shown in Table~\ref{table:reanalysis_baselines}. As can be seen, the climatology baseline performs poorly, as the dynamic in local river discharge varies a lot over time, highlighting the difficulty in predicting flows. We therefore exclude it in Fig.~\ref{fig:reanalysis_baselines} (a) that shows F1-score for floods with a $1.5$-year return period and (b) KGE for river discharge for different lead times from $24$ to $168$ hours. The boxes show distribution quartiles and the evaluation points are represented as points along the y-axis. Fig.~\ref{fig:reanalysis_baselines} (d) shows the F1-score averaged over return periods of $1.5$ to $20$ years and (e) shows the median R2 for river discharge. The persistence baseline predicts the future discharge as the same value of the discharge at time $t$. This achieves good prediction for the short-term forecast, however, the prediction skill drops with lead time. While LSTM outperforms the persistence baseline, RiverMamba outperforms all baselines and methods on all metrics as shown in Table~\ref{table:reanalysis_baselines}. In particular for lead times above 48 hours, the performance gap between RiverMamba and LSTM is large. 
We attribute this to the receptive field and the spatio-temporal modeling of RiverMamba. Fig.~\ref{fig:reanalysis_baselines} (c) plots the F1-score averaged over $24$ to $168$ hours lead time for different flood return periods. The results show that RiverMamba outperforms the other approaches both for more frequent floods and rare severe floods that occur statistically only every 500 years. More results are in the suppl.\ material. 
In the following, we discuss a set of ablation studies that are not performed globally but over Europe.

\begin{table}[!h]
  \vspace{-1mm}
  \caption{Results on GloFAS-Reanalysis. \gray{($\pm$)} denotes the standard deviation for 3 runs.}
  \label{table:reanalysis_baselines}
  \centering
  \small
  \tabcolsep=1.0pt\relax
  \setlength\extrarowheight{0pt}
  \begin{tabular}{r *{8}l}
    \toprule
    & & \multicolumn{3}{c}{Validation (2019-2020)} & & \multicolumn{3}{c}{Test (2021-2024)} \\
    \midrule
    Model & & R2 ($\uparrow$) & KGE ($\uparrow$) & F1 ($\uparrow$) & & R2 ($\uparrow$) & KGE ($\uparrow$) & F1 ($\uparrow$) \\
    \midrule
    Climatology & & \colorm{0.1175} &  \colorm{0.2618} &  \colorf{\NA} & &  \colorm{0.1352} &  \colorm{0.2449} &  \colorf{\NA} \\
    Persistence & & \colorm{0.6778} &  \colorm{0.8380} &  \colorf{0.3138} & &  \colorm{0.6833} &  \colorm{0.8412} &  \colorf{0.3223} \\
   \midrule
   LSTM & & \colorm{0.8539}{\footnotesize\gray{$\pm$0.0031}} & \colorm{0.8931}{\footnotesize\gray{$\pm$0.0034}} & \colorf{0.3511}{\footnotesize\gray{$\pm$0.0068}} & & \colorm{0.8485}{\footnotesize\gray{$\pm$0.0021}} & \colorm{0.8924}{\footnotesize\gray{$\pm$0.0029}} & \colorf{0.3582}{\footnotesize\gray{$\pm$0.0058}} \\
   \midrule
  \bestc RiverMamba & \bestc & \bestc  \textbf{\colorm{0.8803}}{\footnotesize\gray{$\pm$0.0043}} & \bestc \textbf{\colorm{0.9137}}{\footnotesize\gray{$\pm$0.0026}} & \bestc \textbf{\colorf{0.4540}}{\footnotesize\gray{$\pm$0.0056}} & \bestc & \bestc \textbf{\colorm{0.8728}}{\footnotesize\gray{$\pm$0.0013}} & \bestc \textbf{\colorm{0.9125}}{\footnotesize\gray{$\pm$0.0008}} & \bestc \textbf{\colorf{0.4589}}{\footnotesize\gray{$\pm$0.0080}} \\
    \bottomrule
\end{tabular}
\vspace{-1mm}
\end{table}

\begin{table}[!h]
  \caption{Ablation studies on the validation set over Europe.}
  \label{table:ablation}
  \centering
  \tabcolsep=1.6pt\relax
  \setlength\extrarowheight{0pt}
  \begin{tabular}{*{11}c}
  
  \multicolumn{3}{l}{(a) Objective function} & & \multicolumn{3}{l}{(b) Location Embedding} & & \multicolumn{3}{l}{(c) Forecasting strategy}\\
      \cmidrule[\heavyrulewidth]{1-3} \cmidrule[\heavyrulewidth]{5-7} \cmidrule[\heavyrulewidth]{9-11}
      
      $\hat{w}$ & $\hat{u} $  & KGE | F1 ($\uparrow$)  & & LOAN$_{(hind)}$ & LOAN$_{(forc)}$ & KGE | F1 ($\uparrow$) & & S-HRES & T-HRES & KGE | F1 ($\uparrow$) \\

    
      \cmidrule[\heavyrulewidth]{1-3} \cmidrule[\heavyrulewidth]{5-7} \cmidrule[\heavyrulewidth]{9-11}
      
      \xmark & \xmark & \colorm{0.9086} | \colorf{0.2236} & & \xmark & \xmark & \colorm{0.9183} | \colorf{0.2790} & & \xmark & \cmark & \colorm{0.8862} | \colorf{0.2030} \\
      
      \cmark & \xmark & \colorm{0.9127} | \colorf{0.2859} & & \cmark & \xmark & \colorm{0.9160} | \colorf{0.2827} & & \cmark & \xmark & \colorm{0.8869} | \colorf{0.2268} \\

      \xmark & \cmark & \colorm{0.9136} | \colorf{0.2593} & & \xmark & \cmark & \colorm{0.9166} | \textbf{\colorf{0.2931}} & & \cmark \bestc  & \cmark \bestc & \bestc \textbf{\colorm{0.9205}} | \textbf{\colorf{0.2875}} \\
      
      \cmark \bestc & \cmark \bestc & \bestc \textbf{\colorm{0.9205} | \colorf{0.2875}} & & \cmark \bestc & \cmark \bestc & \bestc \textbf{\colorm{0.9205}} | \colorf{0.2875} & & & & \\

  \cmidrule[\heavyrulewidth]{1-3} \cmidrule[\heavyrulewidth]{5-7} \cmidrule[\heavyrulewidth]{9-11}
  \end{tabular}
\end{table}

\textbf{Objective functions.}~ In Table \ref{table:ablation} (a), we evaluate the impact of the weighting factor in the loss \eqref{eq:09}, which is based on $\hat{w}$ Eq.~\eqref{eq:10} and $\hat{u}$. The results show that both terms improve the results. $\hat{w}$ is important to focus on rare and more severe floods, increasing the F1 metric substantially (second row). $\hat{u}$ gives more weight to the forecast in the near future where $\XHRES$ is more reliable, which is important due to the sequential structure of the forecast module. Using only $\hat{u}$ (third row) improves the results on both KGE and F1 metrics. Using both $\hat{w}$ and $\hat{u}$ (fourth row) gives the best results.    

    
\textbf{Location embedding.}~ In Table \ref{table:ablation} (b), we show the benefit of using LOAN. In the first row, we duplicate the static features along the $T$ dimension and concatenate them with the dynamic input. Using the LOAN layer in the hindcast (second row) or forecast blocks (third row) increases the F1 score but decreases KGE. Using LOAN in both hindcast and forecast blocks balances the metrics (fourth row).

\textbf{Forecasting strategy.}~ Table \ref{table:ablation} (c) evaluates the impact of spatio-temporal modeling in the forecast module. In the first row, we remove the spatial relations in the forecast module by replacing the forecast blocks by point-wise MLPs. In this way, the data is processed after the last hindcast layer temporally but not spatially. This makes the model unaware of the spatial biases in the meteorological forcing $\XHRES$. 
The second row denotes a setup where the forecast blocks do not get the features from the previous forecast block (Fig.~\ref{fig:Model}) but directly from the last hindcast layer. In this case, we forecast river discharge for each lead time independently. The results show that in both cases the performance drops compared to our approach (third row), demonstrating the importance of spatio-temporal modeling. More ablation studies can be found in the suppl.\ material. 

\subsection{Experiments on GRDC observational river discharge}
\label{sec:results_obs}

\begin{figure}[!h]
  \centering
  \includegraphics[draft=\draft, width=.99\textwidth]{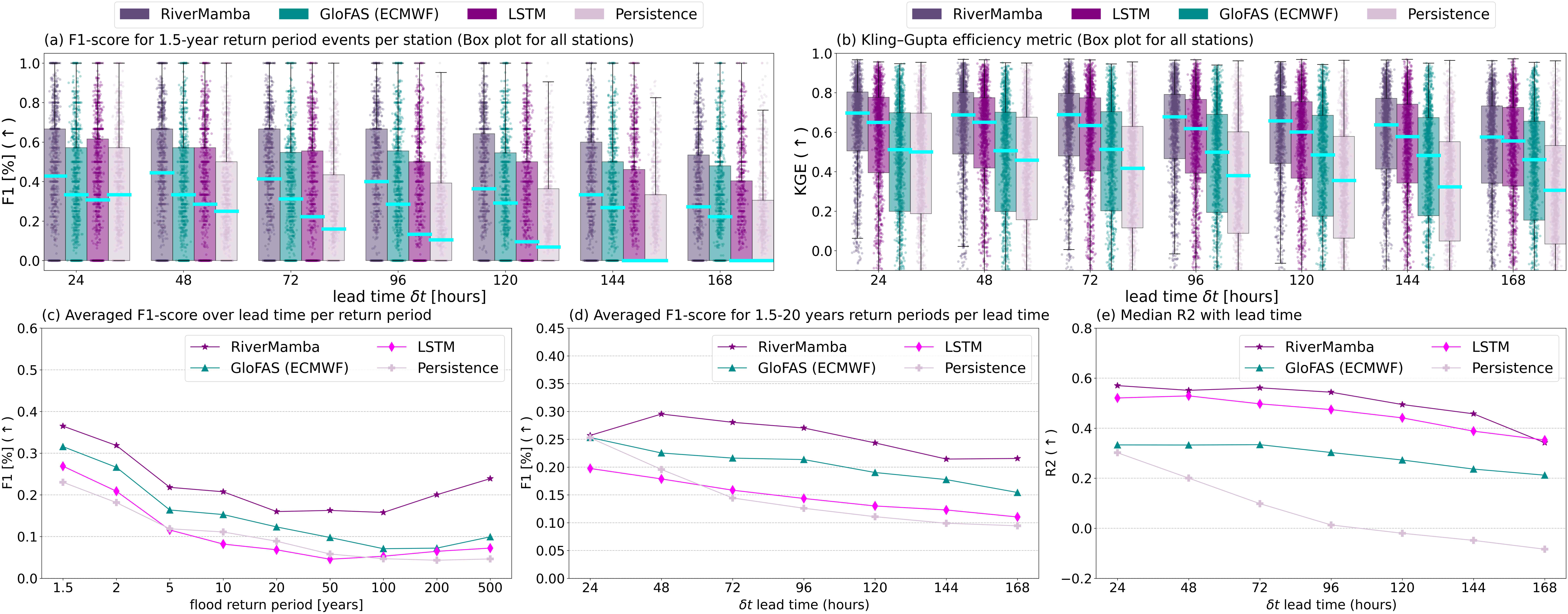}
  \caption{Results on gauged GRDC observations across lead times and flood return periods.} \label{fig:obs_baseline}
\end{figure}

\begin{table}[t]
  \caption{Results on GRDC gauged stations. \gray{($\pm$)} denotes the standard deviation for 3 runs.}
  \label{table:grdc_baselines}
  \centering
  \small
  \tabcolsep=1.0pt\relax
  \setlength\extrarowheight{0pt}
  \begin{tabular}{r *{8}l}
    \toprule
    & & \multicolumn{3}{c}{Validation (2019-2020)} & & \multicolumn{3}{c}{Test (2021-2023)} \\
    \midrule
    Model & & R2 ($\uparrow$) & KGE ($\uparrow$) & F1 ($\uparrow$) & & R2 ($\uparrow$) & KGE ($\uparrow$) & F1 ($\uparrow$) \\
    \midrule
    Climatology & & \colorm{-0.0002} &  \colorm{0.1342} &  \colorf{\NA} & &  \colorm{-0.0013} &  \colorm{0.0870} &  \colorf{\NA} \\
    Persistence & & \colorm{0.1682} & \colorm{0.4569} &  \colorf{0.1626} & & \colorm{0.0660} &  \colorm{0.3918} & \colorf{0.1462} \\
   \midrule
     GloFAS & & \colorm{0.3713} & \colorm{0.5412} & \colorf{0.2135} & & \colorm{0.2892} & \colorm{0.4944} & \colorf{0.2044} \\
    LSTM & & \colorm{0.5437}{\footnotesize\gray{$\pm$0.0025}} & \colorm{0.6572}{\footnotesize\gray{$\pm$0.0010}} & \colorf{0.1724}{\footnotesize\gray{$\pm$0.0017}} & & \colorm{0.4615}{\footnotesize\gray{$\pm$0.0039}} & \colorm{0.6141}{\footnotesize\gray{$\pm$0.0018}} & \colorf{0.1475}{\footnotesize\gray{$\pm$0.0014}} \\
    \midrule
    \bestc RiverMamba & \bestc & \bestc  \textbf{\colorm{0.5943}}{\footnotesize\gray{$\pm$0.0016}} & \bestc \textbf{\colorm{0.7015}}{\footnotesize\gray{$\pm$0.0007}} & \bestc \textbf{\colorf{0.2577}}{\footnotesize\gray{$\pm$0.0046}} & \bestc & \bestc  \textbf{\colorm{0.5057}}{\footnotesize\gray{$\pm$0.0028}} & \bestc \textbf{\colorm{0.6612}}{\footnotesize\gray{$\pm$0.0010}} & \bestc \textbf{\colorf{0.2427}}{\footnotesize\gray{$\pm$0.0111}} \\
    \bottomrule
\end{tabular}
\end{table}

Table \ref{table:grdc_baselines} reports the performance on GRDC river discharge observations at gauged stations, which also includes the physics-based GloFAS reforecast model. As previously, Fig.~\ref{fig:obs_baseline} compares the forecast performance across multiple lead times and flood return periods. Compared to the results on GloFAS reanalysis (Table~\ref{table:reanalysis_baselines}), all models show a noticeable drop in performance when evaluated on GRDC observations (Table~\ref{table:grdc_baselines}). 
This decline likely stems from the fact that GloFAS simulates primarily naturalized discharge, with simplified representations of major reservoirs \cite{GloFAS_Reanalysis, ZAJAC2017}, whereas GRDC reflects fully regulated flow, influenced by complex and unobserved human activities, such as dam operations and irrigation. 
This introduces biases that models cannot learn, especially in the absence of globally available data representing human water management, highlighting the challenge of predicting discharge under human-modified conditions. The results show that traditional baselines such as Climatology and Persistence perform poorly. GloFAS performs much better than the baselines, but the R2 and KGE values are rather low due to the mentioned differences of physics-based models and observations. RiverMamba consistently outperforms the other methods for all metrics. Notably, RiverMamba shows less degradation in F1-score with increasing lead time, highlighting its strength in medium-range flood forecasting. More results are in the suppl.\ material.

\section{Conclusions and limitations}
\label{sec:conclusion}
We introduced RiverMamba, a novel deep learning approach for global, medium-range river discharge and flood forecasting. Due to its efficient structure and specialized scanning paths, RiverMamba maintains a very large receptive field, while scaling linearly with respect to the number of sampled points. As a result, RiverMamba is capable of forecasting high-resolution ($0.05^\circ$) global river discharge maps. Further, the spatio-temporal modeling of the forecast blocks incorporates meteorological forcing and ensures a consistent forecast through space and time. Our analysis reveals that RiverMamba outperforms operational state-of-the-art deep learning and physics-based models on both reanalysis and observational data. While the results show major advancements in river discharge and flood forecasting, the approach has some limitations. For a real operational setting, only data can be used that is available until the current day $t$. For instance, ERA5-Land is publicly available after 5 days whereas we assumed that ERA5-Land is already available after 1 day, i.e., $t-1$. ERA5-Land, however, could be substituted by other near real-time reanalysis data that is earlier available or analysis data until day $t$. 
It also needs to be mentioned that observational data are affected by human interventions like dams and there is a need to integrate such interventions in the model. As it is the case for operational systems, floods are not always correctly forecast. The causes of the errors need to be analyzed more in detail. The forecast errors can be caused by human interventions, errors in the weather forecast for meteorological forcing or river attributes, the rarity of floods, or bias in the data and re-analysis. Given such errors, it is desirable to extend the model such that it estimates its uncertainty for the forecast as well.

Besides these limitations, RiverMamba has the potential for an operational medium-range river discharge and flood forecasting system that predicts flood risks, in particular extreme floods, more accurately and at higher resolution than existing systems. This is essential for stakeholders to make decisions for an effective flood risk mitigation strategy and an early warning system to protect citizens. 


\section{Acknowledgments and Disclosure of Funding}

This work was supported by the Federal Ministry of Research, Technology, and Space under grant no.\ 16IS24075C RAINA and by the Deutsche Forschungsgemeinschaft (DFG, German Research Foundation) – SFB 1502/1–2022 – project no. 450058266 within the Collaborative Research Center (CRC) for the project Regional Climate Change: Disentangling the Role of Land Use and Water Management (DETECT).
We acknowledge EuroHPC Joint Undertaking for awarding us access to Leonardo at CINECA, Italy, through EuroHPC Regular Access Call - proposal No. EHPC-REG-2025R01-218. The authors also gratefully acknowledge the granted access to the Marvin cluster hosted by the University of Bonn. 
Furthermore, we acknowledge the ESM Testprojekt project for supporting this study by providing computing time on the esmtst partition of the supercomputer JUWELS-BOOSTER at J{\"u}lich Supercomputing Centre (JSC).
Finally, we thank Lars Doorenbos for proofreading the manuscript. 
The Mamba logos were generated with Microsoft Designer.


\medskip


\appendix
\label{sec:Appendix}

\begin{center}
{\LARGE
\bfseries 
RiverMamba: A State Space Model for Global River
Discharge and Flood Forecasting} \\
\vspace{0.4cm}
{\Large
\mambalogo
Technical Appendices and Supplementary Material \\
}
\end{center}

{\Large Table of Contents}
\vspace{0.3cm}

\startcontents[sections]
\printcontents[sections]{l}{1}{\setcounter{tocdepth}{3}}
\clearpage

\section{Dataset}
\label{sec:Appendix_dataset}

\subsection{GloFAS reanalysis data}
\label{sec:Appendix_GloFAS_reanalysis_data}
\begin{figure}[!h]
  \centering
  \includegraphics[draft=\draft, width=.8\textwidth]{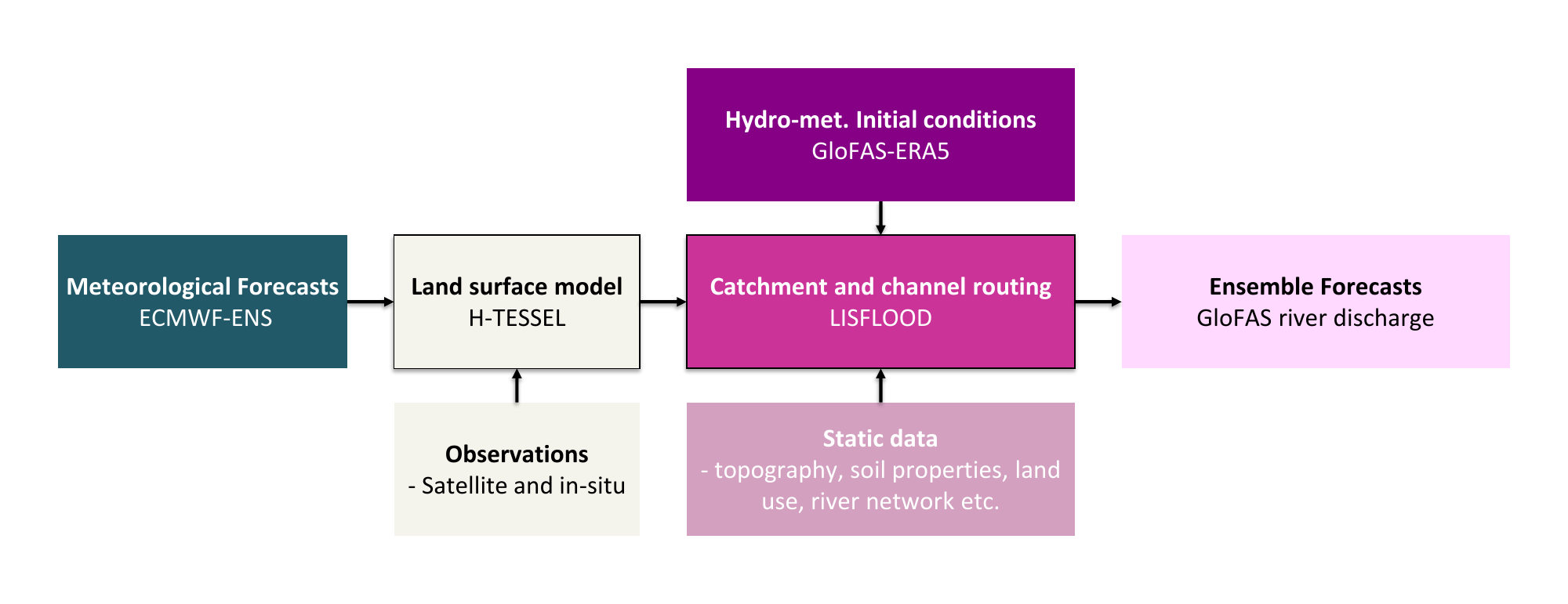}
  \caption{An overview of the key modules in the GloFAS forecasting system. GloFAS-ERA5 reanalysis uses ERA5 meteorological reanalysis data instead of ECMWF ensemble forecasts (ENS).
  Figure outline from \protect\cite{GloFAS_Reanalysis}.} \label{fig:glofas_system}
\end{figure}

The Global Flood Awareness System (GloFAS) is an operational system developed by the European Commission’s Joint Research Centre (JRC) and operated by ECMWF under the Copernicus Emergency Management Service (CEMS) \cite{GloFAS}. It provides real-time global-scale flood forecasts and a long-term hydrological reanalysis dataset, a key resource for flood risk assessment, climate impact studies, and machine learning applications. 
Fig. \ref{fig:glofas_system} shows the workflow of GloFAS to forecast river discharge and flood events.
The GloFAS-ERA5 reanalysis is the long-term retrospective component of GloFAS \cite{GloFAS_Reanalysis}. It delivers daily river discharge estimates from 1979 to present at a spatial resolution of $0.05^\circ$ ($\sim5$ km) and a global coverage ($90^\circ$N-$60^\circ$S, $180^\circ$W-$180^\circ$E). The reanalysis is generated by coupling surface and subsurface runoff from the ERA5 reanalysis, produced by the H-TESSEL and surface model \cite{HTESSEL} with the LISFLOOD hydrological and river routing model \cite{LISFLOOD}. While ERA5 runoff is computed at $\sim31$ km resolution and lacks spatial connectivity, it is downscaled to 0.05° using a nearest-neighbour approach and routed through LISFLOOD to simulate realistic river discharge (dis24, in $m^{3} s^{-1}$) across the global river network. The daily GloFAS reanalysis discharge data represents the mean value between 00:00 UTC previous day and 00:00 UTC current day. 
Similarly to GloFAS, there exists an early warning system for Europe (EFAS) with higher resolution \cite{EFAS}. In our work, GloFAS v4.0 is used for a global application. 
The dataset is publicly available on Climate Data Store and Early Warning Data Store (EWDS) \colorh{\url{https://doi.org/10.24381/cds.a4fdd6b9}}. 
Table \ref{table:glofas_variables} explains the details of four variables we took from the GloFAS reanalysis dataset as the model inputs. The GloFAS-ERA5 reanalysis supports the derivation of flood thresholds (i.e., 2-, 5-, and 20-year return periods) and serves as the initial condition for real-time forecasts such as GloFAS-30d and GloFAS-Seasonal. More details about GloFAS can be found in \cite{GloFAS_Reanalysis}.

Despite its broad applicability, the GloFAS-ERA5 reanalysis is subject to several limitations that researchers should be aware of. Regional biases have been identified, which may stem from uncertainties in the meteorological forcing provided by ERA5, the representation of runoff generation processes within the H-TESSEL land surface model, and limitations in the calibration of the LISFLOOD routing model. When sufficient observational discharge data are available, the LISFLOOD model is calibrated locally for each river catchment larger than $500$ km$^2$, and each calibrated catchment has its own optimized parameter set. This could give the model better performance at the local scale but reduce its generalization ability. Additionally, anthropogenic influences such as dams and reservoirs are incorporated using simplified operational rules, largely due to the lack of globally available real-time release data. Finally, as the dataset is entirely driven by ERA5, it inherits known deficiencies of the reanalysis, including biases in precipitation and the absence of river discharge data assimilation, which may affect the realism of simulated hydrological conditions in some regions.

\begin{table}[h!]
\small
 \caption{Details about the processed variables from GloFAS reanalysis \protect\cite{GloFAS_Reanalysis}.}\label{table:glofas_variables}
 \centering
 \setlength\tabcolsep{3pt}
 \begin{tabular}{*{5}l}
    \toprule
        Variable & Long name & Unit & Height & Surface parameters  \\
    \midrule
        acc\_rod24 & runoff water equivalent &kg/m$^{2}$ & surface and subsurface  & accumulated \\
        dis24      & river discharge & m$^{3}$/s & surface & averaged over 24 hours \\
                   & in the last 24 hours &  &  & \\        
        sd         & snow depth water equivalent & kg/m$^{2}$ & surface & instantaneous \\
        swi        & soil wetness index & - & root zone & instantaneous \\

\bottomrule
\end{tabular}
\end{table}

\subsection{GRDC observational river discharge data}
\label{sec:Appendix_GRDC_data}

We obtain observational river discharge from the Global Runoff Data Centre (GRDC) which is an international data repository that provides access to quality-controlled river discharge observation data from around the world. The GRDC dataset contains time series of daily and monthly river discharge data from over $10000$ hydrological gauging stations across more than 160 countries from small headwater catchments ($\sim 10$ km$^2$ drainage area) to very large river catchment like the Amazon river ($5$ million km$^2$ drainage area). 
GRDC data can be obtained from \colorh{\url{https://grdc.bafg.de/}}. 
All GRDC daily time series measured the value set at 00:00 of the beginning of the day (left-labeled). To keep our evaluation consistent with \cite{Google_nature}, we used the GRDC dataset as the benchmark to evaluate the model performance and followed a similar data processing workflow as in \cite{GloFAS_Forecast, Google_nature}. We first removed the catchments with a drainage area smaller than $500$ km$^2$ and obtained $5524$ GRDC stations to avoid very big discrepancies between the drainage area defined in GRDC and in the GloFAS dataset (\ref{sec:Appendix_GloFAS_reanalysis_data}). Next, we geo-located the GRDC stations to compare them with the GloFAS drainage network and removed the GRDC stations with more than 10\% of drainage area differences. For geo-location, we projected the points on the GloFAS grid, compared each point with its $9$ nearest points, and took the location with the highest KGE value.
Finally, the GRDC stations with no ERA5-Land reanalysis data were discarded. This resulted in $3366$ stations for the global evaluation. This narrowed down the global median drainage area difference to $2.21$\% with an interquartile range of $0.86$\% to $4.73$\%. The discharge observations are recorded at a daily time scale, with the unit $m^3 s^{-1}$ and converted from local time zone to 00:00 UTC via linear interpolation. 
For evaluation, the GRDC observational time series, which are originally left-labeled, were explicitly converted to right-labeled time series to ensure a temporal consistency with the right-labeled predictions from the GloFAS simulations, RiverMamba, LSTM baseline, and Google reforecast. The F1 scores reported for the GRDC evaluation are based on synchronized detection windows between model predictions and observations. 

\begin{figure}[!h]
  \centering
  \includegraphics[width=.99\textwidth]{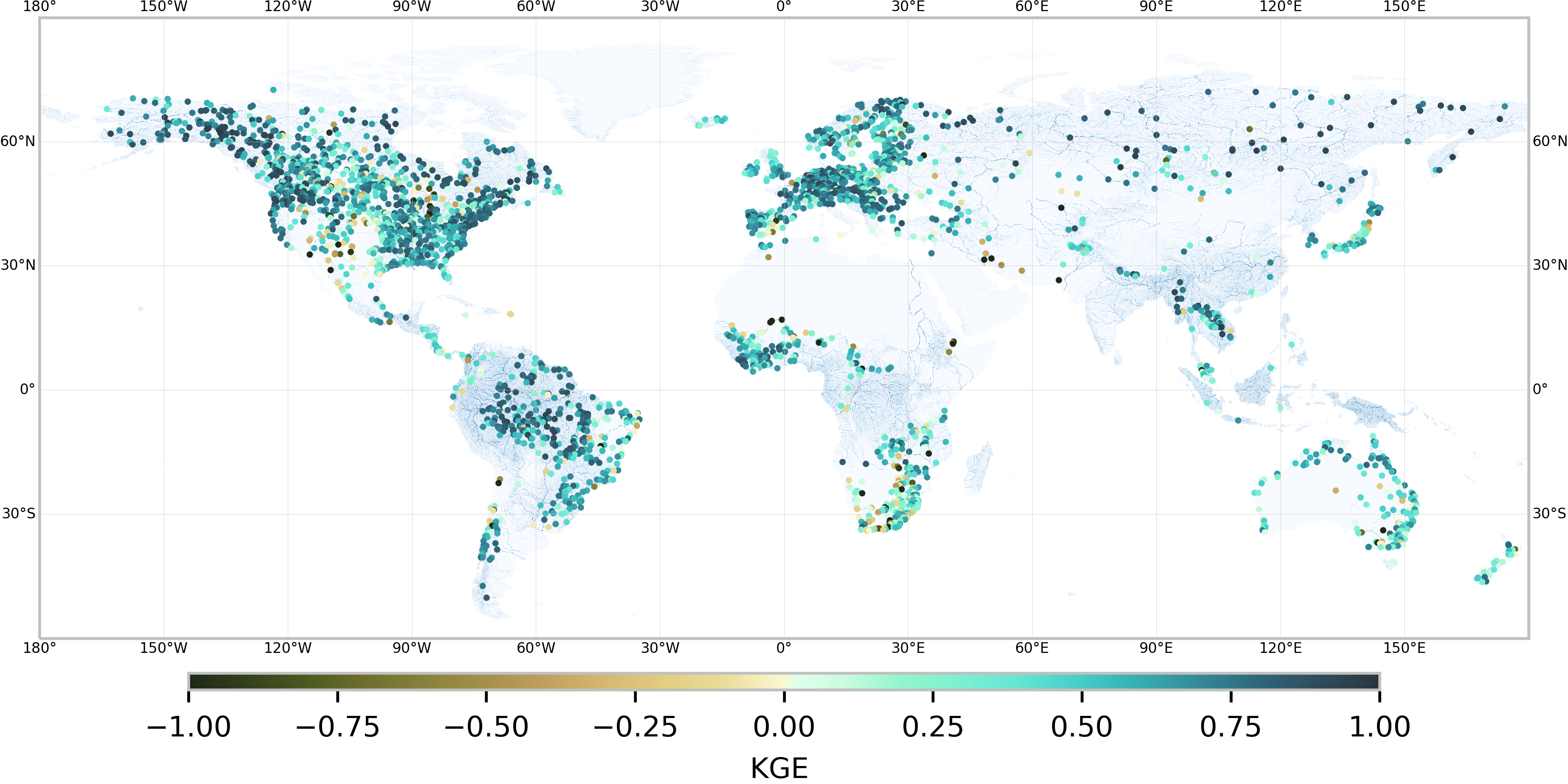}
  \caption{Locations of selected $3366$ GRDC stations used for training and evaluating the RiverMamba model. The colorbar shows the KGE value of GloFAS reanalysis discharge data against the GRDC discharge observations.}\label{fig:GRDC_stations}
\end{figure}

\begin{figure}[!h]
  \centering
  \includegraphics[width=.99\textwidth]{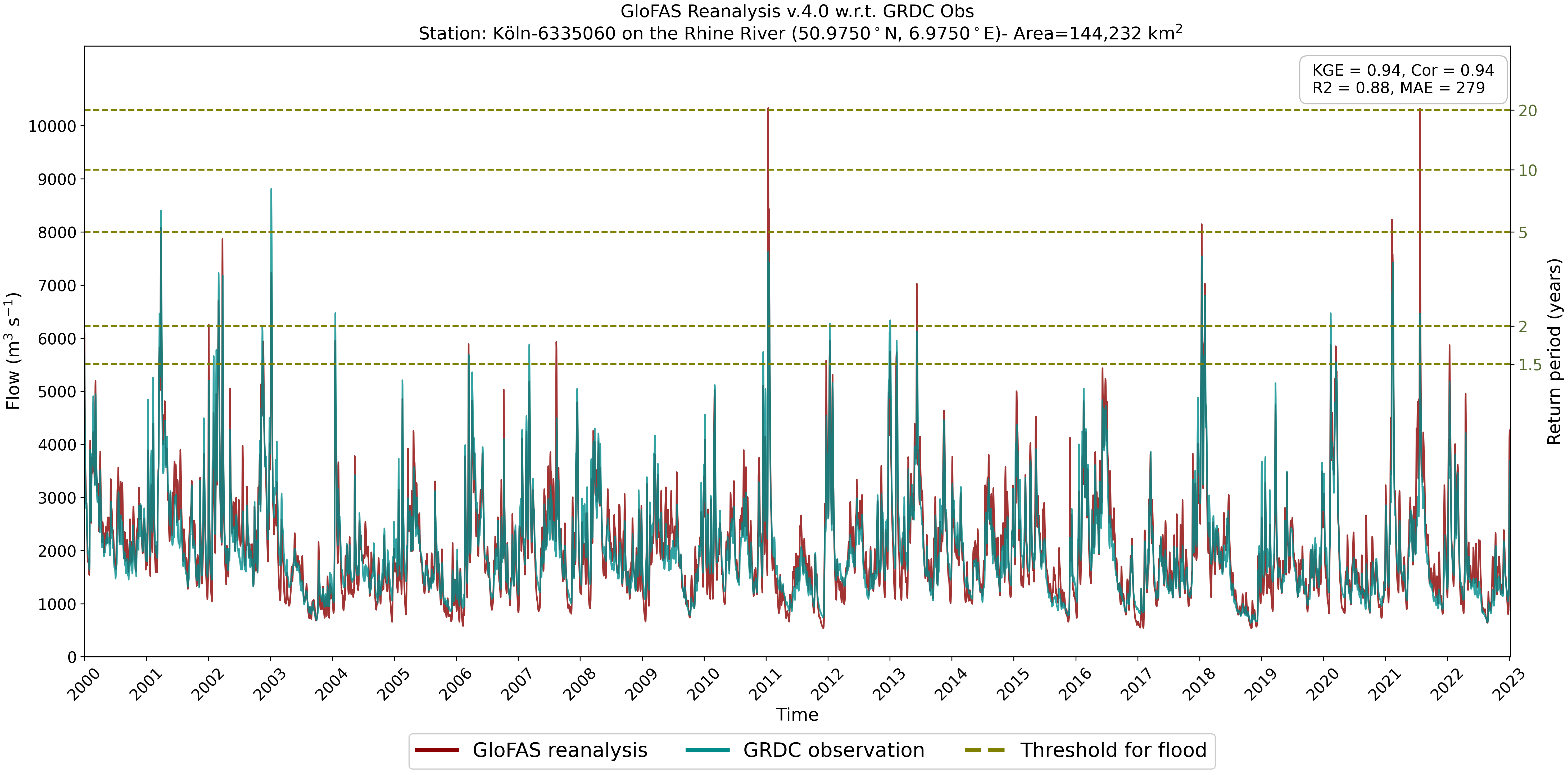}
  \caption{Hydrograph for GloFAS-reanalysis (red line) from 1 January 2000 to 31 December 2023 and observations (dark cyan line), for a gauging station on the Rhine River. The top-right box displays summary statistics from the reanalysis's evaluation against the observations.}\label{fig:KOELN}
\end{figure}

Fig.~\ref{fig:GRDC_stations} shows the KGE values of GloFAS reanalysis data against the GRDC observation at the locations of all 3366 selected stations. In general, there is a good agreement between GloFAS and GRDC data globally, with a median KGE at $0.61$ and an interquartile range of $0.36$ to $0.77$. In regions like south America, south Africa and Australia, the GloFAS reanalysis data have more inconsistency compared to the observations. Fig.~\ref{fig:KOELN} shows an exemplar hydrograph for a gauged station. 

It is important to note that, compared to the GRDC observations, the GloFAS reanalysis dataset only simulates the naturalized flow without considering realistic human interventions such as dams, reservoirs, diversions, irrigation withdrawals, and other water management practices, and this can be a major source of bias in GloFAS compared to the GRDC data. 

\subsection{ERA5-Land data}
\label{sec:Appendix_ERA5_Land_data}

In contrast to operational GloFAS which uses ERA5 as forcing data, we use ERA5-Land as an initial land surface condition for the forecast. The ERA5-Land reanalysis data are described in \cite{ERA5-Land} and retrieved from the Copernicus Climate Change Service (C3S) Climate Data Store (CDS). \colorh{\url{https://doi.org/10.24381/cds.e2161bac}}. 
We processed $14$ instantaneous state variables at $00$:$00$ UTC and $18$ daily accumulated state variables ($00$:$00$ UTC previous day to $00$:$00$ UTC current day). More details are provided in Table \ref{table:era5-land}.
ERA5-Land is provided at $0.1^\circ \times 0.1^\circ$. We mapped the data onto the GloFAS regular latitude and longitude (Plate Carr\'ee projection) using bilinear mapping and implemented by Zhuang et al.~\cite{xESMF}.

\subsection{HRES}
\label{sec:Appendix_HRES_data}

Meteorological conditions serve as the driving forces behind hydrological processes. These are necessary for forecasting the river discharge and potential floods.
We use the deterministic forecast of the ECMWF Integrated Forecast System (IFS) High Resolution (HRES) atmospheric model.
The HRES data were obtained from the ECMWF Archive Catalogue \colorh{\url{https://www.ecmwf.int/en/forecasts/dataset/operational-archive}}.
We use HRES up to $7$ days lead time and once per day at 00:00 UTC. 
The processed data are similar to \cite{Google_nature} except that we used total evaporation as an additional forcing variable. In addition, we do not use any forecast for nowcasting at time step $t$. The processed data include $2$ instantaneous and $5$ daily accumulated variables at the surface level forecasts. 
Technical details regarding the meteorological forcing variables are provided in Tables \ref{table:HRES}.

While the operational archive provides data from $1985$, the quality and the resolution of the forecasts in the earlier years are not sufficient for our application. Therefore, we only processed and used data from $2010$ to $2024$. Data before $2010$ were replaced by ERA5-Land. To match the resolution of the target GloFAS grid, we regridded HRES to $0.05^\circ \times 0.05^\circ$ regular grid.

\begin{table}[h!]
\small
 \caption{Details about the processed variables from ERA5-Land reanalysis \protect\cite{ERA5-Land}.} \label{table:era5-land}
 \centering
 \setlength\tabcolsep{2.5pt}
 \begin{tabular}{*{5}l}
    \toprule
        Variable & Long name & Unit & Height & Surface parameters  \\
    \midrule
        d2m    & 2m dewpoint temperature & K & 2m & instantaneous\\
        e      & total evaporation & m of water equivalent& surface & accumulated\\
        es     & snow evaporation & m of water equivalent& surface & accumulated\\
        evabs  & evaporation from bare soil & m of water equivalent & surface & accumulated\\    
        evaow  & evaporation from open water & m of water & surface & accumulated\\
               & surfaces excluding oceans & equivalent & & \\     
        evatc  & evaporation from the top of  & m of water equivalent & surface & accumulated\\
               & canopy &  & & \\  
        evavt  & evaporation from vegetation  & m of water equivalent & surface & accumulated\\
               & transpiration &  & & \\ 
        lai\_hv& leaf area index & m$^{2}$/m$^{2}$ & 2m & instantaneous\\
               & high vegetation &  & & \\  
        lai\_lv& leaf area index & m$^{2}$/m$^{2}$ & 2m & instantaneous\\
               & low vegetation &  & & \\  

        pev    & potential evaporation & m & 2m & accumulated\\
        sf     & snowfall & m of water equivalent & surface & accumulated\\
        skt    & skin temperature & K & surface & instantaneous\\
        slhf   & surface latent heat flux & J/m$^{2}$  & surface & accumulated\\
        smlt   & snowmelt & m of water equivalent & surface & accumulated\\
        sp     & surface pressure & Pa & surface & instantaneous \\
        src    & skin reservoir content & m of water equivalent & surface & instantaneous \\
        sro    & surface runoff & m & surface & accumulated \\
        sshf   & surface sensible heat flux & J/m$^{2}$ & surface & accumulated\\
        ssr    & surface net solar radiation & J/m$^{2}$ & surface & accumulated\\
        ssrd   & surface solar radiation  & J/m$^{2}$ & surface & accumulated\\
               & downwards &  & & \\ 
        ssro   & subsurface runoff & m & subsurface & accumulated \\
        stl1   & soil temperature & K &  soil layer (0 - 7 cm) & instantaneous \\
        str    & surface net thermal radiation & J/m$^{2}$ & surface & accumulated\\
        strd   & surface thermal radiation & J/m$^{2}$ & surface & accumulated\\
               & downwards &  & & \\ 
        
        swvl1  & volumetric soil water & m$^{3}$/m$^{3}$ & soil layer (0 - 7 cm) & instantaneous \\
        swvl2  & volumetric soil water & m$^{3}$/m$^{3}$ & soil layer (7 - 28 cm) & instantaneous \\
        swvl3  & volumetric soil water & m$^{3}$/m$^{3}$ & soil layer (28 - 100 cm) & instantaneous \\
        swvl4  & volumetric soil water & m$^{3}$/m$^{3}$ & soil layer (100 - 289 cm) & instantaneous \\
        t2m   & 2m temperature & K & 2m & instantaneous \\
        tp    & total precipitation & m & surface & accumulated \\
        u10   & 10 metre U wind component & m/s & 10m & instantaneous \\
        v10   & 10 metre V wind component & m/s & 10m & instantaneous \\

\bottomrule
\end{tabular}
\end{table}

\begin{table}[h!]
\small
 \caption{Details about the processed variables from the ECMWF Integrated Forecast System (IFS) High Resolution (HRES) atmospheric model.} \label{table:HRES}
 \centering
 \setlength\tabcolsep{3pt}
 \begin{tabular}{*{5}l}
    \toprule
        Variable & Long name & Unit & Height & Surface parameters  \\
    \midrule
        e   & total evaporation & m of water equivalent& surface & accumulated\\
        sf  & snowfall & m of water equivalent & surface & accumulated\\
        sp  & surface pressure & Pa & surface & instantaneous \\
        ssr & surface net solar radiation & J/m$^{2}$ & surface & accumulated\\
        str & surface net thermal radiation & J/m$^{2}$ & surface & accumulated\\
        t2m & 2m temperature & K & 2m & instantaneous \\
        tp  & total precipitation & m & surface & accumulated \\

\bottomrule
\end{tabular}
\end{table}

\subsection{CPC data}
\label{sec:Appendix_CPC_data}

Relying solely on the precipitation products from ERA5-Land reanalysis makes the model prune to the biases of the data assimilation which was used to derive the reanalysis. 
Similar to \cite{Google_nature}, we use precipitation estimates as observational input from the National Oceanic and Atmospheric Administration (NOAA), Climate Prediction Center (CPC). The product is called Global Unified Gauge-Based Analysis of Daily Precipitation. The CPC precipitation product is accumulated daily and provided globally at $0.5^\circ \times 0.5^\circ$. To match the resolution of the target river discharge, we mapped CPC data onto the GloFAS domain using nearest point algorithm which preserves the original coarse grid structure but refines the resolution. We did not do any modification for the CPC time zones since it will be considered starting at two days in the past ($t-2$) (see Sec.~\ref{sec:Appendix_implementation_details}).
More details regarding the construction of the daily gauge analysis, the interpolation algorithm, and the gauge algorithm evaluation can be found in \cite{xie2010cpc, xie2007gauge, chen2008assessing}. Operational CPC data can be obtained from \colorh{\url{https://psl.noaa.gov/data/gridded/data.cpc.globalprecip.html}}. 

\subsection{LISFLOOD static features}
\label{sec:Appendix_LISFLOOD_static_data}

River attributes and static maps are crucial to capture the sub-grid variability for the river discharge. For consistency and to make a fair comparison with GloFAS, we used LISFLOOD input static maps \cite{LISFLOOD_Static} similar to the operational GloFAS. This includes $96$ time-invariant variables from $7$ different categories (Table \ref{table:static}). The maps are provided at the same resolution as GloFAS at $3$ arcmin and covering the globe ($90^\circ$N-$60^\circ$S, $180^\circ$W-$180^\circ$E). We excluded the lakes, reservoirs and some static water demand maps. 

In addition, we add the Cartesian coordinates for the points on the WGS-84 ellipsoid to enhance the positional encoding:
\begin{align}
x ={}& (N + H)\cos{\phi}\cos{\lambda},~~~~ y = (N + H)\cos{\phi}\sin{\lambda},~~ z = ((1 - e^2) + H)\sin{\phi}\,,\label{eq:11}
\end{align}
\begin{align}
N ={}& \frac{a}{\sqrt{(1 - e^{2} \sin^2{(\phi)})}},~~~~ e^2 = \frac{a^2 - b^2}{a^2}\,,\label{eq:12}
\end{align}
where $N$ is the radius of curvature in the prime vertical, H is the height from the elevation model, $\phi$ and $\lambda$ are the geographic latitude and longitude, respectively, $a$ and $b$ are the semi-major and semi-minor axes of the ellipse, and $e$ is the eccentricity. We set $a=6,378.137$ km and $b=6,356.752$ km.

The LISFLOOD static maps can be obtained from the Joint Research Centre Data Catalogue \colorh{\url{http://data.europa.eu/89h/68050d73-9c06-499c-a441-dc5053cb0c86}}.

In Sec.~\ref{sec:hydrorivers}, we show experiments using the widely used HydroRIVERS river attributes data \cite{HydroRIVERS, HydroATLAS}.

\begin{table}[h!]
\small
 \caption{The processed LISFLOOD static and parameter maps \protect\cite{LISFLOOD_Static}.} \label{table:static}
 \centering
 \setlength\tabcolsep{3pt}
 \begin{tabular}{l c}
    \toprule     
        Category & \# Static features \\
    \midrule
         catchment morphology and river network & 12 \\
         grid & 2 \\
         land use & 6 \\
         vegetation properties & 45 \\
         soil properties & 14 \\
         water demand & 3 \\
         GloFASv4.0 calibrated parameters & 14 \\

\bottomrule
\end{tabular}
\end{table}

\subsection{Diagnostic river points}
\label{sec:Appendix_river_points}

\begin{sidewaysfigure}
  \centering
  \includegraphics[draft=\draft, width=.99\textwidth]{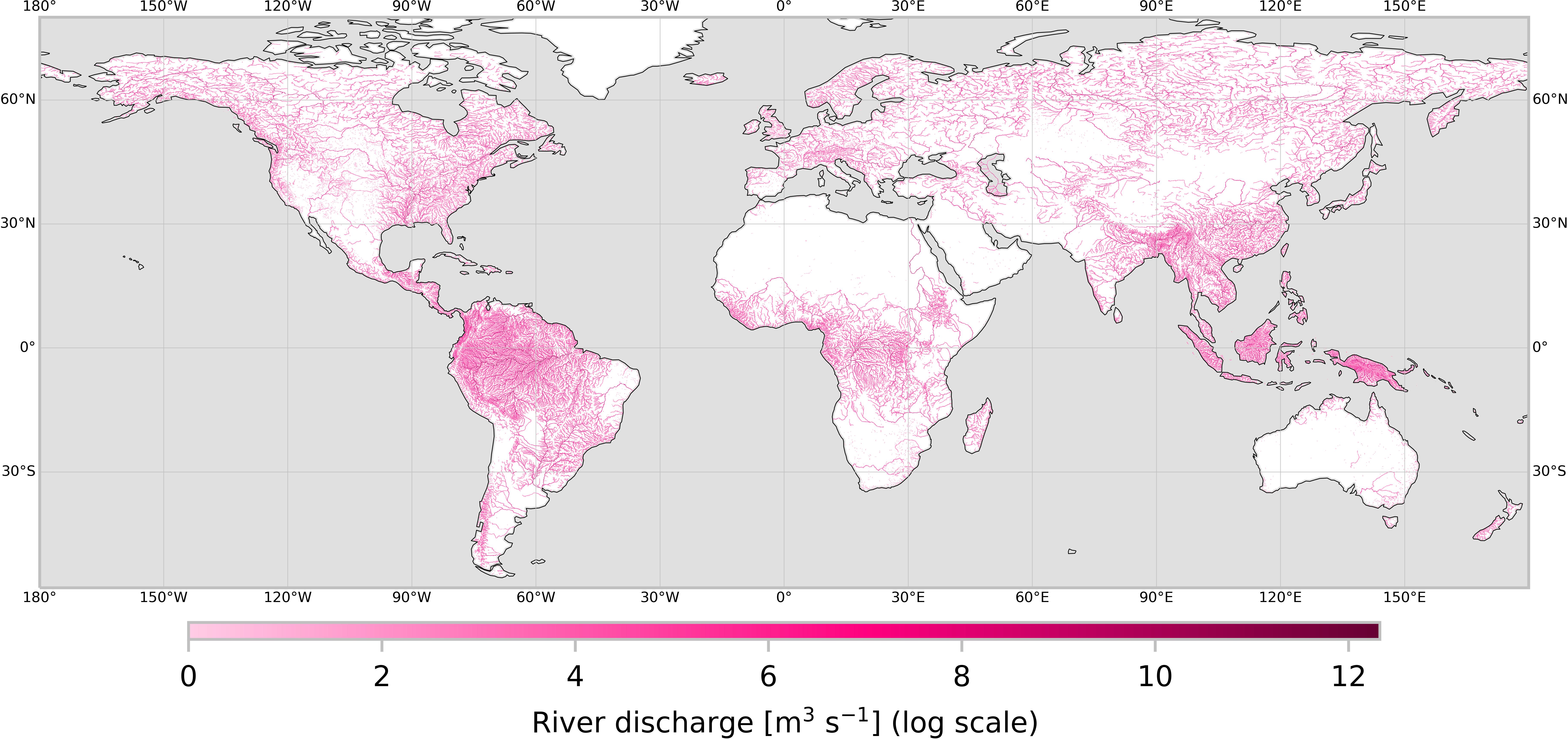}
  \caption{Overview of the selected river points.}\label{fig:river_points}
\end{sidewaysfigure}

The original resolution of GloFAS v.4.0 is $3$ arcmin with an image resolution of $3000 \times 7200$ (21 million pix). In order to run experiments efficiently, we sampled points. For this, we remove all points that are not located on the land surface, i.e., points over ocean or sea. This reduced the points from $21,000,000$ to $6,221,926$ points. We excluded points with median river discharge less than $10$ $m^3 s^{-1}$ since river discharge is more relevant where there is a water flow, i.e., points that are located near to water bodies and not located over desert or glacier regions.
Points which are close to rivers (distance 1 pix to points with discharge > 10 $m^3 s^{-1}$) were not excluded. We also do not exclude points defined as GRDC stations. This reduced the points further to $1,529,667$ diagnostic river points on which we train and test. Figure \ref{fig:river_points} gives an overview of the filtered diagnostic river points used in this study. Note that the trained model can generate river discharge maps at full resolution as can be seen in Sec.~\ref{sec:Appendix_case_study}.  

\newpage

\section{Return periods and flood definition}
\label{sec:Appendix_return_periods}

In hydrology, the return period (also known as the recurrence interval) is a statistical measure that estimates how often a given hydrological event such as a flood, drought, or heavy rainfall is expected to occur on average over a long period. In this study, the return periods refer to the flood frequency. For example, a 2-year flood has a 50\% chance of being exceeded in any given year. The return period ($RP$) is defined as the inverse of the annual exceedance probability (AEP):
\begin{equation}
RP = \frac{1}{AEP} \,. \label{eq:22}
\end{equation}
\noindent

In practice, flood return periods are used to define flood thresholds, i.e., a flood warning is triggered when discharge exceeds the 2-year threshold. 
Note that a high return period i.e., 20 years does not necessarily imply actual flooding in all regions, particularly in highly regulated or flood-resilient areas. A high return period event simply reflects the statistical rarity in streamflow magnitude, and should not be equated with a flood event without additional context (e.g., thresholds, inundation). The return period is used here as a proxy indicator of hydrological extremity, which we call flood severity.
In this study, we adapted the GloFAS approach to define flood thresholds corresponding to selected return periods (or recurrence intervals) of 1.5, 2, 5, 10, 20, 50, 100, 200, and 500 years. These thresholds are derived from the LISFLOOD reanalysis simulations, which are forced with ERA5 meteorological data. The return levels are estimated by fitting a Gumbel extreme value distribution to the annual maxima for the period 1979–2022, using the L-moments method. For the evaluation on GloFAS reanalysis data, we use the pre-defined return periods data from the Copernicus Emergency Management Service (\colorh{\url{https://confluence.ecmwf.int/display/CEMS/Auxiliary+Data}}). 
Fig.~\ref{fig:KOELN} shows the flood thresholds defined by different return periods.
For the GRDC observation dataset, we calculated the return periods at individual stations from the first available observation date to 2022. To allow a fair evaluation of GloFAS reanalysis data on GRDC observation, the return period of GloFAS data is also calculated at GRDC stations but on the local available observation time period. 
Note that while we calculated return period thresholds separately from both the GRDC observations and the GloFAS reanalysis, we did not calculate return periods from the trained ML models reforecast as this would require generating a long reforecast climatology to fit a statistical extreme value distribution. 

\section{Implementation and training details}
\label{sec:Appendix_implementation_details}

The training was done on clusters with NVIDIA A100 80GB and 48GB GPUs.
In Table \ref{table:aifas_parameter}, we highlight the main hyperparameters used for training RiverMamba.

\begin{table}[!h]
\small
 \caption{Implementation details of RiverMamba}\label{table:aifas_parameter}
 \centering
 \setlength\tabcolsep{2pt}
 \begin{tabular}{l c c}
 \toprule
 \multirow{2}{*}{Configuration} & Pre-training & Fine-tuning \\
   & (GloFAS-Reanalysis) & (GRDC) \\
 \midrule
 Optimizer & AdamW & AdamW \\
 Learning rate & 0.0006 & 0.0001 \\
 Minimum learning rate & 0.00009 & 0.00009\\
 Batch size ($B$) & 1 & 1 \\
 Learning rate scheduler & Cosine annealing & Cosine annealing \\
 Weight decay & 0.001 & 0.01 \\
 Training epochs & 60 & 20 \\
 Warmup epochs & 4 & 6 \\
 Gradient clip & 10 & 10 \\
 \midrule
 Input hindcast length ($T$) & 4 & 4 \\
 Lead time ($L$) & 7 & 7 \\
 $\alpha$ for $\hat{u}$ & 0.25 & 0.25 \\
 Number of input points $P$ & 245,954 & 245,954$^*$  \\
 \midrule
 Embedding dimension for GloFAS reanalysis & 48 & 48 \\
 Embedding dimension for ERA5-Land reanalysis & 128 & 128 \\
 Embedding dimension for CPC & 16 & 16 \\
 Number of hindcast layers & 3 & 3 \\
 Hidden dimension in hindcast block ($K$) & 192 & 192 \\
 Depth of hindcast layers & [2, 2, 2] & [2, 2, 2] \\
 Curves in hindcast layers & \{Sweep\_H, Sweep\_V, & \{Sweep\_H, Sweep\_V,  \\
 & Gilbert, Gilbert trans\} & Gilbert, Gilbert trans\} \\
 Grouping size in hindcast block & [(4, 254945), (2, 254945), & [(4, 254945), (2, 254945), \\
 & (1, 254945)] & (1, 254945)] \\
 Dropout in hindcast block & 0.2 & 0.4 \\
 D\_state in hindcast block & 16 & 16 \\
 D\_conv in hindcast block & 4 & 4 \\
 \midrule
 Hidden dimension in forecast block ($K$) & 192+64 & 192+64\\
 Embedding dimension for HRES ($K_{HRES}$) & 64 & 64 \\
 Number of forecast layers & 7 & 7 \\
 Depth of forecast layers & [1, 1, 1, 1, 1, 1, 1] & [1, 1, 1, 1, 1, 1, 1] \\
 Curves in forecast layers & \{Sweep\_H, Sweep\_V, & \{Sweep\_H, Sweep\_V, \\
  & Gilbert, Gilbert trans\} & Gilbert, Gilbert trans\} \\
 Grouping size in forecast block & [(1, 254945)] * 7 & [(1, 254945)] * 7 \\
 Dropout in forecast block & 0.2 & 0.4 \\
 D\_state in forecast block & 16 & 16 \\
 D\_conv in forecast block & 4 & 4 \\
 \midrule
 Hidden dimension in forecasting head ($K_{head}$) & 64 & 64 \\
 Dropout in head & 0.1 & 0.3 \\
\bottomrule
\multicolumn{3}{l}{\footnotesize{$^*$}For GRDC fine-tuning, we only compute the loss where GRDC observations are available.} \\
    
\end{tabular}
\end{table}

\begin{figure}[!b]
  \centering
  \includegraphics[width=.7\textwidth]{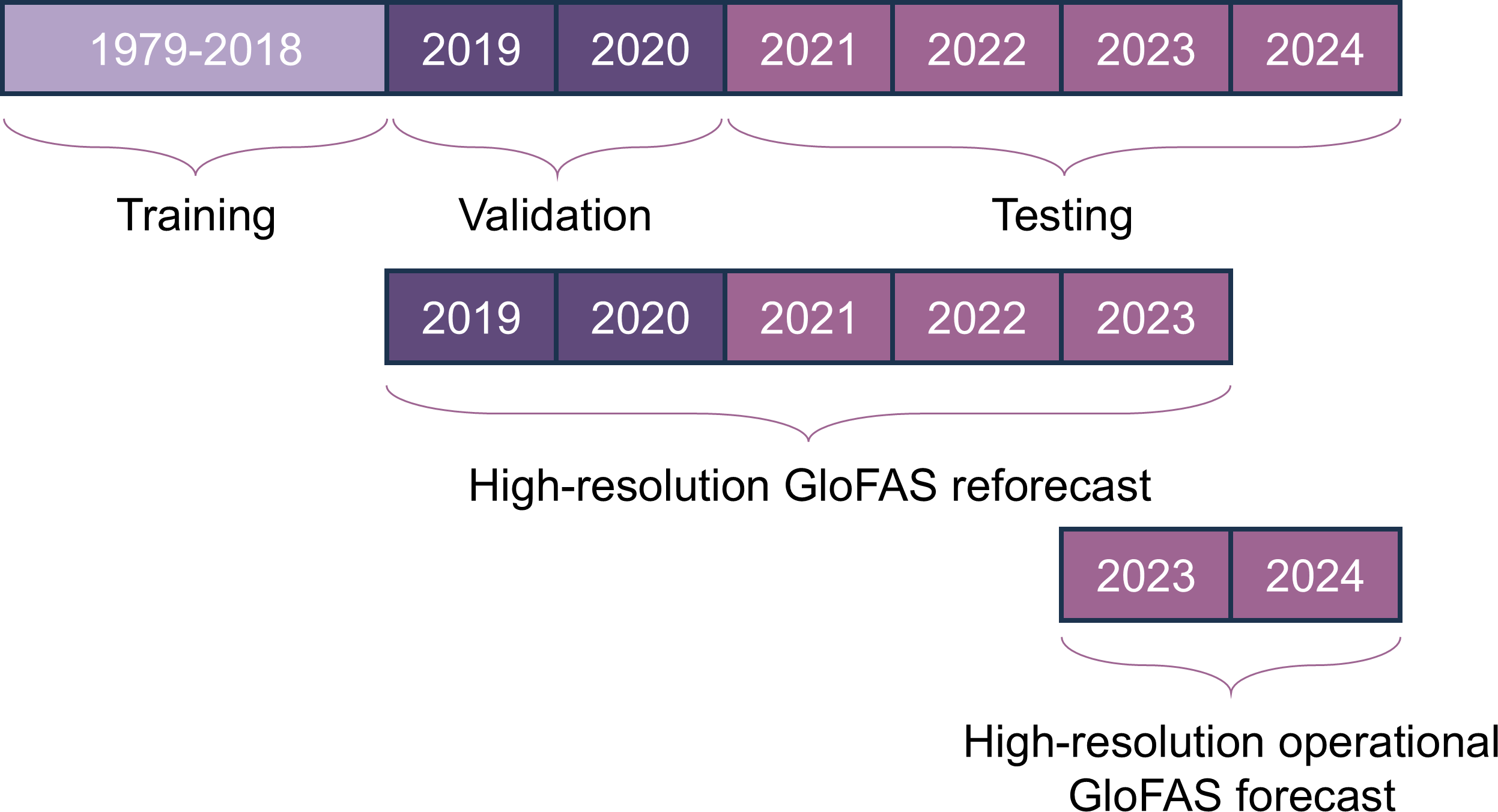}
  \caption{Details about the data splits. 
  }\label{fig:train_val_test_splits}
\end{figure}

To mimic a real operational setting, the initial conditions from CPC data starts at day $t-2$, GLoFAS reanalysis at day $t-1$, and ERA5-Land reanalysis at day $t-1$ in the past.
All input data are normalized based on the computed mean and standard deviation from the training set.
To handle missing data in the reanalysis, we first use the pre-computed statistics to normalize the data. Then, we replace the invalid pixels with zero values.
HRES data is always used for validating and testing. During training, we replace IFS meteorological forcing by ERA5 if they are unavailable, i.e., before 2010.
The training, validation, and testing splits are shown in Fig.~\ref{fig:train_val_test_splits}.

To accelerate training and to fit the data into the memory, we use bfloat16 floating point precision. During inference, we use float32 floating point precision.
Pre-training RiverMamba took about 3 days on 16 GPUs. Finetuning on GRDC data took about 4 hours on 16 GPUs. 

\clearpage

\section{Mamba Block}
\label{sec:Appendix_mamba_block}

\begin{algorithm}[!h]
\caption{Mamba block} \label{alg:block}
\small
\begin{algorithmic}[1]
\REQUIRE{}
\STATE token sequence $\X^l$ : \textcolor{shapecolor}{$(\mathtt{B}, \mathtt{T}, \mathtt{P}, \mathtt{K})$}
\STATE token sequence $\X_{static}$ : \textcolor{shapecolor}{$(\mathtt{B}, \mathtt{P}, \mathtt{K})$}
\STATE curve ID ($S_{block}$) specific to the block $l$
\ENSURE{}
\STATE transformed token sequence $\X^{l+1}$ : \textcolor{shapecolor}{$(\mathtt{B}, \mathtt{T}, \mathtt{P}, \mathtt{K})$}
\\\hrulefill
\STATE \textcolor{gray}{\text{\# serialize the input sequence along the P dimension}}
\STATE $\X^l$ : \textcolor{shapecolor}{$(\mathtt{B}, \mathtt{T}, \mathtt{P}, \mathtt{K})$}, $\X_{static}$ : \textcolor{shapecolor}{$(\mathtt{B}, \mathtt{P}, \mathtt{V_{s}})$} $\leftarrow$ $\mathbf{Serialization}(\X^{l}, S_{block})$, $\mathbf{Serialization}(\X_{static}, S_{block})$
\STATE \textcolor{gray}{\text{\# adaptively normalize the input sequence $\X^{l}$}}
\STATE $\X^{l'}$ : \textcolor{shapecolor}{$(\mathtt{B}, \mathtt{T}, \mathtt{P}, \mathtt{K})$} $\leftarrow$ $\mathbf{LOAN}_{1}(\X^{l}, \X_{static})$
\STATE \textcolor{gray}{\text{\# projection of $\X^{l'}$ into $\mathbf{x}$ and $\mathbf{z}$}, here E is equal to K in our work since we do not expand the dimension}
\STATE $\mathbf{x}$ : \textcolor{shapecolor}{$(\mathtt{B}, \mathtt{T}, \mathtt{P}, \mathtt{E})$}, $\mathbf{z}$ : \textcolor{shapecolor}{$(\mathtt{B}, \mathtt{T}, \mathtt{P}, \mathtt{E})$} $\leftarrow$ $\mathbf{Linear}^\mathbf{xz}(\X^{l'})$
\STATE \textcolor{gray}{\text{\# process with different direction }}
\FOR{$o$ in \{forward, backward\}}
\STATE \textcolor{gray}{\text{\# flip the curve along the spatial dimension P}}
    \IF {$d$ = 'backward'}
        \STATE $\mathbf{x}$ : \textcolor{shapecolor}{$(\mathtt{B}, \mathtt{T}, \mathtt{P}, \mathtt{E})$} $\leftarrow$ $\mathbf{Flip}(\mathbf{x})$
    \ENDIF
\STATE \textcolor{gray}{\text{\# flatten the curve along the temporal dimension 'spatial-first'}}
\STATE $\mathbf{x'}$ : \textcolor{shapecolor}{$(\mathtt{B}, \mathtt{(T \times P)}, \mathtt{E})$} $\leftarrow$ $\mathbf{Flatten}(\mathbf{x})$
\STATE \textcolor{gray}{\text{\# selective state space model, here N is the D\_state}}
\STATE $\mathbf{x}'_o$ : \textcolor{shapecolor}{$(\mathtt{B}, \mathtt{(T \times P)}, \mathtt{E})$} $\leftarrow$ $\mathbf{SiLU}(\mathbf{Conv1d}_o(\mathbf{x'}))$
\STATE $\mathbf{B}_o$ : \textcolor{shapecolor}{$(\mathtt{B}, \mathtt{(T \times P)}, \mathtt{N})$}, $\mathbf{C}_o$ : \textcolor{shapecolor}{$(\mathtt{B}, \mathtt{(T \times P)}, \mathtt{N})$} $\leftarrow$ $\mathbf{Linear}_o^{\mathbf{B}}(\mathbf{x}'_o)$, $\mathbf{Linear}^{\mathbf{C}}_o(\mathbf{x}'_o)$
\STATE \textcolor{gray}{\text{\# initialize $D_o$ with ones}}
\STATE $\mathbf{D}_o$ : \textcolor{shapecolor}{$(\mathtt{E})$} $\leftarrow$ $\mathbf{Parameter}$ Ones : \textcolor{shapecolor}{$(\mathtt{E})$}
\STATE \textcolor{gray}{\text{\# softplus ensures positive $\mathbf{\Delta}_o$ }}
\STATE $\mathbf{\Delta}_o$ : \textcolor{shapecolor}{$(\mathtt{B}, \mathtt{(T \times P)}, \mathtt{E})$} $\leftarrow$ $\log(1 + \exp(\mathbf{Linear}_o^{\mathbf{\Delta}}(\mathbf{x}'_o) + \mathbf{Parameter}_o^{\mathbf{\Delta}}))$
\STATE \textcolor{gray}{\text{\# shape of $\mathbf{Parameter}_o^{\mathbf{A}}$ is \textcolor{shapecolor}{$(\mathtt{E}, \mathtt{N})$} }}
\STATE $\overline{\mathbf{A}_o}$ : \textcolor{shapecolor}{$(\mathtt{B}, \mathtt{(T \times P)}, \mathtt{E}, \mathtt{N})$} $\leftarrow$ $\mathbf{\Delta}_o \bigotimes \mathbf{Parameter}_o^{\mathbf{A}}$ 
\STATE $\overline{\mathbf{B}_o}$ : \textcolor{shapecolor}{$(\mathtt{B}, \mathtt{(T \times P)}, \mathtt{E}, \mathtt{N})$} $\leftarrow$ $\mathbf{\Delta}_o \bigotimes \mathbf{B}_o$
\STATE \textcolor{gray}{\text{\# initialize $h_o$ and $\mathbf{y}_o$ with zeros }}
\STATE $h_o$ : \textcolor{shapecolor}{$(\mathtt{B}, \mathtt{E}, \mathtt{N})$} $\leftarrow$ Zeros : \textcolor{shapecolor}{$(\mathtt{B}, \mathtt{E}, \mathtt{N})$}
\STATE $\mathbf{y}_o$ : \textcolor{shapecolor}{$(\mathtt{B}, \mathtt{(T \times P)}, \mathtt{E})$} $\leftarrow$ Zeros : \textcolor{shapecolor}{$(\mathtt{B}, \mathtt{(T \times P)}, \mathtt{E})$}
\STATE \textcolor{gray}{\text{\# SSM recurrent }}
\FOR{$i$ in \{0, ..., $L$-1\}}
\STATE $h_o$ = $\overline{\mathbf{A}_o}[:,i,:,:] \bigodot h_o + \overline{\mathbf{B}_o}[:,i,:,:] \bigodot \mathbf{x}_o'[:,i,:,\textcolor{shapecolor}{\mathtt{None}}] $
\STATE $\mathbf{y}_o[:,i,:]$ = $h_o \bigotimes \mathbf{C}_o[:,i,:]$ + $\mathbf{D}_o[\textcolor{shapecolor}{\mathtt{None}}, :] \bigodot \mathbf{x}_o'[:,i,:] $
\ENDFOR
\STATE \textcolor{gray}{\text{\# reshape $\mathtt{(T \times P)}$ to (T, P)}}
 \STATE $\mathbf{y_o}$ : \textcolor{shapecolor}{$(\mathtt{B}, \mathtt{T}, \mathtt{P}, \mathtt{E})$} $\leftarrow$ $\mathbf{Reshape}(\mathbf{y_o})$
\STATE \textcolor{gray}{\text{\# flip the curve along the spatial dimension P}}
    \IF {$o$ = 'backward'}
        \STATE $\mathbf{y_o}$ : \textcolor{shapecolor}{$(\mathtt{B}, \mathtt{T}, \mathtt{P}, \mathtt{E})$} $\leftarrow$ $\mathbf{Flip}(\mathbf{y_o})$
    \ENDIF
\ENDFOR
\STATE \textcolor{gray}{\text{\# get gated $\mathbf{y}$ }}
\STATE $\mathbf{y}_{forward}'$ : \textcolor{shapecolor}{$(\mathtt{B}, \mathtt{T}, \mathtt{P}, \mathtt{E})$}, $\mathbf{y}_{backward}'$ : \textcolor{shapecolor}{$(\mathtt{B}, \mathtt{T}, \mathtt{P}, \mathtt{E})$} $\leftarrow$ $\mathbf{y}_{forward} \bigodot \mathbf{SiLU}(\mathbf{z})$, $\mathbf{y}_{backward} \bigodot \mathbf{SiLU}(\mathbf{z}) $

\STATE \textcolor{gray}{\text{\# post normalization and residual connection }}
\STATE $\mathbf{X}^{l+1'}$ : \textcolor{shapecolor}{$(\mathtt{B}, \mathtt{(T \times P)}, \mathtt{K})$} $\leftarrow$ $\mathbf{Linear}^\mathbf{X}(\mathbf{LayerNorm}((\mathbf{y}_{forward}' + \mathbf{y}_{backward}')/2)) + \mathbf{X}^{l}$
\STATE \textcolor{gray}{\text{\# adaptively normalize the output sequence $\X^{l+1}$}}
\STATE $\X^{l+1}$ : \textcolor{shapecolor}{$(\mathtt{B}, \mathtt{T}, \mathtt{P}, \mathtt{K})$} $\leftarrow$ $\mathbf{LOAN}_{2}(\X^{l+1'}, \X_{static})$
\STATE \textcolor{gray}{\text{\# feed-forward layer and residual connection}}
\STATE $\X^{l+1}$ : \textcolor{shapecolor}{$(\mathtt{B}, \mathtt{T}, \mathtt{P}, \mathtt{K})$} $\leftarrow$ $\mathbf{MLP}(\X^{l+1}) + \X^{l+1'}$

\STATE \textcolor{gray}{\text{\# resort the input sequence along the $P$ dimension}}
\STATE $\X^{l+1}$ : \textcolor{shapecolor}{$(\mathtt{B}, \mathtt{T}, \mathtt{P}, \mathtt{K})$} $\leftarrow$ $\mathbf{Resort}(\X^{l+1}, S_{block})$
\STATE Return: $\X^{l+1}$ 
\end{algorithmic}
\end{algorithm}

\clearpage

\section{Evaluation metrics}
\label{sec:Appendix_evaluation_metrics}

To assess model performance, we used 8 metrics that are commonly used for hydrological modeling and flood forecasting evaluation \cite{Gauch_2023}. This includes MAE (Mean Absolute Error), RMSE (Root Mean Square Error), R (Pearson Correlation Coefficient), R2 (Coefficient of Determination), KGE (Kling–Gupta Efficiency), Precision, Recall and F1 score. Below are the details about the individual metrics:

\textbf{Mean Absolute Error (MAE)} represents the average of the absolute differences between the predicted and observed values. It provides a straightforward measure of model accuracy. MAE is less sensitive to outliers than RMSE:

\begin{equation}\label{eq:MAE}
\mathrm{MAE} = \frac{1}{P} \sum_{p=1}^{P} \left| \X_{p}^{\text{obs}} - \X_{p}^{\text{pred}} \right| \,,
\end{equation}
where $\X_{p}^{\mathrm{obs}}$ is the observed river discharge at point $p$, $\X_{p}^{\mathrm{pred}}$ is the predicted river discharge, and $P$ is the total number of points.

\textbf{Root Mean Square Error (RMSE)} measures the square root of the average squared differences between predicted and observed values. It penalizes large errors more heavily than MAE:

\begin{equation}
\mathrm{RMSE} = \sqrt{ \frac{1}{P} \sum_{p=1}^{P} (X_{p}^{\text{obs}} - X_{p}^{\text{pred}})^2 } \,.
\end{equation}

\textbf{Pearson Correlation Coefficient (R)} measures the linear relationship between observed and predicted values, ranging from –1 (perfect negative correlation) to +1 (perfect positive correlation):

\begin{equation}
R = \frac{\sum_{p=1}^{P} (X_p^{\text{obs}} - \bar{X}^{\text{obs}})(X_p^{\text{pred}} - \bar{X}^{\text{pred}})}{\sqrt{\sum_{p=1}^{P} (X_p^{\text{obs}} - \bar{X}^{\text{obs}})^2} \sqrt{\sum_{p=1}^{P} (X_p^{\text{pred}} - \bar{X}^{\text{pred}})^2}} \,,
\end{equation}
where $\bar{X}^{\mathrm{pred}}$ is the mean of predicted river discharge, and $\bar{X}^{\mathrm{obs}}$ is the mean of observed river discharge.  

\textbf{Coefficient of Determination (R2)} evaluates the predictive power of a model relative to the observed mean. It has the same meaning as Nash–Sutcliffe Efficiency (NSE) which is commonly used in hydrology. Values closer to 1 indicate better performance, while values below 0 suggest that the model performs worse than using the observed mean:


\begin{equation}
\mathrm{R2} = 1 - \frac{\sum_{p=1}^{P} (X_p^{\text{obs}} - X_p^{\text{pred}})^2}{\sum_{p=1}^{P} (X_p^{\text{obs}} - \bar{X}^{\text{obs}})^2} \,.
\end{equation}


\textbf{Kling–Gupta Efficiency (KGE)} is a composite metric that combines correlation, bias, and variability. It addresses some weaknesses of NSE by ensuring balance across multiple aspects of model performance. Like NSE, values near 1 indicate good performance, while values below 0 indicate performance worse than the observed mean:

\begin{equation}\label{eq:KGE}
\mathrm{KGE} = 1 - \sqrt{ (r - 1)^2 + (\beta - 1)^2 + (\gamma - 1)^2 } \,,
\end{equation}
\vspace{0.3cm}
\begin{equation}
\beta = \frac{\bar{X}^{\text{pred}}}{\bar{X}^{\text{obs}}}, \quad
\gamma = \frac{\text{CV}^{\text{pred}}}{\text{CV}^{\text{obs}}}, \quad
r = \text{Pearson correlation coefficient} \,.
\end{equation}
where $r$ is the Pearson correlation between observed and simulated, $\beta$ is the bias ratio, $\gamma$ is the variability ratio, and $\mathrm{CV}^{\text{obs}} = \sigma^{\text{obs}} / \bar{X}^{\text{obs}}$ and $\mathrm{CV}^{\text{pred}} = \sigma^{\text{pred}} / \bar{X}^{\text{pred}}$ are the coefficients of variation, where $\sigma^{\text{obs}}$ and $\sigma^{\text{pred}}$ are the standard deviations of the observed and predicted river discharge, respectively. 

\textbf{Precision} is the proportion of correctly identified positive cases (i.e., flood events) among all predicted positives. High precision indicates a low false-positive rate:
\begin{equation}\label{eq:prec}
\mathrm{Precision} = \frac{\mathrm{TP}}{\mathrm{TP + FP}} \,,
\end{equation}
where TP is the number of true positives and FP is the number of false positives. 

\textbf{Recall} is the proportion of correctly identified positives among all actual positives. High recall indicates a low false-negative rate:
\begin{equation}
\mathrm{Recall} = \frac{\mathrm{TP}}{\mathrm{TP + FN}} \,,
\end{equation}
where FN is the number of false negatives.

\textbf{F1-score} is the harmonic mean of precision and recall, particularly useful in imbalanced classification tasks (i.e., flood detection where flood events are rare):

\begin{equation}\label{eq:F1}
\mathrm{F1} = 2 \cdot \frac{\mathrm{Precision} \cdot \mathrm{Recall}}{\mathrm{Precision} + \mathrm{Recall}} \,.
\end{equation}

In this study, the metrics \eqref{eq:MAE}-\eqref{eq:KGE} are used to evaluate the agreement between observed and forecast discharge time series. During the evaluation, these metrics are calculated on the time series at single grid points and then averaged over all the grid points. The metrics \eqref{eq:prec}-\eqref{eq:F1} are applied to assess the model’s ability to detect flood events at different return periods. For example, on a day classified as exceeding the 2-year return period threshold, a correct prediction of discharge above this threshold is considered a true positive. 
If not otherwise specified, we report F1-score averaged over 1.5-20 year return periods and all 3366 points.


\newpage

\section{Ablation studies}
\label{sec:Appendix_ablation}

For the ablation studies, we conducted experiments over the European domain ($60^\circ$N $30^\circ$S, $-10^\circ$W $40^\circ$E) which has 82,804 points from the filtered diagnostic river points defined in Sec.~\ref{sec:Appendix_river_points} and includes $675$ GRDC stations for evaluation.

\subsection{Mamba vs.\ Transformer}
\label{sec:Appendix_ablation_ssm_transformer}

\begin{figure}[!h]
  \centering
  \includegraphics[width=.98\textwidth]{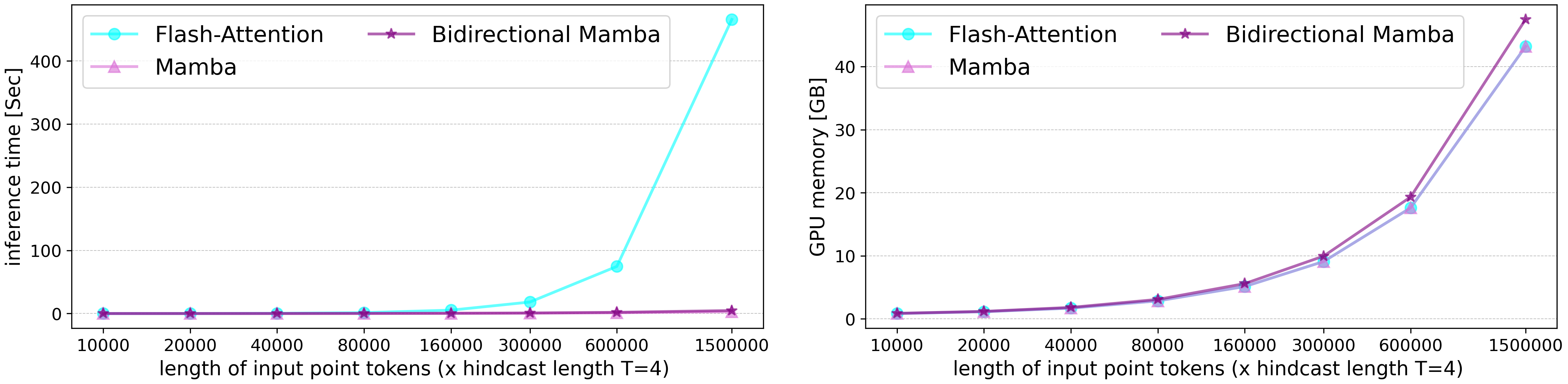}
  \caption{Comparison between Mamba and Transformer-based backbones.}\label{fig:ablation_transformer_vs_ssm}
\end{figure}

In Fig.~\ref{fig:ablation_transformer_vs_ssm}, we compare the model using Mamba blocks to a variant using Transformer blocks with Flash-Attention \cite{dao2022flashattention, dao2023flashattention2}. Both Mamba and Flash-Attention are efficient compared to a typical self-attention. However, Mamba scales better with the number of input tokens (Fig.~\ref{fig:ablation_transformer_vs_ssm} (left)), important for global modeling. The Transformer-based approach becomes computationally infeasible regarding the runtime for a larger number of input points. Both approaches have similar memory consumption which scales linearly with the sequence length. Using the bidirectional Mamba block increases the memory consumption slightly (Fig.~\ref{fig:ablation_transformer_vs_ssm} (right)).

Table \ref{table:ablation_ssm_transformer_training_time} shows a comparison between Mamba and Flash-Attention regarding training time with different configurations. Since Flash-Attention does not scale with global data, we split the space-filling curve and do a local modeling (see Fig.\ref{fig:local_vs_global_mamba} (b)). This requires rearranging the curves at each block, which becomes the main bottleneck for Flash-Attention compared to Mamba.

\begin{table}[!h]
 \caption{Training time on the reanalysis data for 1 epoch (1,529,667 points).}\label{table:ablation_ssm_transformer_training_time}
 \centering
 \setlength\tabcolsep{6pt}
 \begin{tabular}{l *{3}l}
 \toprule
 Model & Time (min) & GPUs & CPUs \\
\midrule
Flash-Attention & $\sim 145 $ & $16 \times \text{A}100$ & $ 64 \times 4 $ \\
Mamba & $\sim 82$ & $16 \times \text{A}100$ & $ 64 \times 4 $ \\
\bottomrule
\end{tabular}
\end{table}

Table \ref{table:ablation_backbone} shows the results for RiverMamba with different backbones. Flash-Attention and Mamba2 \cite{mamba2} achieve slightly lower performance compared to Mamba. 

\begin{table}[h!]
  \caption{Ablation studies for the RiverMamba backbone on the validation set over Europe.}
  \label{table:ablation_backbone}
  \centering
  \tabcolsep=4.0pt\relax
  \setlength\extrarowheight{0pt}
  \begin{tabular}{*{3}l}
    \toprule
    Backbone & \# Params &  KGE | F1 ($\uparrow$) \\
    \cmidrule[\heavyrulewidth]{1-3}
    Flash-Attention & 5.03 M & \colorm{0.9161} | \colorf{0.2804} \\
    \bestc Mamba & \bestc 4.38 M & \bestc \textbf{\colorm{0.9205}} | \textbf{\colorf{0.2875}} \\
    Mamba2 & 4.02 M & \colorm{0.9169} | \colorf{0.2793} \\
    \bottomrule
  \end{tabular}
\end{table}

\subsection{Feature importance}
\label{sec:Appendix_ablation_feature_importance}

In Table \ref{table:ablation_feature_importance}, we study the performance of RiverMamba with different input features. All input features have an impact on the performance. Removing the observational CPC data only slightly reduces the results for GloFAS reanalysis, but the decrease is larger for observational GRDC.
Removing GloFAS reanalysis and ERA5-Land initial conditions reduces the performance as well, but GloFAS reanalysis is more important. 
For GloFAS reanalysis (fourth row), we only drop GloFAS from the input and keep the data as a target to train the model. During inference, we still use the last step of GloFAS reanalysis and add it to $\Delta \X_{dis24}$ to generate the discharge. This ensures consistency within the table and isolates the impact of input GloFAS reanalysis on the model. RiverMamba still works without taking GloFAS as input. This highlights that the model is more than a post-processor of discharge data and can in fact be used as a backbone for hydrological modeling. Note that if we want to drop GloFAS completely, we need to change the objective function, i.e., by predicting the absolute value of river discharge or the change of discharge w.r.t. climatology.
Finally, removing the meteorological forcing forecasts HRES has the biggest impact since weather forecasting is an important source of information. The best performance is achieved when we use all input variables.

\begin{table}[h!]
  \caption{Ablation studies for feature importance on the validation set over Europe.}
  \label{table:ablation_feature_importance}
  \centering
  \tabcolsep=4.0pt\relax
  \setlength\extrarowheight{0pt}
  \begin{tabular}{*{7}c}
  
    \toprule
    
    Static & CPC & ERA5 & GloFAS-reanalysis & HRES & KGE | F1 ($\uparrow$) & KGE | F1 ($\uparrow$) \\

    {\small(LISFLOOD)} &  &  &   &  & {\small(Reanalysis)} & {\small(GRDC Obs)} \\
      
    \cmidrule[\heavyrulewidth]{1-7}
    
    \xmark & \cmark & \cmark & \cmark & \cmark & \colorm{0.9091} | \colorf{0.2505} & \colorm{0.7633} | \colorf{0.1118} \\
        
    \cmark & \xmark & \cmark & \cmark & \cmark & \colorm{0.9151} | \colorf{0.2842} & \colorm{0.7681} | \colorf{0.1102} \\
    
    \cmark & \cmark & \xmark & \cmark & \cmark & \colorm{0.9077} | \colorf{0.2521} & \colorm{0.7731} | \colorf{0.1110} \\

    \cmark & \cmark & \cmark & \xmark & \cmark & \colorm{0.9060} | \colorf{0.2450} & \colorm{0.7521} | \colorf{0.1226} \\
    

    \cmark & \cmark & \cmark & \cmark & \xmark & \colorm{0.7972} | \colorf{0.1276} & \colorm{0.6757} | \colorf{0.0640} \\

    \cmark \bestc & \cmark \bestc & \cmark \bestc & \cmark \bestc & \cmark \bestc & \bestc \textbf{\colorm{0.9205} | \colorf{0.2875}} & \bestc \textbf{\colorm{0.7838} | \colorf{0.1335}} \\
        
    \bottomrule
  \end{tabular}
\end{table}


\subsection{Pretraining on reanalysis}
\label{sec:Appendix_ablation_pretraining}

In Table \ref{table:ablation_pretraining}, we show the value of pretraining on GloFAS reanalysis data for the GRDC prediction. From our experiments, we can see a clear benefit of training on river discharge reanalysis before training the model on GRDC observations.

\begin{table}[h!]
  \caption{Ablation studies for pretraining on GloFAS reanalysis on the validation set over Europe.} \label{table:ablation_pretraining}
  \centering
  \tabcolsep=4.0pt\relax
  \setlength\extrarowheight{0pt}
  \begin{tabular}{*{3}c}
  
    \toprule
    
      Pretrained on GloFAS reanalysis & KGE | F1 ($\uparrow$) \\
      
      \cmidrule[\heavyrulewidth]{1-2}

      \xmark & \colorm{0.7406} | \colorf{0.0882}  \\
      \bestc \cmark & \bestc \textbf{\colorm{0.7838} | \colorf{0.1335}} \\

    \bottomrule
  
  \end{tabular}
\end{table}

\subsection{Space-filling curves}
\label{sec:Appendix_ablation_curves}

In Table \ref{table:ablation_space_filling_curves_1}, we investigate the impact of various serialization patterns for RiverMamba. 
Our experiments show that sweep curves perform better than the other curves. Iterating between sweep and Gilbert curves (fourth column) improves the F1-score further. Iterating over all curves improves F1-score, but decreases KGE. For simplicity, we use thus the combination of sweep and Gilbert curves for our experiments.   

\begin{table}[h!]
  \caption{Ablation studies for space-filling curves on the validation set over Europe. The columns indicate the serialization patterns: G for Gilbert, Z for Zigzag, S for Sweep, Shuffle represents shuffling the order inside the hindcast layers. S and Z curves use both direction H and V and G uses both regular and trans Gilbert versions.} \label{table:ablation_space_filling_curves_1}
  \centering
  \tabcolsep=2.75pt\relax
  \setlength\extrarowheight{0pt}
  \begin{tabular}{l*{5}c}
  
    \toprule
    
      Curve type & G & Z & S & \bestc S + G & S + G + Z  \\
      
      \cmidrule[\heavyrulewidth]{1-6}

      KGE | F1 ($\uparrow$) & \colorm{0.9156} | \colorf{0.2733} & \colorm{0.9156} | \colorf{0.2719} & \textbf{\colorm{0.9205}} | \colorf{0.2826} & \bestc \textbf{\colorm{0.9205}} | \colorf{0.2875} & \colorm{0.9163} | \textbf{\colorf{0.2962}}  \\
      
    \bottomrule
  
  \end{tabular}
\end{table}

In Table \ref{table:ablation_space_filling_curves_3}, we remove the spatial modeling in RiverMamba completely and do the scanning only along the temporal dimension (first row). The results show the importance of spatiotemporal modeling.

\begin{table}[h!]
  \caption{Ablation regarding the spatiotemporal modeling on the validation set over Europe.} \label{table:ablation_space_filling_curves_3}
  \centering
  \tabcolsep=4.0pt\relax
  \setlength\extrarowheight{0pt}
  \begin{tabular}{*{3}c}
  
    \toprule
    
      Temporal modeling & Spatiotemporal modeling & KGE | F1 ($\uparrow$) \\
      
      \cmidrule[\heavyrulewidth]{1-3}
      \cmark &  \xmark & \colorm{0.8726} | \colorf{0.1952} \\
      \bestc \xmark & \bestc \cmark & \bestc \textbf{\colorm{0.9205}} | \textbf{\colorf{0.2875}} \\
      
    \bottomrule
  
  \end{tabular}
\end{table}

The design of the scanning also plays a role. From Table \ref{table:ablation_space_filling_curves_2} (a), we found that sequential scanning along the spatial dimension ($P$) works better. This is represented as scanning from $P$ to $T$ (P$\rightarrow$T). In other words, the points are connected over time by scanning at time step $t$ and continuing the scan at $t+1$. In the second case (T$\rightarrow$P), each point will be scanned along the time dimension ($T$) and then connected to the next point along the spatial dimension $P$.

\begin{figure}[!h]
  \centering
  \includegraphics[width=.98\textwidth]{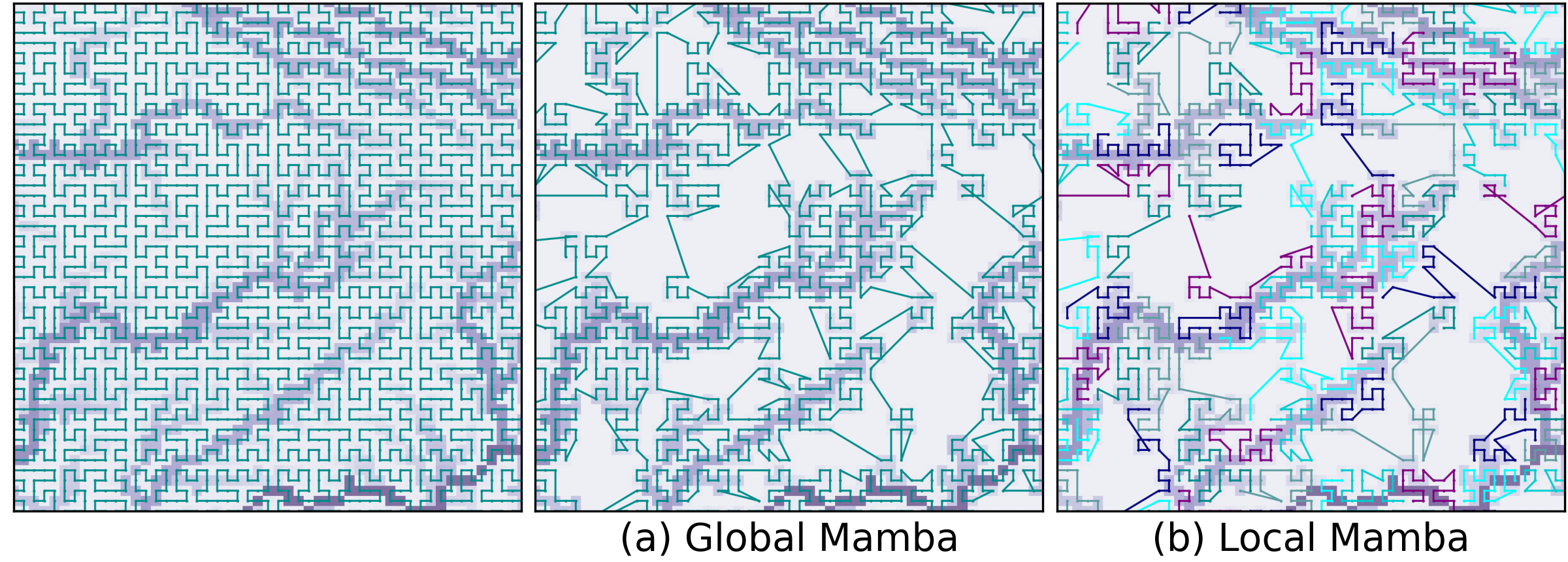}
  \caption{Illustration of the difference between global curve (middle) and grouped local curves (right). The left image shows a Gilbert space-filing curve for all points. In our experiments, the global curve is used (middle). For the experiments with Flash-Attention, we use the local curves (right).}\label{fig:local_vs_global_mamba}
\end{figure}

In Table \ref{table:ablation_space_filling_curves_2} (b), we split the curve into local curves similar to PointTransformer \cite{Wu_2024_CVPR} (Fig.~\ref{fig:local_vs_global_mamba} (b)). Using a larger receptive field gives the model more capability to extract up- and downstream features and to model adjacent catchments. In addition, local modeling needs more computations along the network i.e., sorting, resorting and padding. Finally, bidirectional Mamba (Table \ref{table:ablation_space_filling_curves_2} (c)) collects information about the streamflow from both side of the curve thus covering the whole domain and achieving a better performance than unidirectional Mamba.

\begin{table}[t!]
  \caption{Ablation studies for scan patterns on the validation set over Europe.}
  \label{table:ablation_space_filling_curves_2}
  \centering
  \tabcolsep=2.5pt\relax
  \setlength\extrarowheight{0pt}
  \begin{tabular}{*{11}c}
      
      \multicolumn{3}{l}{(a) Curve order} & & \multicolumn{3}{l}{(b) Curve type} & & \multicolumn{3}{l}{(c) Bidirectional Curve}\\
      
      \cmidrule[\heavyrulewidth]{1-3} \cmidrule[\heavyrulewidth]{5-7} \cmidrule[\heavyrulewidth]{9-11}
        
      T$\rightarrow$P & P$\rightarrow$T  & KGE | F1 ($\uparrow$) & & Local & Global & KGE | F1 ($\uparrow$) & & SSM & Bi-SSM & KGE | F1 ($\uparrow$) \\
      
      \cmidrule[\heavyrulewidth]{1-3} \cmidrule[\heavyrulewidth]{5-7} \cmidrule[\heavyrulewidth]{9-11}
      
      \cmark & \xmark & \colorm{0.9153} | \colorf{0.2807} & & \cmark & \xmark & \colorm{0.9164} | \textbf{\colorf{0.2893}} & & \cmark & \xmark & \colorm{0.9113} | \colorf{0.2482} \\
      \xmark \bestc & \cmark \bestc  & \bestc \textbf{\colorm{0.9205} | \colorf{0.2875}} & & \xmark \bestc & \cmark \bestc & \bestc \textbf{\colorm{0.9205}} | \colorf{0.2875} & & \xmark \bestc &  \cmark \bestc & \bestc \textbf{\colorm{0.9205} | \colorf{0.2875}} \\

      \cmidrule[\heavyrulewidth]{1-3} \cmidrule[\heavyrulewidth]{5-7} \cmidrule[\heavyrulewidth]{9-11}
      
  \end{tabular}
\end{table}

\begin{figure}[!h]
  \centering
  \includegraphics[draft=\draft, width=.4\textwidth]{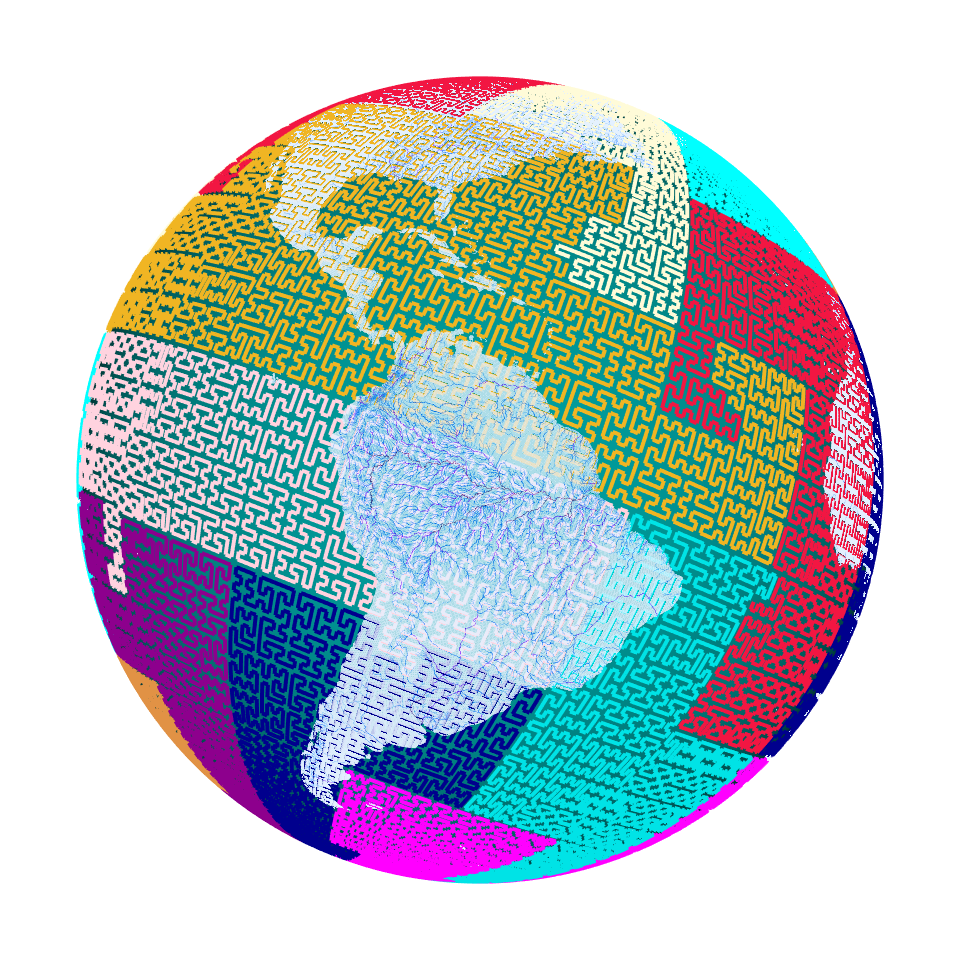}
  \caption{A simplified view of splitting along Gilbert space-filling curve. 
  }\label{fig:split_curves_on_sphere}
\end{figure}

For training and inference on the global dataset, it is impractical to fit all the input points ($\sim 6$ million points) into the memory. For this, we first define a Gilbert space-filling curve on the globe and then we split the curve into smaller curves along the space-filling curve, i.e., we split the curve into sequences with $\sim311$K points for each.
A simplified version of the splits is shown in Fig.~\ref{fig:split_curves_on_sphere}.

\subsection{Weighting in the objective function}
\label{sec:Appendix_ablation_objective_function}

In this experiment, we study the effect of weighting floods not just by their return period but also by augmenting it with an additive flood offset of $1$. To this end, we trained a model with weighted flood events by their return periods + flood offset of $1$. The F1 results are shown below in Table \ref{table:ablation_weighting_objective_function} for the reanalysis dataset and different return periods. Adding an offset of $1$ does not improve the results. 

\begin{table}[h!]
  \caption{Ablation study on the validation set over Europe regarding the weighting in the objective function. Shown is F1-score for reanalysis data across different return periods.} \label{table:ablation_weighting_objective_function}
  \centering
  \tabcolsep=4.5pt\relax
  \setlength\extrarowheight{0pt}
  \begin{tabular}{l*{5}c}
  
    \toprule
      Return period & 1.5 & 2.0 & 5.0 & 10.0 & 20.0 \\
      \midrule
    Validation (2019-2020) & & & & &\\
        \midrule
      W/ offset & \colorf{0.4820} & \colorf{0.3760} & \colorf{0.2358} & \colorf{0.1790} & \colorf{0.1181} \\
      W/o offset & \bestc \textbf{\colorf{0.4870}} & \bestc \textbf{\colorf{0.3767}} & \bestc \textbf{\colorf{0.2516}} & \bestc \textbf{\colorf{0.2015}} & \bestc \textbf{\colorf{0.1208}} \\
        \midrule
        Testing (2021-2024) & & & & &\\
        \midrule         
      W/ offset & \colorf{0.6114} & \bestc \bestc \textbf{\colorf{0.5080}} & \bestc \textbf{\colorf{0.3125}} & \bestc \textbf{\colorf{0.2486}} & \colorf{0.1656} \\
      W/o offset & \bestc \textbf{\colorf{0.6122}} & \colorf{0.5072} & \colorf{0.3031} & \colorf{0.2434} & \bestc \textbf{\colorf{0.1669}} \\
    \bottomrule
  \end{tabular}
\end{table}

\subsection{Activation function in LOAN}
\label{sec:Appendix_ablation_gelu_loan}

We conducted an additional experiment where we replaced the activation function in the LOAN layer by ReLU activation. GELU \cite{GELU} avoids the dying ReLU problem and improves optimization. As can be seen from Table \ref{table:ablation_gelu_loan}, GELU performs slightly better.

\begin{table}[h!]
  \caption{Ablation study on the validation set over Europe regarding the activation function in LOAN layers.} \label{table:ablation_gelu_loan}
  \centering
  \tabcolsep=4.5pt\relax
  \setlength\extrarowheight{0pt}
  \begin{tabular}{l*{3}c}

    \toprule
      Activation function & ReLU & \bestc GELU \\
      \midrule
      KGE | F1 ($\uparrow$) & \colorm{0.9143} | \colorf{0.2833} & \bestc \textbf{\colorm{0.9205} | \colorf{0.2875}} \\
    \bottomrule
  \end{tabular}
\end{table}

\newpage

\section{Computational time}
\label{sec:Appendix_computational_time}

Neither Google \cite{Google_nature} nor GloFAS \cite{GloFAS_Forecast} provided the compute time for the operational forecast. The inference time for RiverMamba is reported in supp. Fig. \ref{fig:ablation_transformer_vs_ssm}. In Table \ref{table:computational_time}, we report the inference time (seconds) w.r.t. the number of input points for our model with $4$ days as a hindcast (first row), a trained version of Google's LSTM with $4$ days as a hindcast (second row), and a trained Google's LSTM version as in \cite{Google_nature} with one year hincast (third row). We use one A100 GPU for all runs. All machine learning approaches are very fast. We expect that GloFAS is by several magnitudes slower, which is a practical advantage of machine learning approaches for this task.

\begin{table}[h!]
  \caption{Inference time in seconds.} \label{table:computational_time}
  \centering
  \tabcolsep=4.5pt\relax
  \setlength\extrarowheight{0pt}
  \begin{tabular}{l*{8}l}
  
    \toprule
      Model & 10K & 20K & 40K & 80K & 160K & 300K & 600K & 1500K \\
    \midrule
    RiverMamba (4 days hindcast) & 0.026 & 0.044 & 0.086 & 0.190 & 0.423 & 0.874 & 1.914 & 4.739 \\
    LSTM (4 days hindcast) & 0.005 & 0.009 & 0.015 & 0.027 & 0.053 & 0.098 & - & - \\
    LSTM (one year hindcast) & 0.069 & - & - & - & - & - & - & - \\
    \bottomrule
  \end{tabular}
\end{table}

\newpage

\section{Baselines}
\label{sec:Appendix_baselines}

\subsection{Climatology}
\label{sec:Appendix_baselines_climatology}

We followed \cite{GloFAS_Forecast} to define the climatology baseline. For this, we computed climatology for the long-term record of river discharge data (1979-2018) with a moving window of 31 days centered on the day-of-the-year. Then, we computed 11 fixed quantiles at $10$\% interval for each day-of-the-year. As a result, the climate distribution changes with lead time, reflecting the dynamic changes in local river discharge patterns over time. Climatology is commonly applied to medium- and extended range lead times, where seasonal patterns predominantly influence the river discharge forecast \cite{GloFAS_Forecast}.

\subsection{Persistence}
\label{sec:Appendix_baselines_persistence}

We defined the persistence baseline as the daily river discharge of GloFAS reanalysis from the day preceding the day at which the forecast was issued, i.e., for a forecast starting at 00:00 UTC $\X^t$, the persistence is defined as the averaged river discharge between 00:00 UTC $\X^{t-1}$ and 00:00 UTC $\X^{t}$. This value was used as a prediction for the entire lead time. Persistence is primarily applied to short lead times, where the correlation of sequential river discharge values predominantly influences the forecasts \cite{GloFAS_Forecast}. Note that this baseline is unrealistic since no reanalysis is available directly at time $t$.

\subsection{LSTM}
\label{sec:Appendix_baselines_lstm}

We adopted the same LSTM architecture as described in \cite{Google_nature}.
The model follows an encoder–decoder structure, where the encoder is a bi-directional “hindcast” LSTM that processes historical input data, and the decoder is a uni-directional “forecast” LSTM that generates predictions over a 7-day forecast horizon based on forecast inputs.
To ensure fair comparison and benchmarking, we used the same input data (i.e., we include GloFAS reanalysis and exclude IMERG and nowcasting data for LSTM), train–test split, and normalization strategies as in the RiverMamba model.
To remain consistent with \cite{Google_nature}, we trained the model only at locations with available gauge observations, specifically the 3366 GRDC stations (see Sec.~\ref{sec:Appendix_GRDC_data}) rather than using a global training setup. Thus for LSTM, we do not include any spatial connections and space filling curves are not used in combination with the LSTM baseline.

The model leverages both dynamic and static inputs. For the hindcast LSTM, we used a 14-day sequence of dynamic inputs including CPC precipitation, GloFAS reanalysis, and ERA5-Land reanalysis data. At each time step, static attributes derived from the LISFLOOD model are embedded and concatenated with the dynamic inputs. For the forecast LSTM, we used ECMWF HRES forecasts as dynamic inputs over the 7-day horizon, with the static attributes concatenated in the same manner.

To connect the encoder and decoder, we employed a “state” layer consisting of two transfer networks (\colorh{\url{https://neuralhydrology.github.io/}}): a linear cell-state transfer network and a nonlinear hidden-state transfer network (a fully connected layer with hyperbolic tangent activation). A linear output head is applied at each forecast step to predict streamflow, and the model is trained using the mean squared error (MSE) loss. Unlike \cite{Google_nature}, we focus on deterministic prediction, so we do not implement a probabilistic output head or probabilistic loss function. In total, the model has 834,421 parameters.

In \cite{Google_nature}, an input sequence length of 365 days was used. This is because the model in \cite{Google_nature} has to simulate the states (i.e., soil moisture) and current runoff from the meteorological forcing input. 
In our experiments, since the states and the streamflow already integrate the meteorological signal of the past, we trained the LSTM model using a range of input sequence lengths from 4 to 90 days. We observed only marginal performance gains beyond a certain point, and identified 14 days as an optimal input sequence length.

The reported LSTM results are averaged over an ensemble of three independently trained models, each initialized with a different random seed.
Each training batch contains data from all 3366 GRDC stations at a given time step, with a batch size of 1—effectively training on 3366 samples per mini-batch. Training takes approximately 12 hours on four NVIDIA A100 GPUs for 35 epochs.
More details about the model architecture can be found in \cite{Google_nature}, as well as in the NeuralHydrology GitHub repository (\colorh{\url{https://neuralhydrology.github.io/}}).
Table \ref{table:lstm_parameter} summarizes the key hyperparameters used in our implementation of the LSTM model.

\begin{table}[!h]
\small
 \caption{Implementation details of the LSTM model.}\label{table:lstm_parameter}
 \centering
 \setlength\tabcolsep{2pt}
 \begin{tabular}{l c}
 \toprule
 Configuration & Value \\
 \midrule
Hidden size in hindcast LSTM  & 256 \\
Hidden size in forecast LSTM & 128 \\
Hidden size in static embedding layer & 20 \\
Hidden size in dynamic embedding layer & 20 \\
Hidden size in state layer  & 128 \\
Number of layers & 1 \\
Dropout at output regression head  & 0.4 \\
Dropout at state layer  & 0.1 \\
Learning rate & 0.0003 \\
Learning rate scheduler & Cosine annealing \\
Batch size & 1 with (3366 samples) \\
Optimizer & Adam \\
beta1 momentum term & 0.9 \\
beta2 momentum term  & 0.999 \\
weight decay & 0 \\
\bottomrule
\end{tabular}
\end{table}

It is important to note that \cite{Google_nature} did not release the full code or the full hyperparameter configurations of their final model, but only the pretrained checkpoints were made available. 
Although the saved models can be loaded for inference using the original inputs, it is not possible to retrain or adapt these models to a different input setup, which was required for our experiments.
We therefore used the published checkpoints and the NeuralHydrology GitHub repository as a reference to re-implement and train the LSTM.

All results shown in the paper for the LSTM baseline are obtained by our trained LSTM, except in sections~\ref{sec:Appendix_google_reforecast_ungauged} and \ref{sec:Appendix_google_reforecast_gauged}, where we compare with the published reforecast of Google’s LSTM obtained from \cite{Google_nature}. 

\subsection{GloFAS Forecast}
\label{sec:Appendix_baselines_glofas_forecast}

Operational forecast from GloFAS was obtained from the ECMWF Early Warning Data Store (EWDS) \colorh{\url{https://doi.org/10.24381/cds.ff1aef77}}. 
This represents real-time data from the official system for operational flood forecasting from the Copernicus Emergency Management Service (CEMS) and managed and developed by the European Commission’s Joint Research Centre. GloFAS forecast is produced by forcing the LISFLOOD model with the ECMWF ensemble forecast (ENS) up to $30$ days.
GloFAS forecast uses ENS meteorological forcing twice a day at 00:00 UTC. The high-resolution GloFAS v.4.0 forecast is available from 2023-07-26.
We compare to this baseline in Sec.~\ref{sec:Appendix_comparison_glofas_forecast}.

\subsection{GloFAS Reforecast}
\label{sec:Appendix_baselines_glofas_reforecast}

This baseline is similar to GloFAS forecast (Sec.~\ref{sec:Appendix_baselines_glofas_forecast}), however, GloFAS reforecast are forecasts run over the past with the new system version 4.0. The reforecast is available until 2023 and does not span the full testing split. We use this baseline for the main comparison with GloFAS in the main paper and in Sec.~\ref{sec:Appendix_results_obs}.

\newpage

\section{Space-filling curves}
\label{sec:Appendix_space_filling_curves}

\begin{figure}[!h]
  \centering
  \includegraphics[width=.96\textwidth]{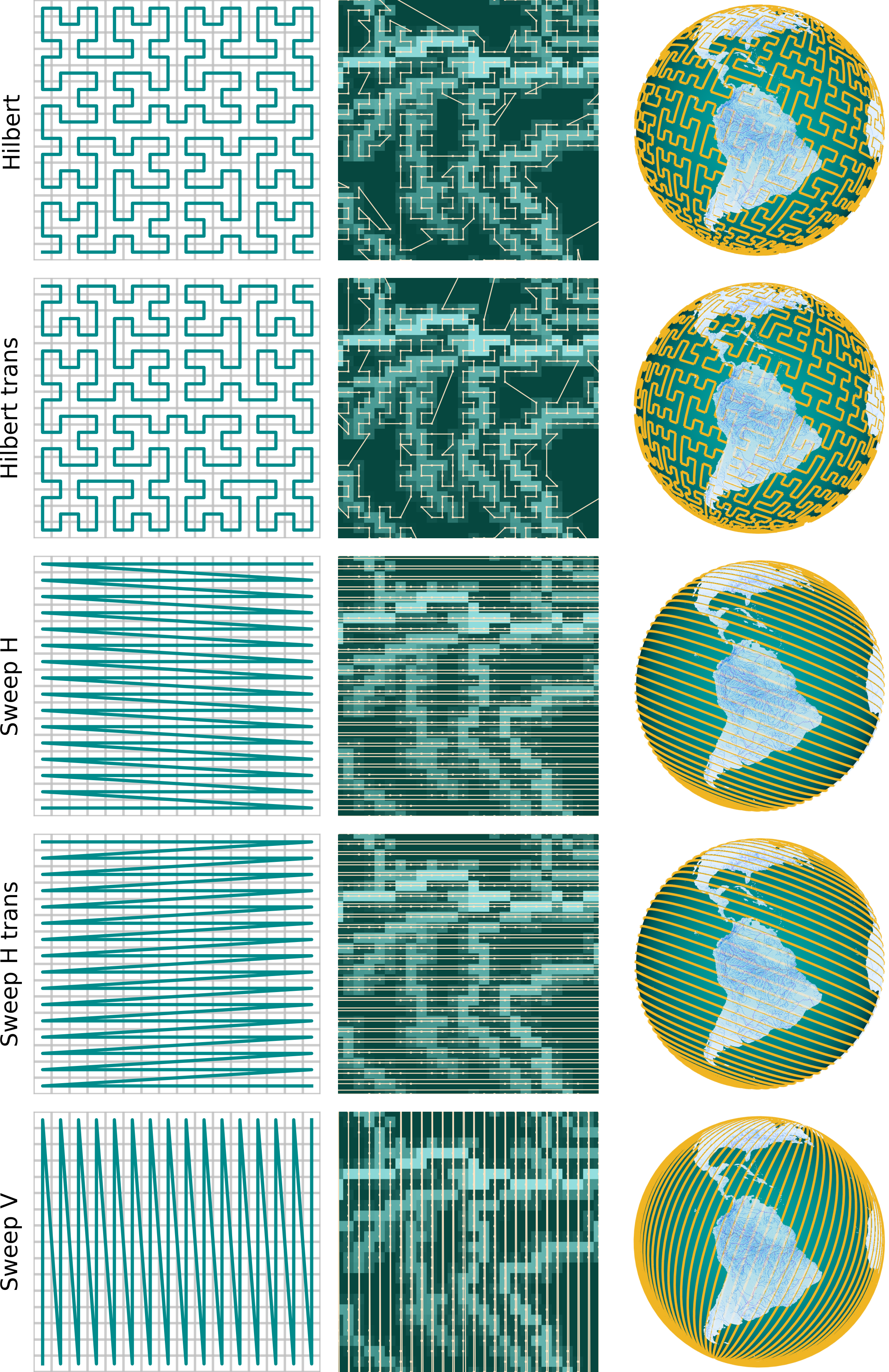}
  \caption{
  Visualization of different types of space-filling curves. 
  For each type, we show the space-filling curve over a 2D discrete space (left), zoomed in version over the Earth where the points are sorted via a specific serialization order within the space-filling curve (middle), and simplified 3D visualization of the curve over the Earth (right). }\label{fig:curve_type_1}
\end{figure}

\begin{figure}[!h]
  \centering
  \includegraphics[width=.96\textwidth]{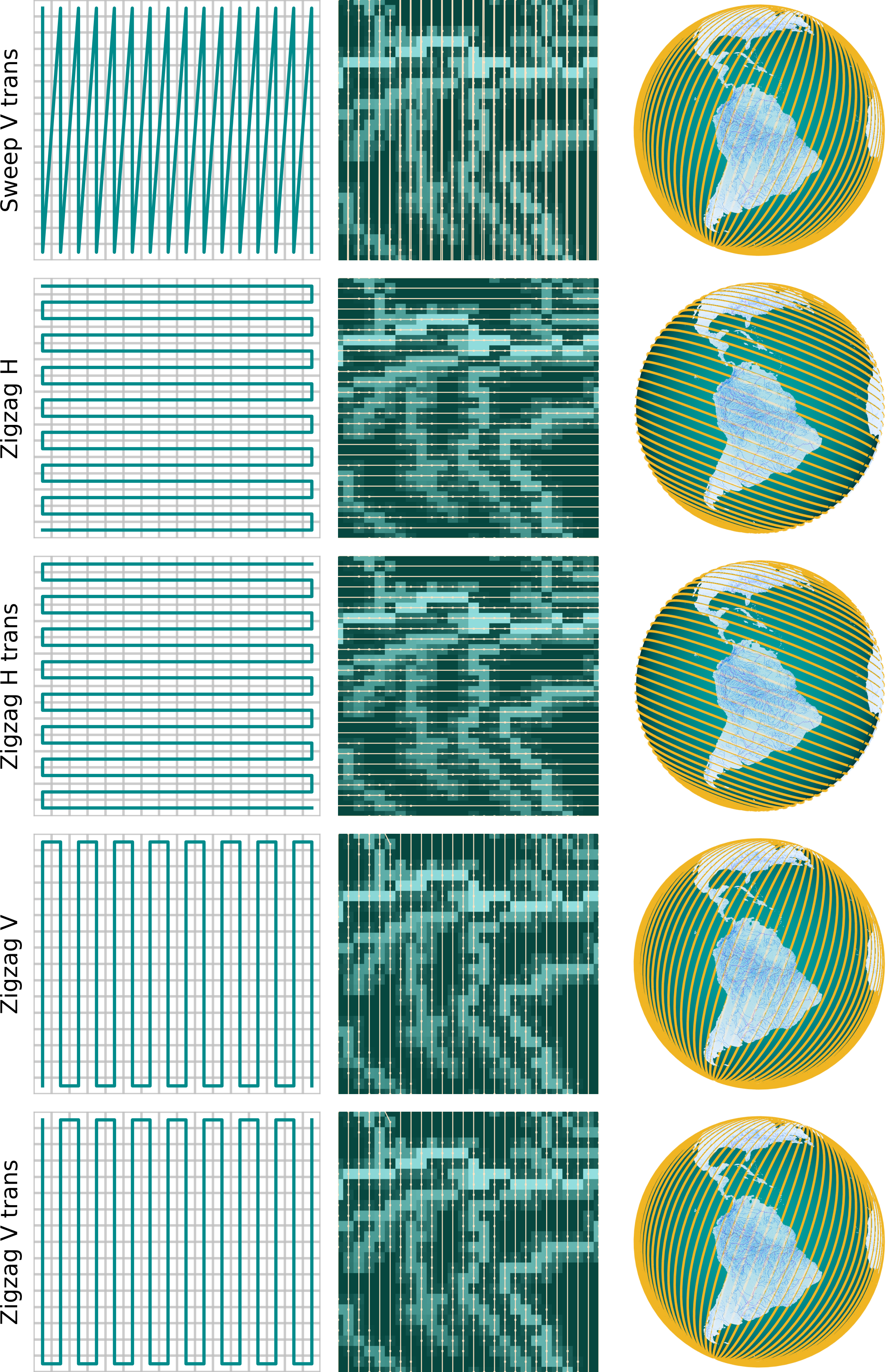}
  \caption{Visualization of different types of space-filling curves. 
  For each type, we show the space-filling curve over a 2D discrete space (left), zoomed in version over the Earth where the points are sorted via a specific serialization order within the space-filling curve (middle), and simplified 3D visualization of the curve over the Earth (right). }\label{fig:curve_type_2}
\end{figure}

Serialized encoding maps a point’s position into an integer index representing its order within the given space-filling curve. Each point is stored as a 64-bit integer.
For simplicity, we define the curves on the 2D PlateCarree projection of the Earth. As illustrated in Figs.~\ref{fig:curve_type_1} and \ref{fig:curve_type_2}, the serialization is done according to the sorted serialized encoding of all points with $\Phi: \ZZ^3 \rightarrow \NN$.
Due to the nature of the bijective transformation, there is an inverse mapping $\Phi^{-1}: \NN \rightarrow \ZZ^3$ which allows for the mapping of the encoded index back into the point’s position $p_i \in \ZZ^3$ (or $p_i \in \RR^3$ in case of a continuous space). This inverse mapping is called the serialized decoding or the deserialization.
In the following, we describe the mapping for each curve:

\textbf{Sweep.} This curve fills in the space like a spherical helix or a Luxodrome around the sphere. 

\textbf{Zigzag.} This curve is similar to the Sweep curve. The main difference is that the transformation ensures that every neighboring points on the curve are also neighboring in the physical space. 

\textbf{Generalized Hilbert.} Generalized Hilbert (Gilbert) is a Hilbert space-filling curve \cite{hilbert1935stetige} for rectangular domains of arbitrary non-power of two sizes \cite{pseudo-hilbert}. We used the numpy implementation of (\colorh{\url{https://github.com/jakubcerveny/gilbert}}) to generate the curves.
Transposed Gilbert is generated as $y_{(transpose)} = H - y$, where $y \in [1, H]$.

\clearpage

\section{Experiments on HydroRIVERS}
\label{sec:hydrorivers}

\begin{figure}[!h]
  \centering
  \includegraphics[width=.98\textwidth]{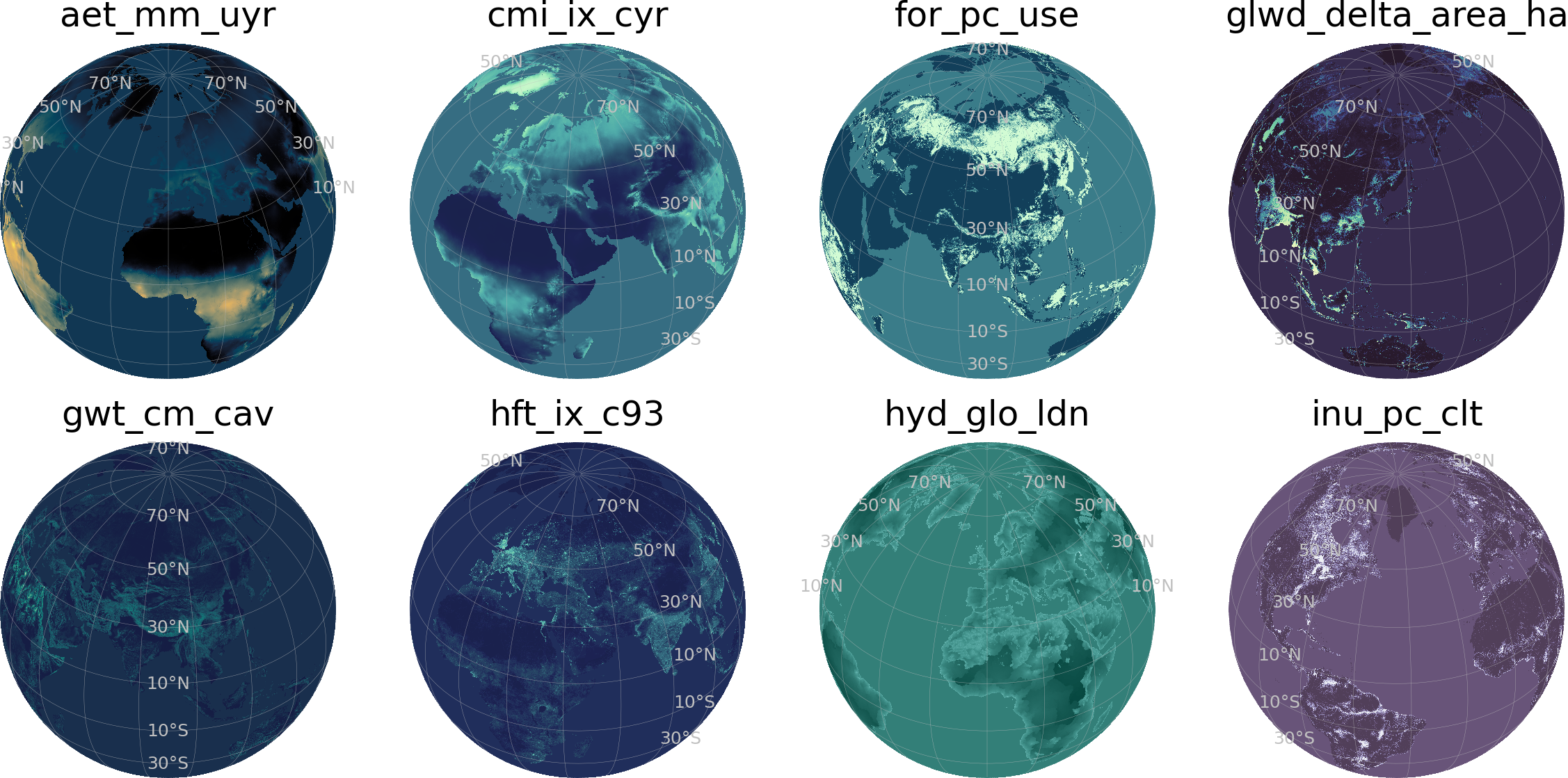}
  \caption{An overview of the processed HydroRIVERS static features. HydroRIVERS is mapped into the GloFAS domain.}\label{fig:hydrorivers}
\end{figure}

HydroRIVERS data are widely used to train deep learning models in hydrology. In this section, we explore the performance of RiverMamba with HydroRIVERS \cite{HydroRIVERS, HydroATLAS}. For this, we obtained static river attributes from \colorh{\url{https://www.hydrosheds.org/products/hydrorivers}}. 
The data is stored as a shape file. To map them onto the GloFAS domain, we first extract the coordinates of the rivers and then project them with the grid points on the WGS-84 ellipsoid (Eq.~\ref{eq:11} and \ref{eq:12}). Then, for each GloFAS grid point, depending on the attribute type, we either average the attributes or take the most frequent attribute within a radius of $5$ km. If no attributes were found, we increase the radius to $12$ km, and $24$ km, respectively.
We processed 299 river feature attributes overall and experimented with 103 features, i.e., we removed the monthly attribute statistics from the static features. Fig.~\ref{fig:hydrorivers} gives an overview of the processed HydroRIVERS data.

In Table \ref{table:ablation_hydrorivers}, we compare the LISFLOOD with the HydroRIVERS static maps for prediction on both GloFAS reanalysis and GRDC data. Using HydroRIVERS performs worse than using LISFLOOD static maps.

\begin{table}[h!]
  \caption{Ablation studies on the validation set over Europe.} \label{table:ablation_hydrorivers}
  \centering
  \tabcolsep=6.0pt\relax
  \setlength\extrarowheight{0pt}
  \begin{tabular}{*{4}c}
  
    \toprule
    
    HydroRIVERS & LISFLOOD & KGE | F1 ($\uparrow$) & KGE | F1 ($\uparrow$) \\

     &  & {\small(Reanalysis)} & {\small(GRDC Obs)} \\
    
    \cmidrule[\heavyrulewidth]{1-4}

    \xmark & \xmark & \colorm{0.9091} | \colorf{0.2505} & \colorm{0.7633} | \colorf{0.1118} \\

    \cmark & \xmark & \colorm{0.9174} | \colorf{0.2622} & \colorm{0.7406} | \colorf{0.1227} \\
        
    \xmark \bestc & \cmark \bestc & \bestc \textbf{\colorm{0.9205}} | \textbf{\colorf{0.2875}} & \bestc \textbf{\colorm{0.7838} | \colorf{0.1335}} \\
        
    \bottomrule
  
  \end{tabular}
\end{table}

\clearpage

\section{Additional results}
\label{sec:Appendix_results}

\subsection{Comparison with Google reforecast on ungauged GRDC}
\label{sec:Appendix_google_reforecast_ungauged}

In this section, we evaluate RiverMamba against the published reforecast by \cite{Google_nature} and available from \cite{grey_nearing_2023_10397664}. For this, we split our data into 8 folds and evaluate on ungauged stations. All stations were predicted similar to \cite{Google_nature} where each station was evaluated out-of-sample in both time and space. 
Note that the LSTM model for Google reforecast used more stations ($\sim$5680), while we used much less stations ($3366$). In addition, LSTM takes one year input as a hindcast, while RiverMamba takes only 4 days as input. Furthermore, RiverMamba does not use nowcasting data at time $t$ and starts the input initial conditions at $t-1$ to mimic an operational forecast.
There are also differences in the input initial conditions, i.e., LSTM uses precipitation estimates from the NASA Integrated Multi-satellite Retrievals for GPM (IMERG) early run as input. In addition, it uses HydroATLAS \cite{HydroATLAS} as geophysical and anthropogenic basin attributes. RiverMamba uses GloFAS reanalysis as an initial condition and LISFLOOD as static basin attributes.

\begin{figure}[!h]
  \centering
  \includegraphics[draft=\draft, width=.98\textwidth]{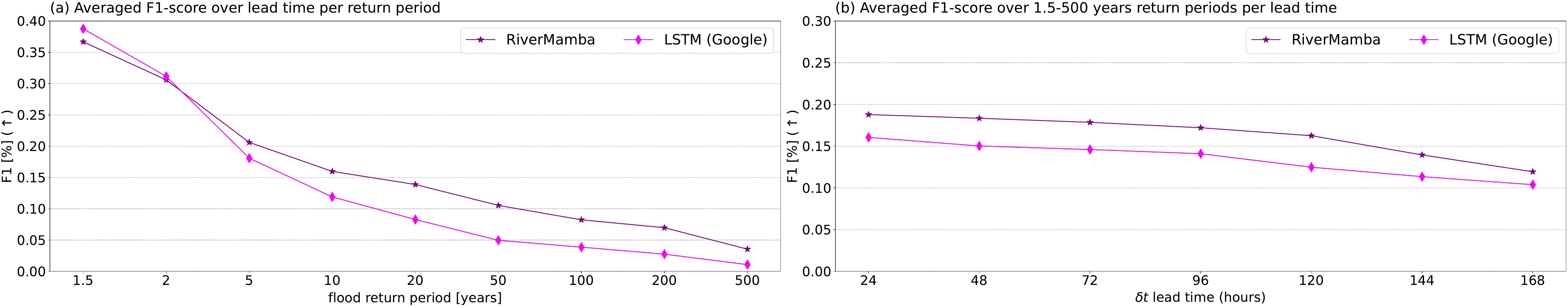}
  \caption{Comparison to Google reforecast on ungauged GRDC stations (test set 2014-2021 out-of-sample in space and time).}\label{fig:obs_google_ungauged}
\end{figure}

\begin{figure}[!h]
  \centering
  \includegraphics[draft=\draft, width=.8\textwidth]{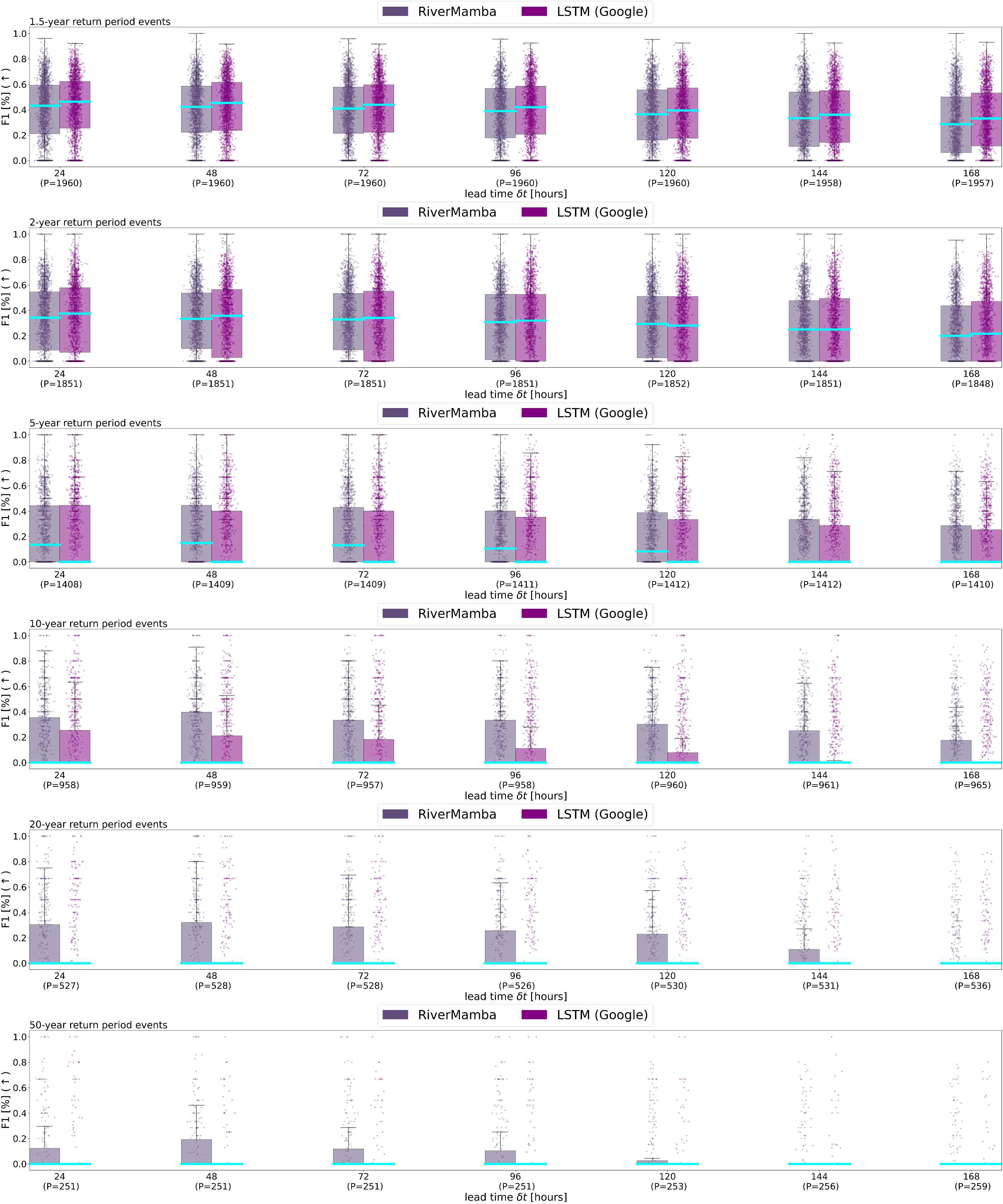}
  \caption{Comparison to Google reforecast. Shown is F1-score of flood forecasting for different return periods and lead time on ungauged GRDC stations (test set 2014-2021 out-of-sample in space and time). Distribution quartiles are displayed in boxes, and the entire range excluding outliers is displayed in whiskers. The median score for the model is shown by the cyan line in the box.}\label{fig:obs_google_ungauged_f1_cls}
\end{figure}

\begin{figure}[!h]
  \centering
  \includegraphics[draft=\draft, width=.8\textwidth]{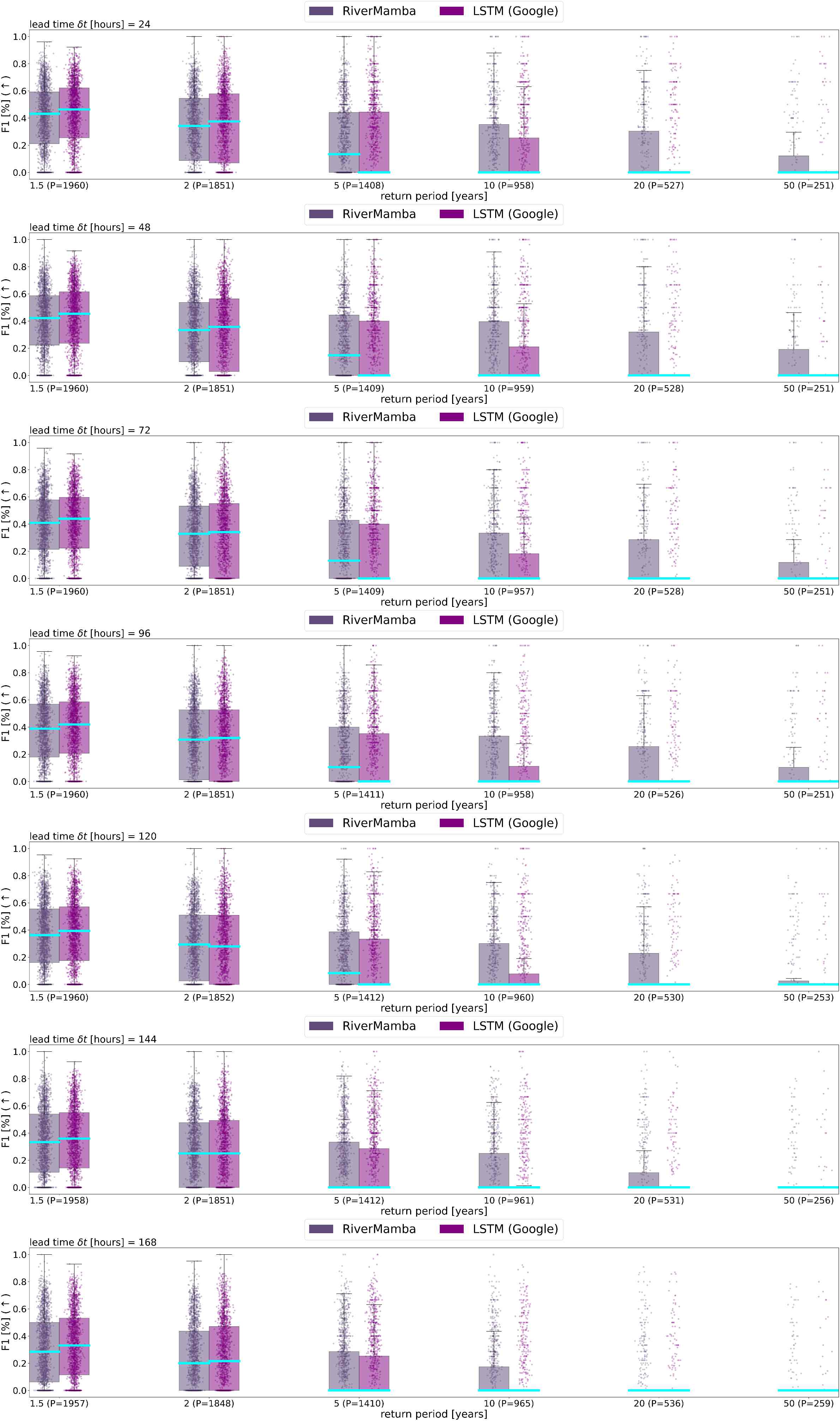}
  \caption{Comparison to Google reforecast. Shown is F1-score of flood forecasting for different lead time and return periods ($1.5$ - $50$ years) on ungauged GRDC stations (test set 2014-2021 out-of-sample in space and time). Distribution quartiles are displayed in boxes, and the entire range excluding outliers is displayed in whiskers. The median score for the model is shown by the cyan line in the box.}\label{fig:obs_google_ungauged_f1_time}
\end{figure}

Table \ref{table:appendix_ungauged_google_reforecast} shows the overall performance for the years 2014-2021. The F1-score is averaged for all lead times and 1.5-20 year return periods.
We expect that adding nowcasting (analysis data) and IMERG as input and an ensemble would improve the results of RiverMamba on ungauged basins further. The ungauged streamflow forecast becomes also better when the number of stations increases.
More results are shown in Figs.\ref{fig:obs_google_ungauged}-\ref{fig:obs_google_ungauged_f1_time}.

\begin{table}[!h]
  \caption{Comparison to Google reforecast on ungauged GRDC stations for the years 2014-2021. Shown is the averaged F1-score $(\uparrow)$ for all lead times and 1.5-20 year return periods.}
  \label{table:appendix_ungauged_google_reforecast}
  \centering
  \tabcolsep=2.5pt\relax
  \setlength\extrarowheight{0pt}
  \begin{tabular}{*{2}c}
    \toprule
    LSTM (Google reforecast from \cite{Google_nature}) & \bestc RiverMamba \\
    \midrule
    \colorf{0.2164} & \bestc \textbf{\colorf{0.2355}} \\
    \bottomrule
\end{tabular}
\end{table}

\clearpage

\subsection{Comparison with Google reforecast on gauged GRDC}
\label{sec:Appendix_google_reforecast_gauged}

Similar to Section~\ref{sec:Appendix_google_reforecast_ungauged}, here we compare to the published reforecast by \cite{Google_nature} and available from \cite{grey_nearing_2023_10397664} but on gauged stations where all stations were evaluated out-of-sample in time for the years 2019-2021.
For the differences between RiverMamba and the LSTM model by \cite{Google_nature}, see Sec.~\ref{sec:Appendix_google_reforecast_ungauged}.

\begin{figure}[!h]
  \centering
  \includegraphics[draft=\draft, width=.98\textwidth]{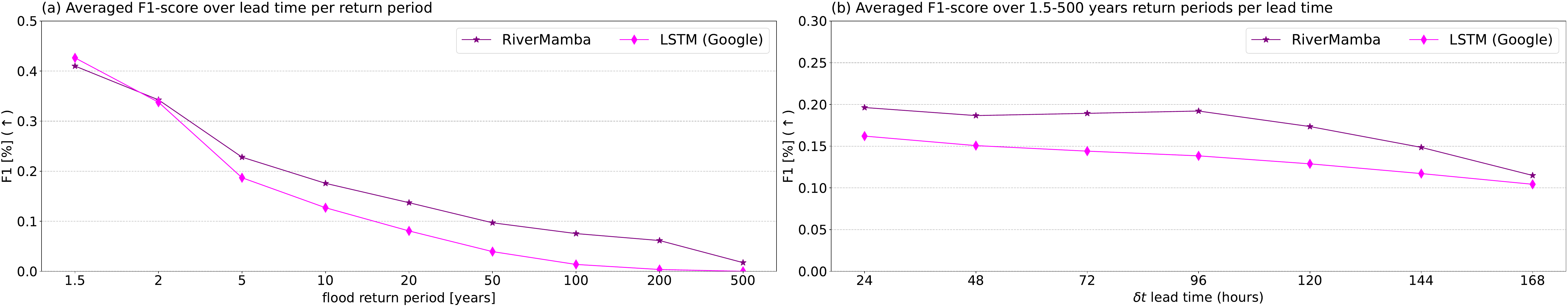}
  \caption{Comparison to Google reforecast on gauged GRDC stations (test set 2019-2021 out-of-sample in time).}\label{fig:obs_google_gauged}
\end{figure}

\begin{figure}[!h]
  \centering
  \includegraphics[draft=\draft, width=.8\textwidth]{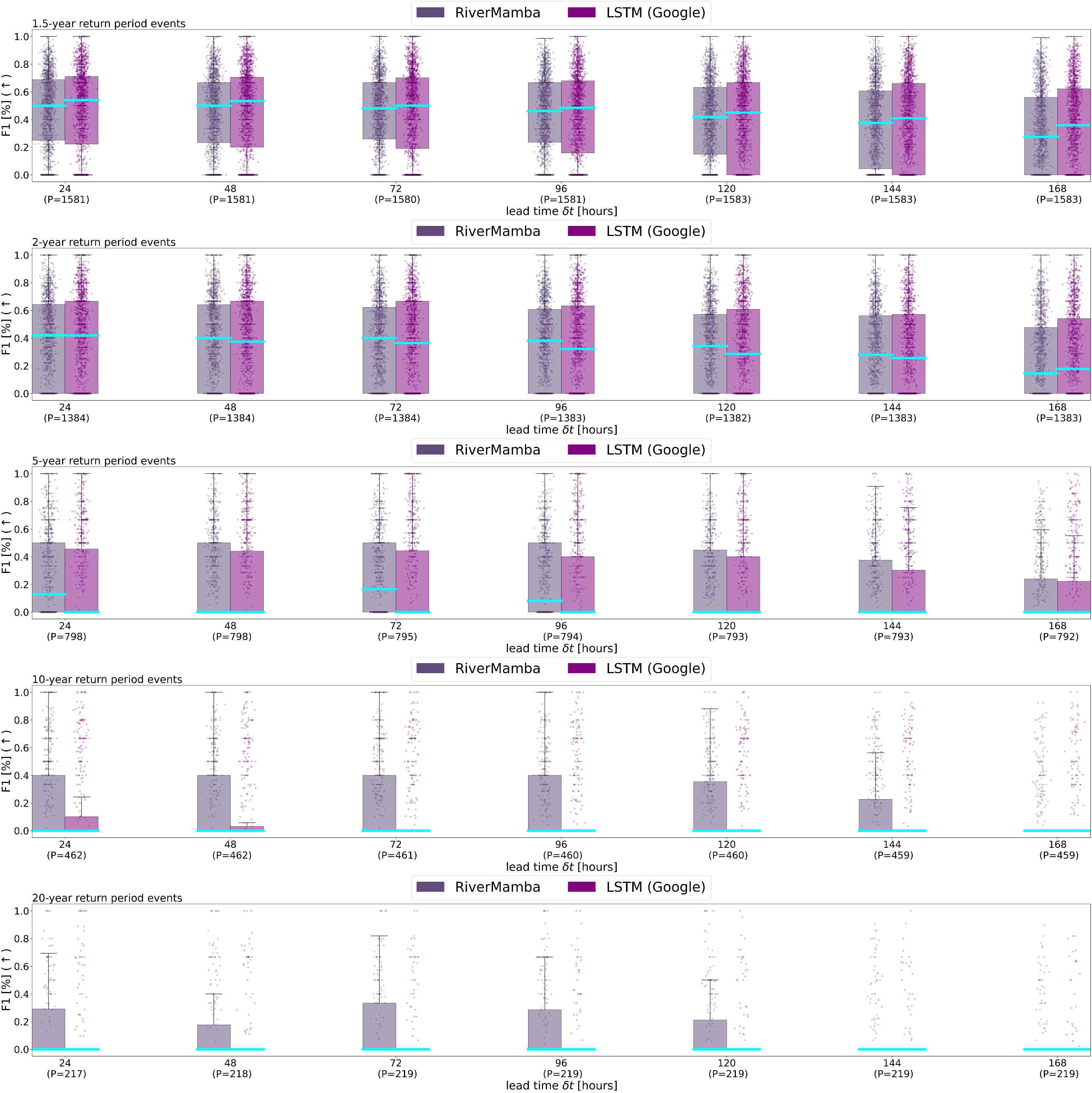}
  \caption{Comparison to Google reforecast. Shown is F1-score of flood forecasting for different return periods and lead time on gauged GRDC stations (test set 2019-2021 out-of-sample in time). Distribution quartiles are displayed in boxes, and the entire range excluding outliers is displayed in whiskers. The median score for the model is shown by the cyan line in the box.}\label{fig:obs_google_gauged_f1_cls}
\end{figure}

\begin{figure}[!h]
  \centering
  \includegraphics[draft=\draft, width=.8\textwidth]{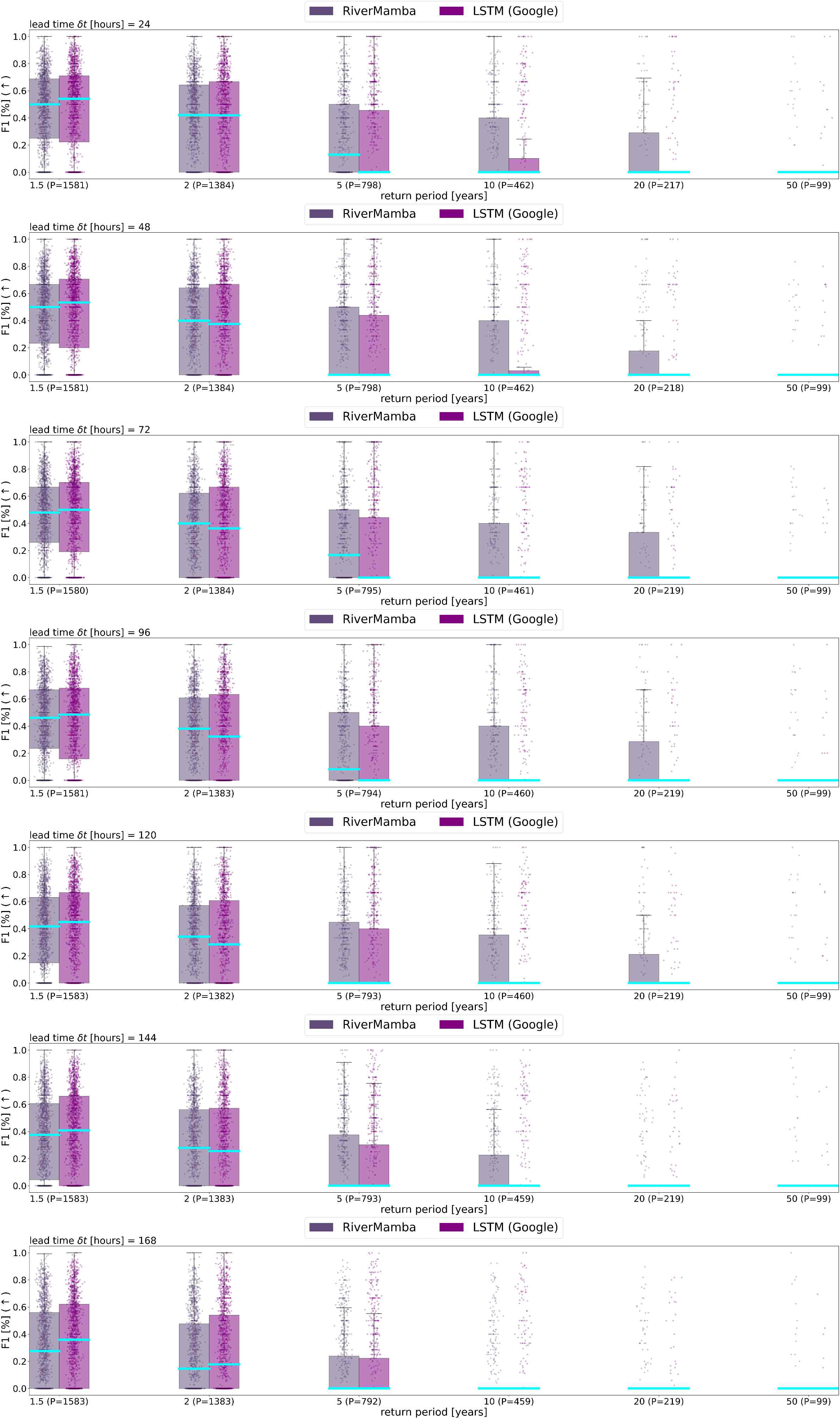}
  \caption{Comparison to Google reforecast. Shown is F1-score of flood forecasting for different lead time and return periods ($1.5$ - $50$ years) on gauged GRDC stations (test set 2019-2021 out-of-sample in time). Distribution quartiles are displayed in boxes, and the entire range excluding outliers is displayed in whiskers. The median score for the model is shown by the cyan line in the box.}\label{fig:obs_google_gauged_f1_time}
\end{figure}

Table \ref{table:appendix_gauged_google_reforecast} shows the overall performance for the years 2019-2021. The F1-score is averaged for all lead times and 1.5-20 year return periods. 
More results are shown in Figs.~\ref{fig:obs_google_gauged}-\ref{fig:obs_google_gauged_f1_time}.

\begin{table}[!h]
  \caption{Comparison to Google reforecast on gauged GRDC stations for the years 2019-2021. Shown is the averaged F1-score $(\uparrow)$ for all lead times and 1.5-20 year return periods.}
  \label{table:appendix_gauged_google_reforecast}
  \centering
  \tabcolsep=2.5pt\relax
  \setlength\extrarowheight{0pt}
  \begin{tabular}{*{2}c}
    \toprule
    LSTM (Google reforecast from \cite{Google_nature}) & \bestc RiverMamba \\
    \midrule
    \colorf{0.2318} & \bestc \textbf{\colorf{0.2587}} \\
    \bottomrule
\end{tabular}
\end{table}

\clearpage

\subsection{Additional results on gauged GloFAS reanalysis}
\label{sec:Appendix_results_reanalysis}

\begin{figure}[!h]
  \centering
  \includegraphics[draft=\draft, width=.98\textwidth]{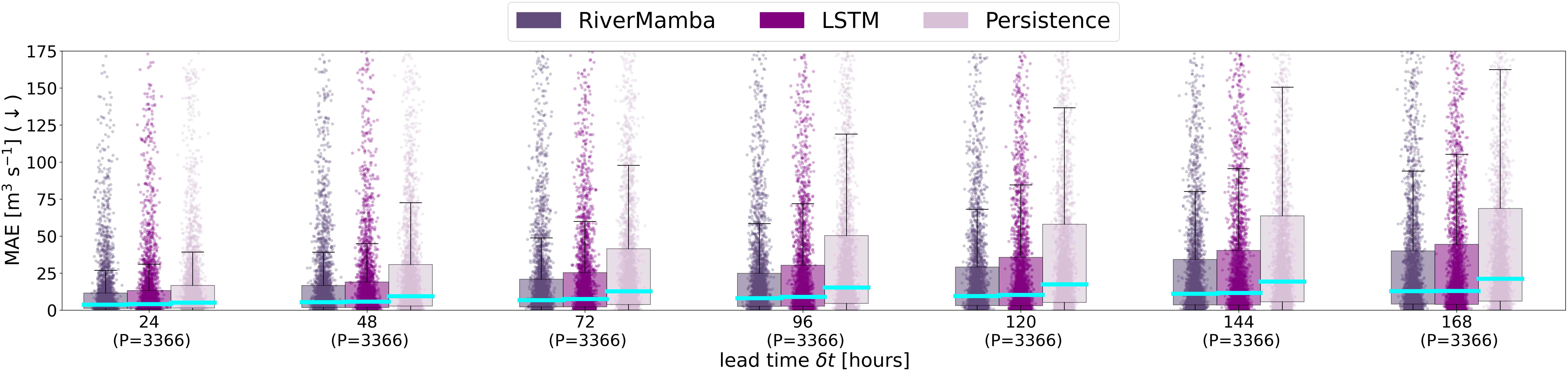}
  \caption{MAE of the river discharge forecasting with different lead time on GloFAS reanalysis (test set 2021-2024 temporally out-of-sample).}\label{fig:reanalysis_MAE}
\end{figure}

\begin{figure}[!h]
  \centering
  \includegraphics[draft=\draft, width=.98\textwidth]{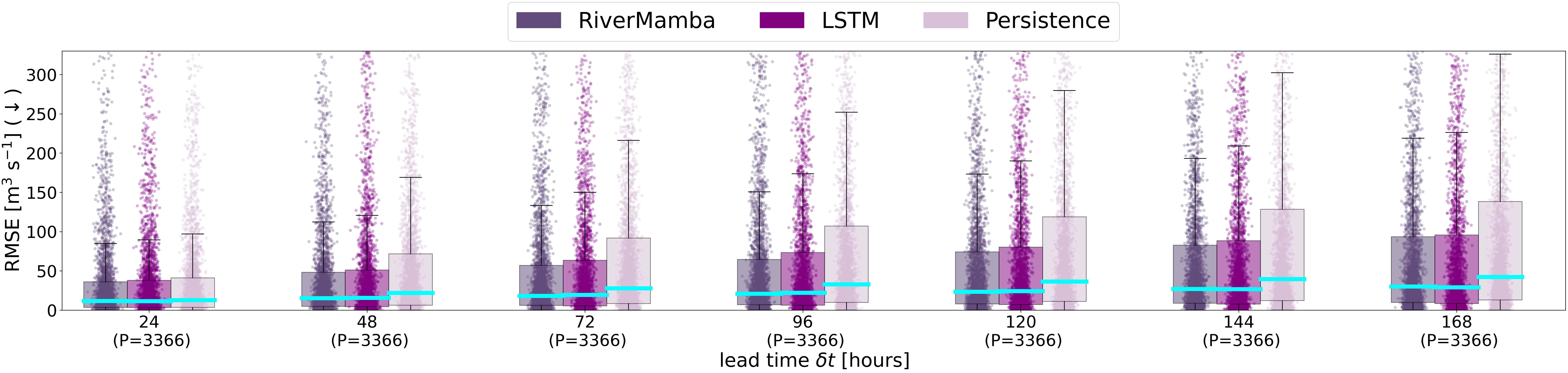}
  \caption{RMSE of the river discharge forecasting with different lead time on GloFAS reanalysis (test set 2021-2024 temporally out-of-sample).}\label{fig:reanalysis_RMSE}
\end{figure}

\begin{figure}[!h]
  \centering
  \includegraphics[draft=\draft, width=.98\textwidth]{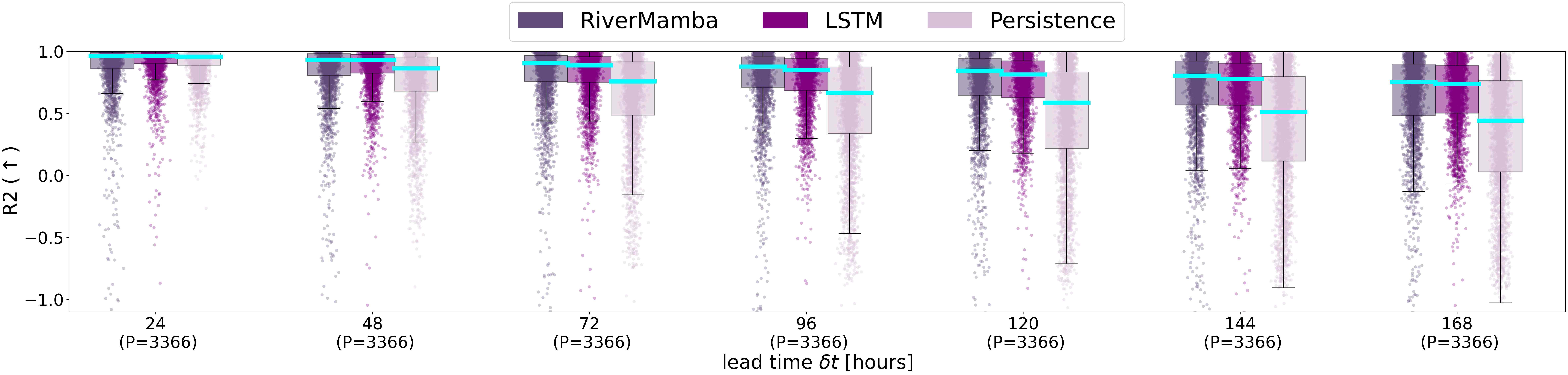}
  \caption{R2 (NSE) of the river discharge forecasting with different lead time on GloFAS reanalysis (test set 2021-2024 temporally out-of-sample).}\label{fig:reanalysis_R2}
\end{figure}

\begin{figure}[!h]
  \centering
  \includegraphics[draft=\draft, width=.98\textwidth]{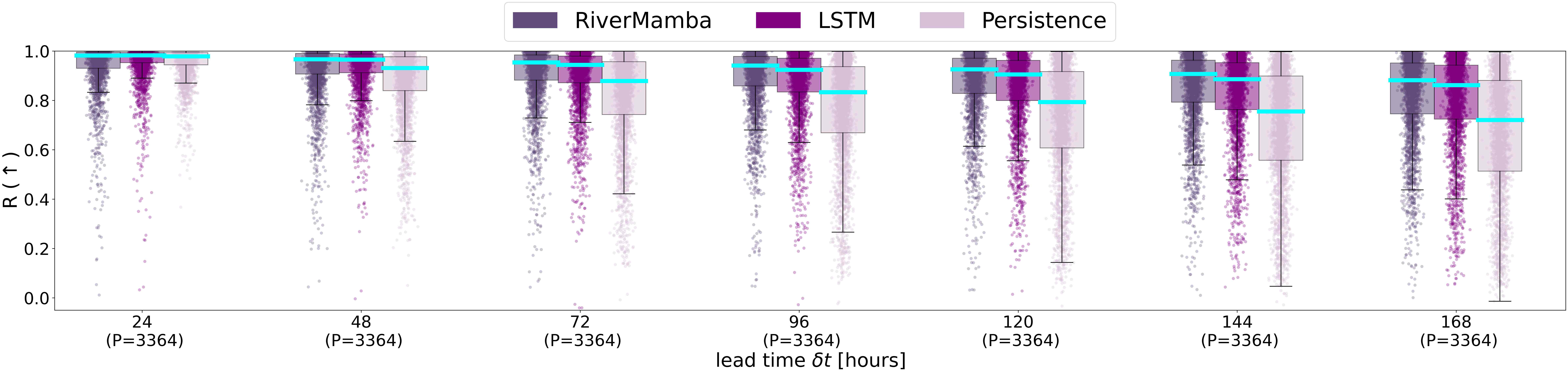}
  \caption{Pearson correlation (R) of the river discharge forecasting with different lead time on GloFAS reanalysis (test set 2021-2024 temporally out-of-sample).}\label{fig:reanalysis_R}
\end{figure}

\begin{figure}[!h]
  \centering
  \includegraphics[draft=\draft, width=.8\textwidth]{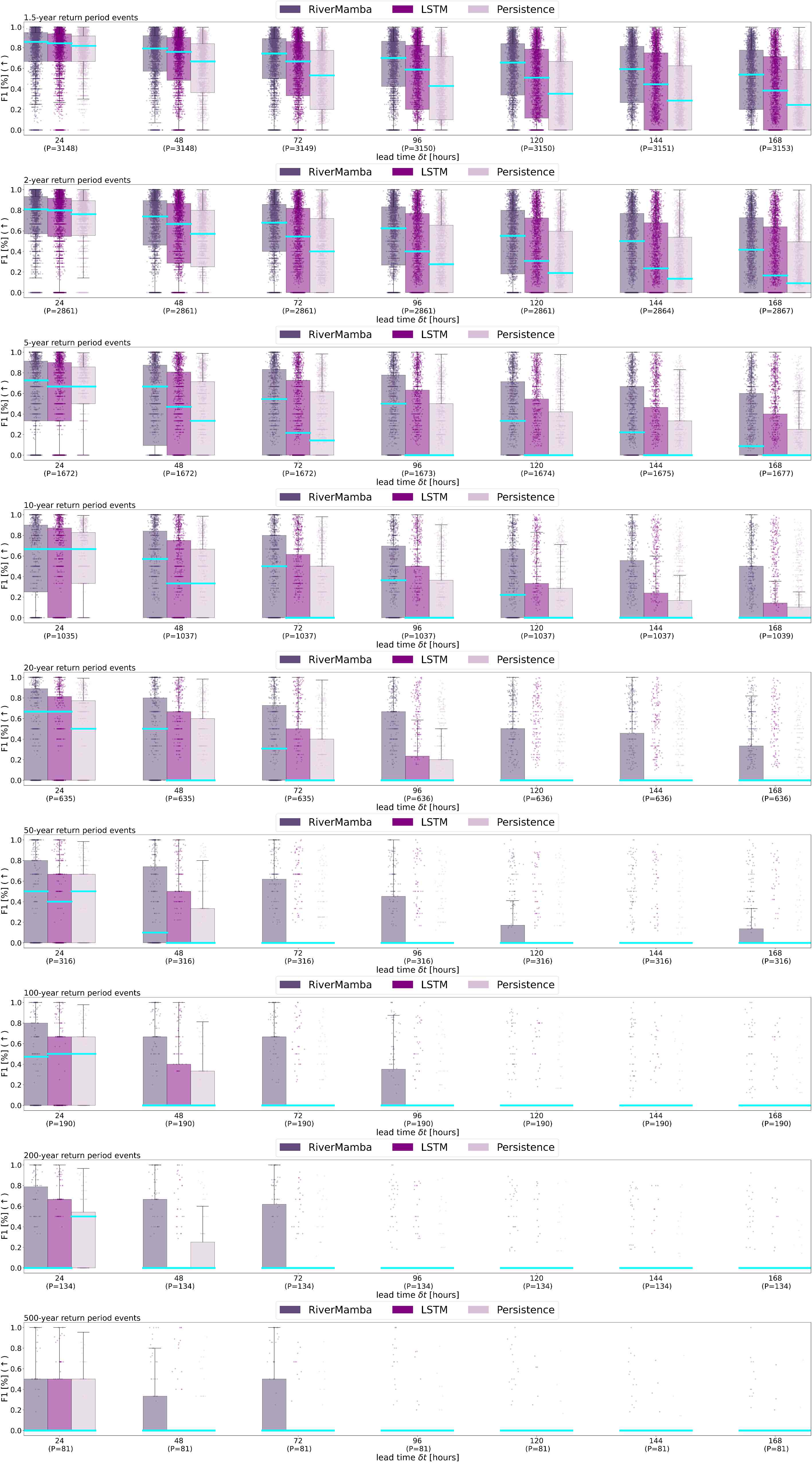}
  \caption{F1-score of flood forecasting for different return periods and lead time on GloFAS reanalysis (test set 2021-2024 temporally out-of-sample). Distribution quartiles are displayed in boxes, and the entire range excluding outliers is displayed in whiskers. The median score for the model is shown by the cyan line in the box.}\label{fig:reanalysis_f1_cls}
\end{figure}

\begin{figure}[!h]
  \centering
  \includegraphics[width=.8\textwidth]{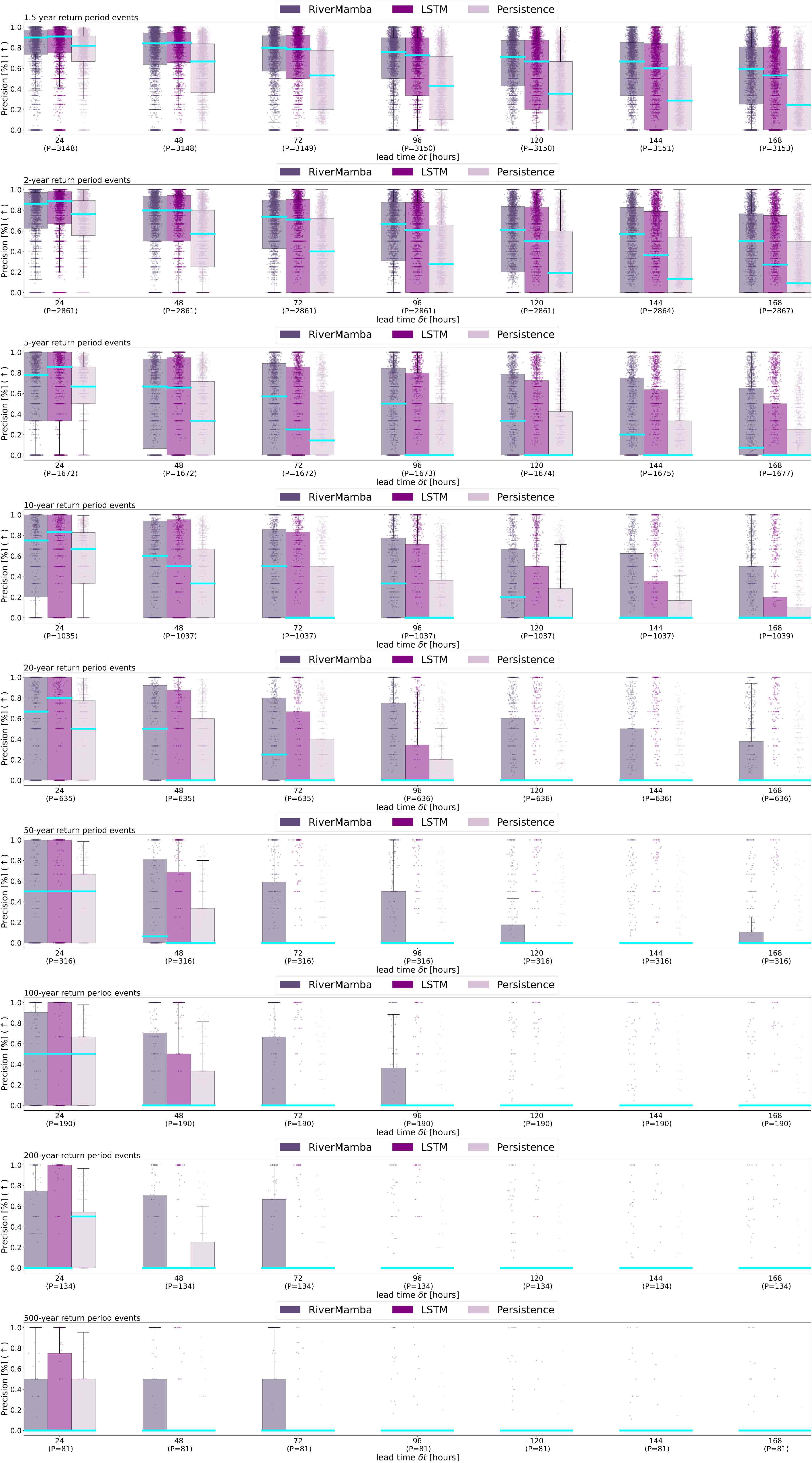}
  \caption{Precision of flood forecasting for different return periods and lead time on GloFAS reanalysis (test set 2021-2024 temporally out-of-sample). Distribution quartiles are displayed in boxes, and the entire range excluding outliers is displayed in whiskers. The median score for the model is shown by the cyan line in the box.}\label{fig:reanalysis_precision_cls}
\end{figure}

\begin{figure}[!h]
  \centering
  \includegraphics[width=.8\textwidth]{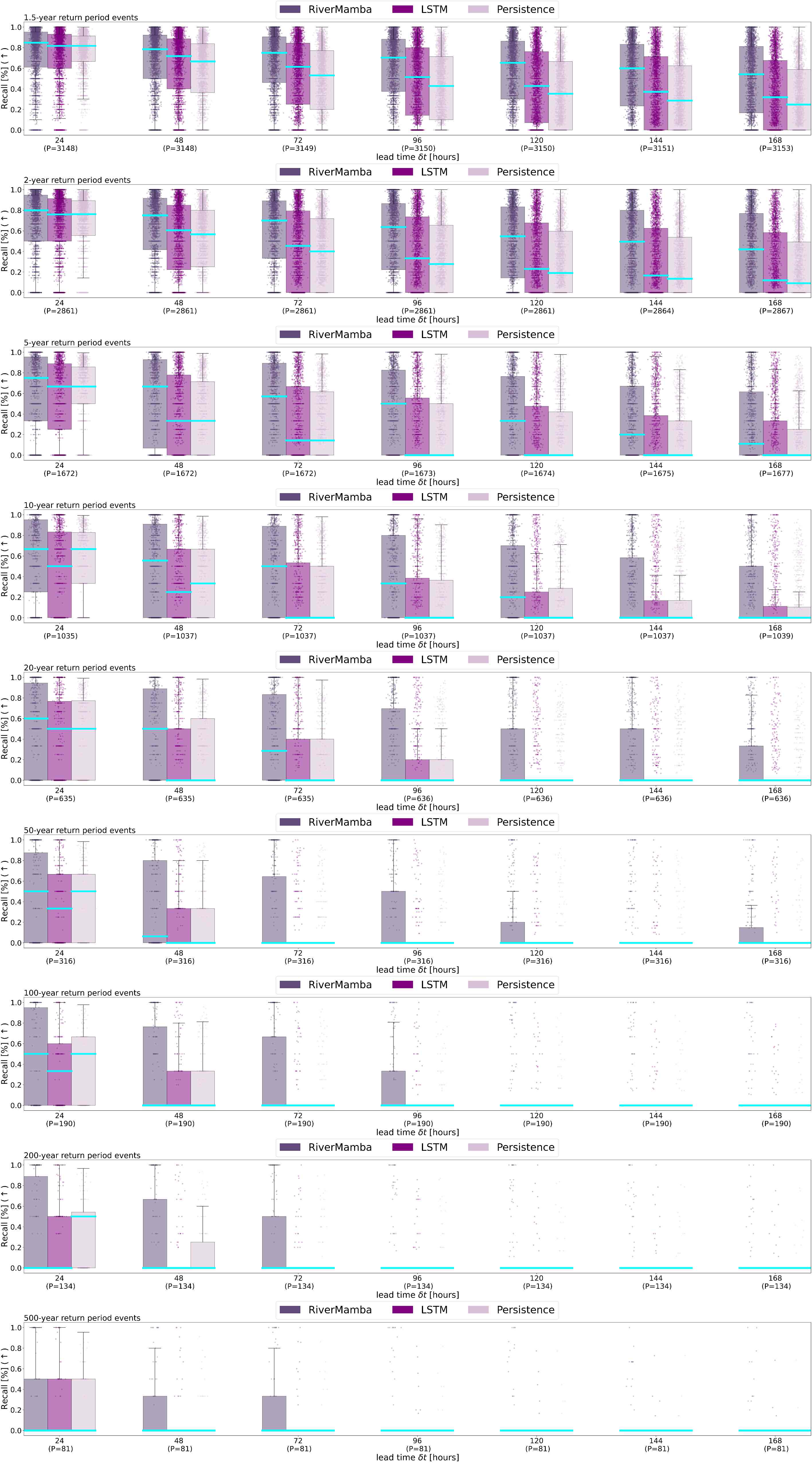}
  \caption{Recall of flood forecasting for different return periods and lead time on GloFAS reanalysis (test set 2021-2024 temporally out-of-sample). Distribution quartiles are displayed in boxes, and the entire range excluding outliers is displayed in whiskers. The median score for the model is shown by the cyan line in the box.}\label{fig:reanalysis_recall_cls}
\end{figure}

\begin{figure}[!h]
  \centering
  \includegraphics[draft=\draft, width=.8\textwidth]{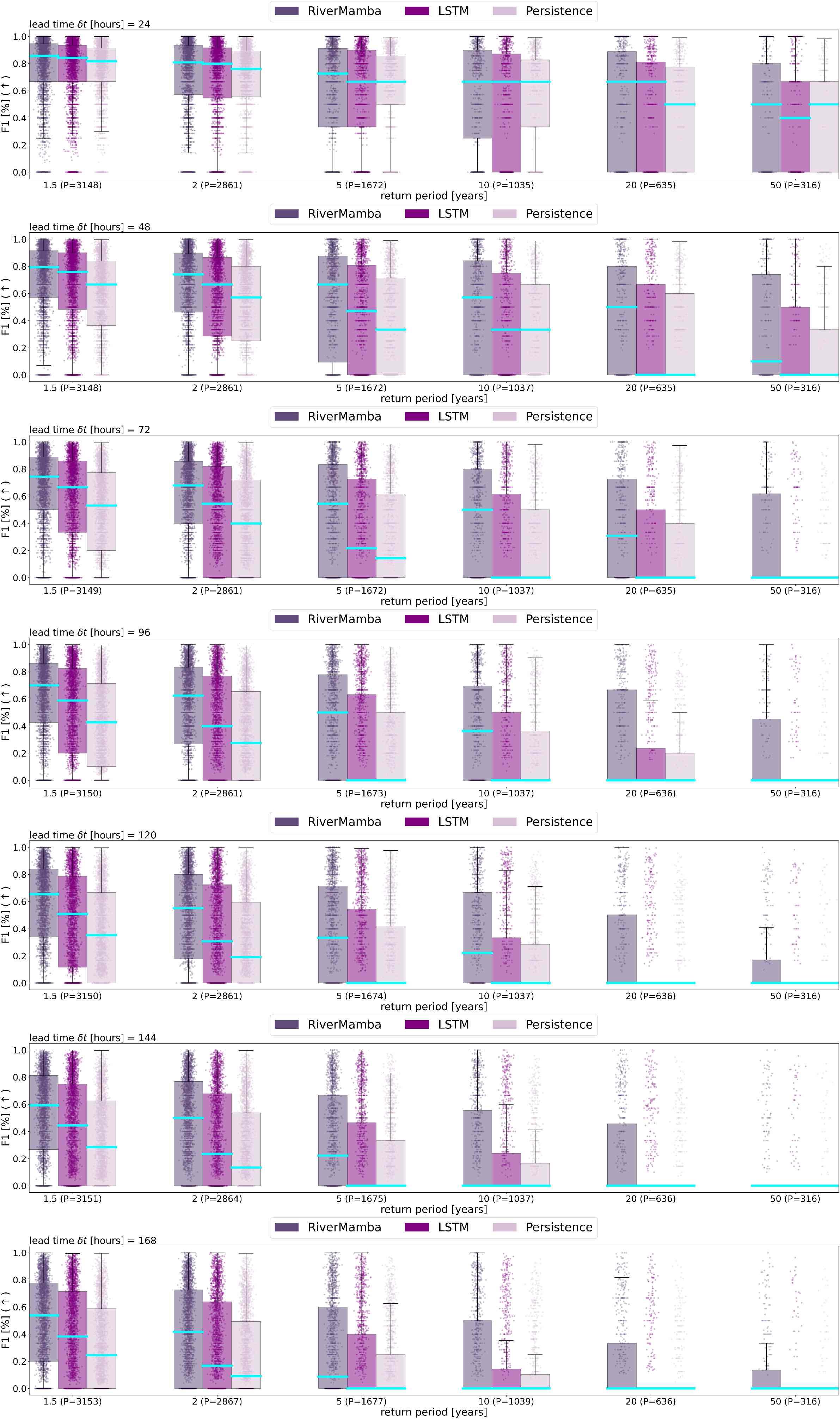}
  \caption{F1-score of flood forecasting for different lead time and return periods ($1.5$ - $50$ years) on GloFAS reanalysis (test set 2021-2024 temporally out-of-sample). Distribution quartiles are displayed in boxes, and the entire range excluding outliers is displayed in whiskers. The median score for the model is shown by the cyan line in the box.}\label{fig:reanalysis_f1_time}
\end{figure}

\begin{figure}[!h]
  \centering
  \includegraphics[draft=\draft, width=.8\textwidth]{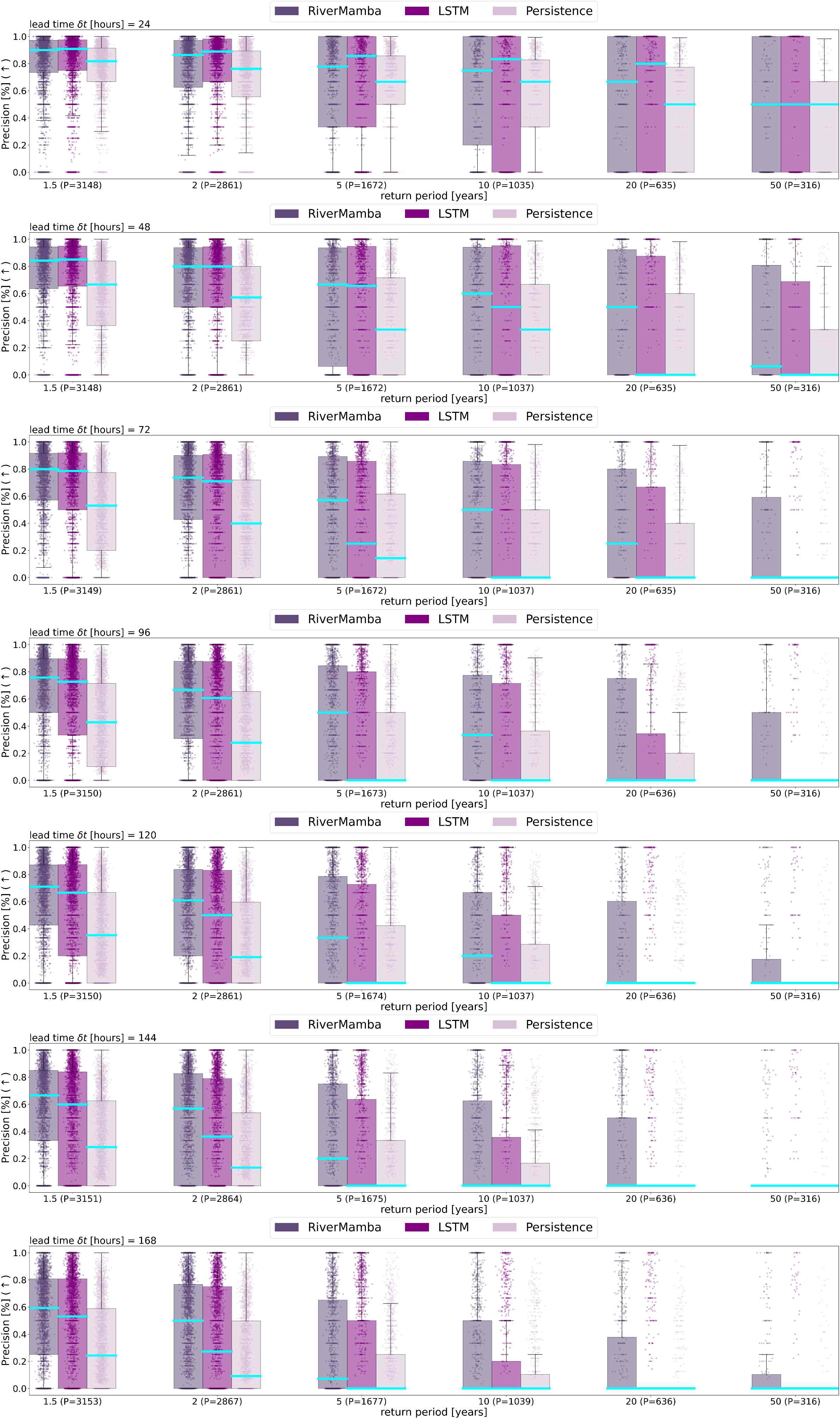}
  \caption{Precision of flood forecasting for different lead time and return periods ($1.5$ - $50$ years) on GloFAS reanalysis (test set 2021-2024 temporally out-of-sample). Distribution quartiles are displayed in boxes, and the entire range excluding outliers is displayed in whiskers. The median score for the model is shown by the cyan line in the box.}\label{fig:reanalysis_precision_time}
\end{figure}

\begin{figure}[!h]
  \centering
  \includegraphics[width=.8\textwidth]{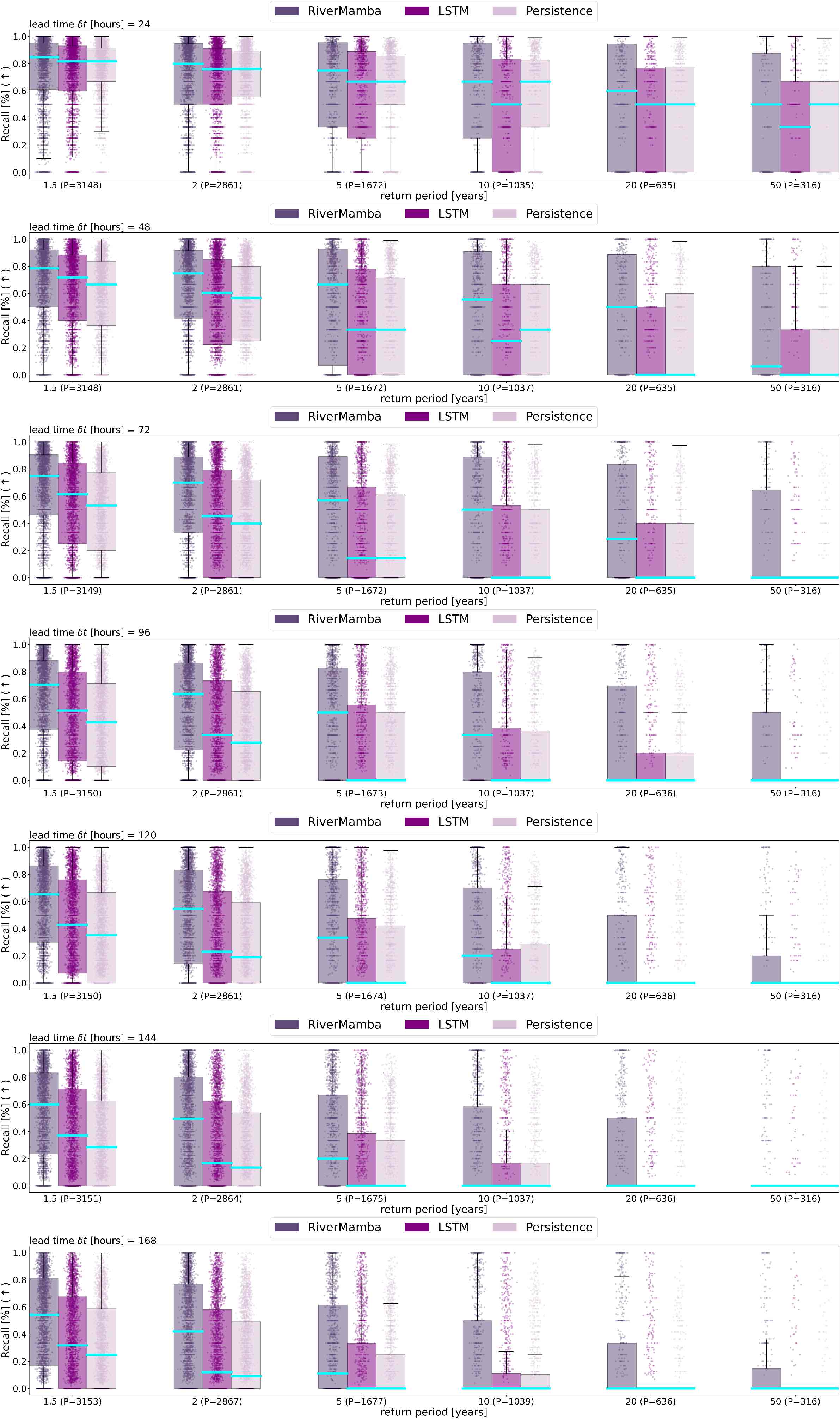}
  \caption{Recall of flood forecasting for different lead time and return periods ($1.5$ - $50$ years) on GloFAS reanalysis (test set 2021-2024 temporally out-of-sample). Distribution quartiles are displayed in boxes, and the entire range excluding outliers is displayed in whiskers. The median score for the model is shown by the cyan line in the box.}\label{fig:reanalysis_recall_time}
\end{figure}

\begin{figure}[!h]
  \centering
  \includegraphics[width=.97\textwidth]{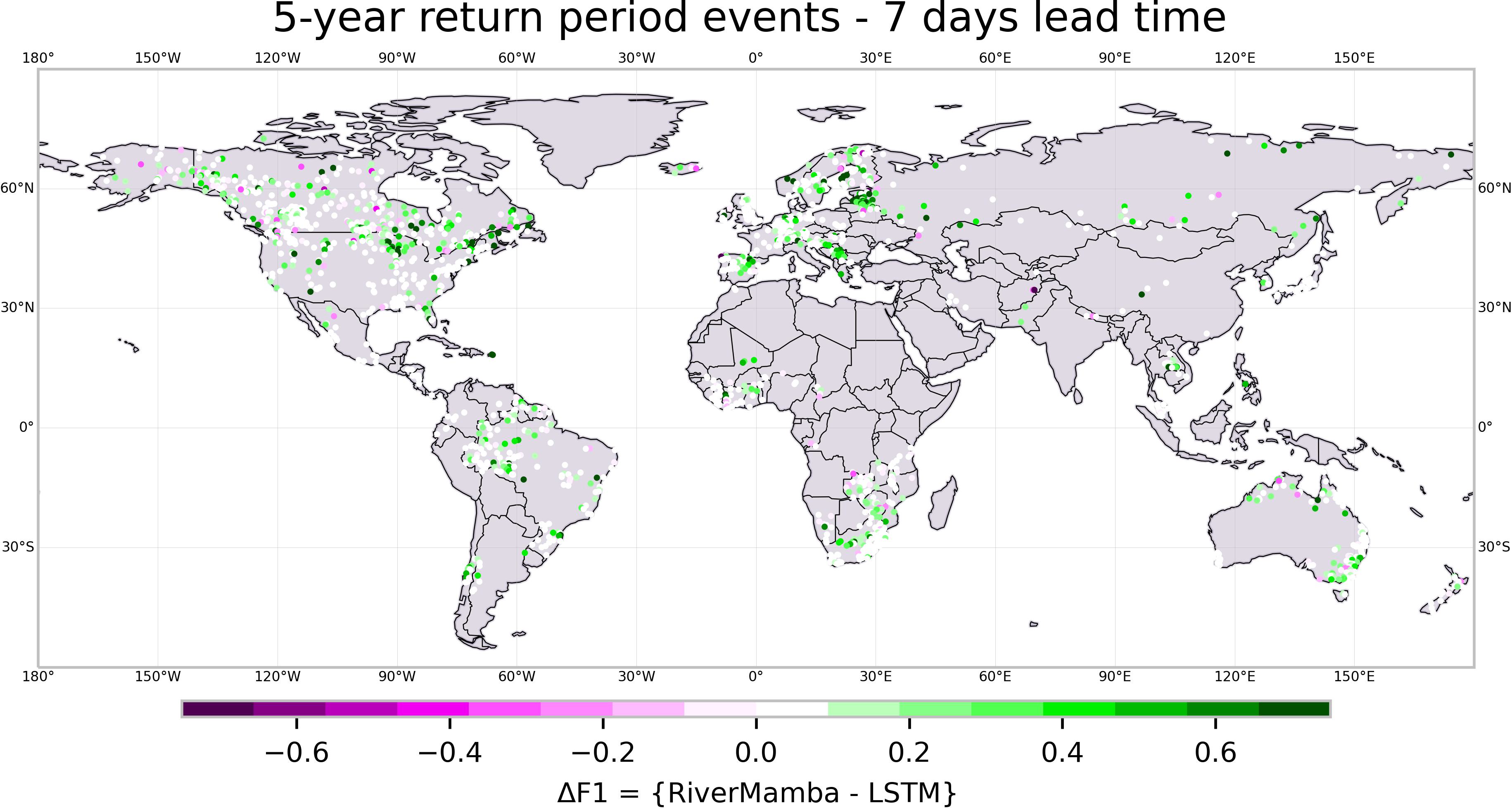}
  \caption{Comparison of F1-score between RiverMamba and LSTM on GloFAS reanalysis for the 5-year return period events (test set 2021-2024 temporally out-of-sample). RiverMamba improves over LSTM in 41\% of the stations (P=1677) and being better or equally better in 89\% of the stations.}\label{fig:reanalysis_delta_f1_lstm}
\end{figure}

\begin{figure}[!h]
  \centering
  \includegraphics[width=.97\textwidth]{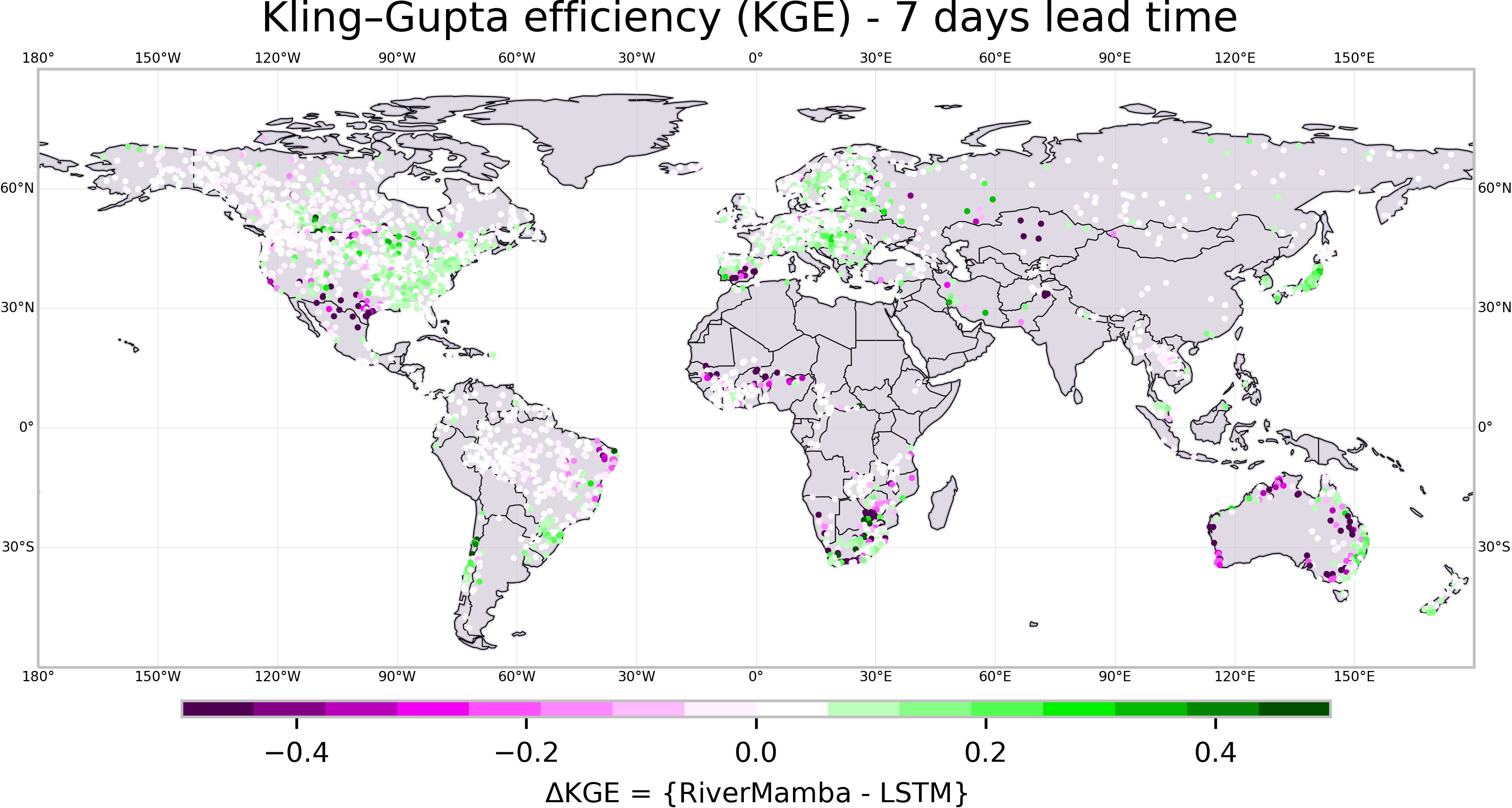}
  \caption{Comparison of KGE between RiverMamba and LSTM on GloFAS reanalysis (test set 2021-2024 temporally out-of-sample). RiverMamba improves over LSTM in 73\% of the stations (P=3364).}\label{fig:reanalysis_delta_kge_lstm}
\end{figure}

\begin{figure}[!h]
  \centering
  \includegraphics[draft=\draft, width=.98\textwidth]{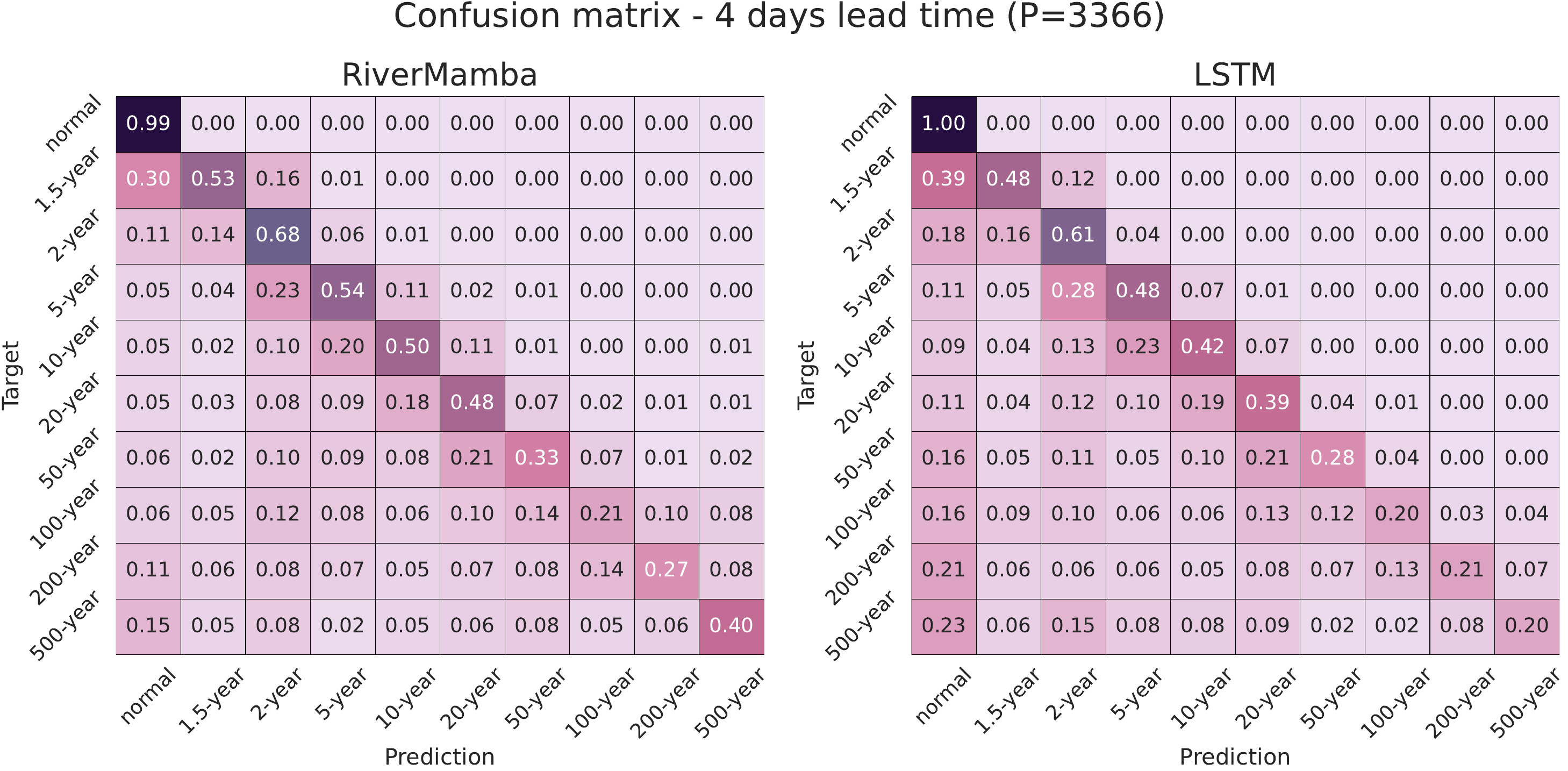}
  \caption{Comparison to LSTM on GloFAS reanalysis (test set 2021-2024 temporally out-of-sample).}\label{fig:reanalysis_confusion_matrix}
\end{figure}

In this section, we plot additional results for the experiments on GloFAS river discharge reanalysis. In Figs.~\ref{fig:reanalysis_MAE}-\ref{fig:reanalysis_R}, we report the results for MAE, RMSE, R2, and R metrics with lead time. In Figs.~\ref{fig:reanalysis_f1_cls}-\ref{fig:reanalysis_recall_time}, we report the results of F1-score, Precision, and Recall metrics for different return periods and lead times.
In Figs.~\ref{fig:reanalysis_delta_f1_lstm} and \ref{fig:reanalysis_delta_kge_lstm}, we compare the results between RiverMamba and LSTM for F1-score and KGE metrics spatially.
Finally, Fig.~\ref{fig:reanalysis_confusion_matrix} shows the confusion matrix for both RiverMamba and LSTM. 

\clearpage

\subsection{Additional results on gauged GRDC}
\label{sec:Appendix_results_obs}

\begin{figure}[!h]
  \centering
  \includegraphics[draft=\draft, width=.98\textwidth]{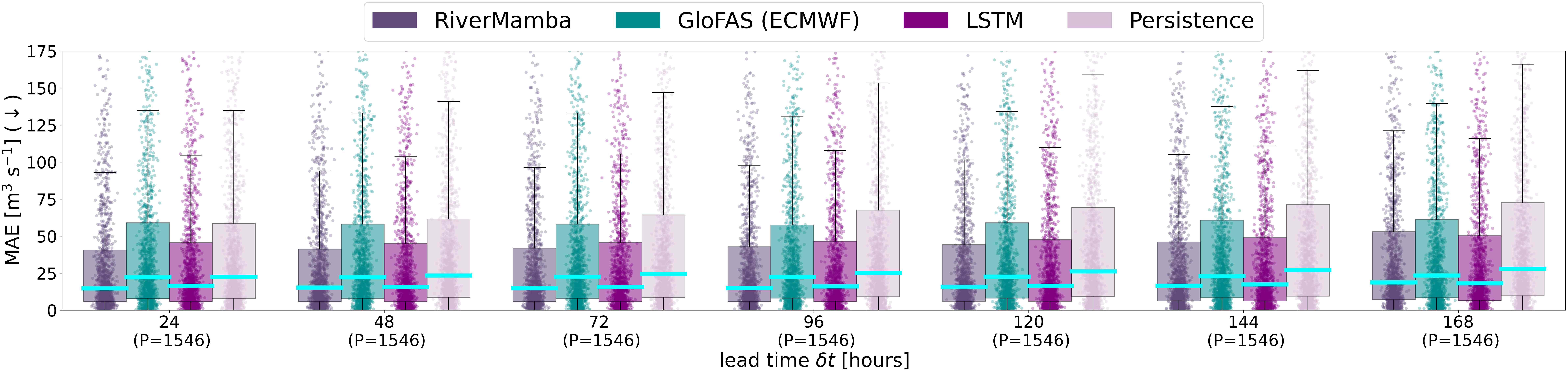}
  \caption{MAE of the river discharge forecasting with different lead time on GRDC observations (test set 2021-2023 temporally out-of-sample).}\label{fig:obs_MAE}
\end{figure}

\begin{figure}[!h]
  \centering
  \includegraphics[draft=\draft, width=.98\textwidth]{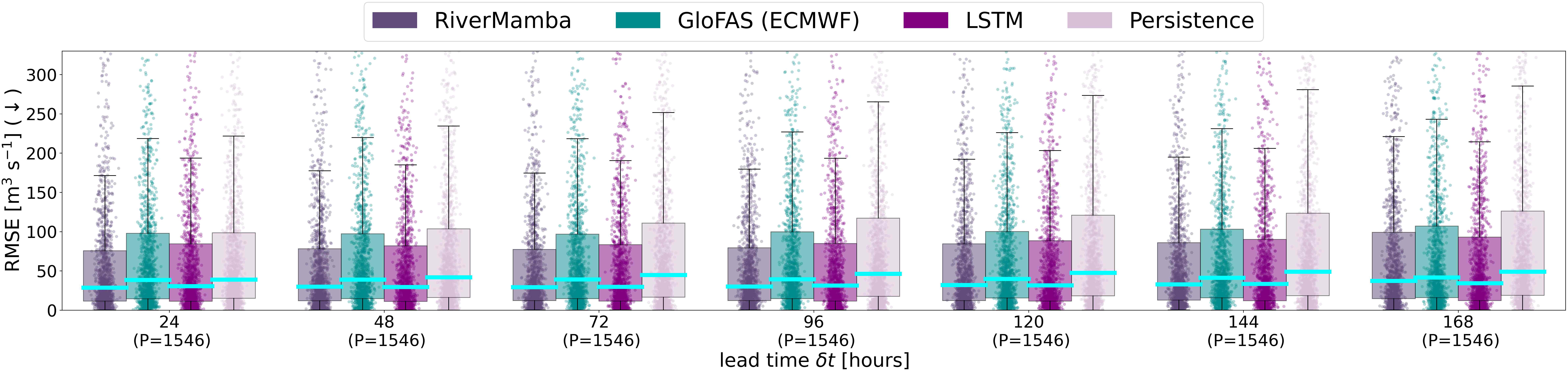}
  \caption{RMSE of the river discharge forecasting with different lead time on GRDC observations (test set 2021-2023 temporally out-of-sample).}\label{fig:obs_RMSE}
\end{figure}

\begin{figure}[!h]
  \centering
  \includegraphics[draft=\draft, width=.98\textwidth]{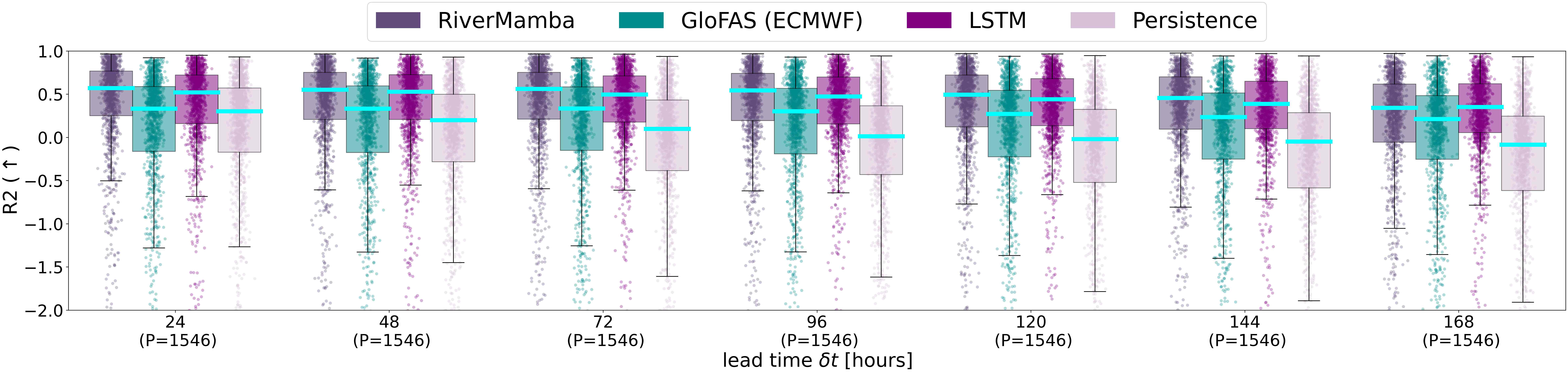}
  \caption{R2 (NSE) of the river discharge forecasting with different lead time on GRDC observations (test set 2021-2023 temporally out-of-sample).}\label{fig:obs_R2}
\end{figure}

\begin{figure}[!h]
  \centering
  \includegraphics[draft=\draft, width=.98\textwidth]{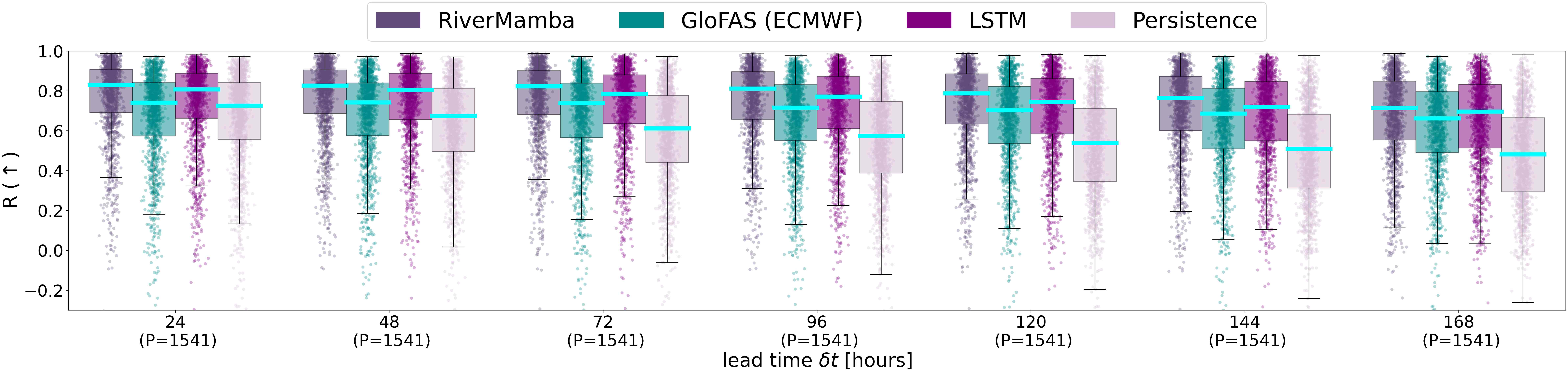}
  \caption{Pearson correlation (R) of the river discharge forecasting with different lead time on GRDC observations (test set 2021-2023 temporally out-of-sample).}\label{fig:obs_R}
\end{figure}

\begin{figure}[!h]
  \centering
  \includegraphics[draft=\draft, width=.98\textwidth]{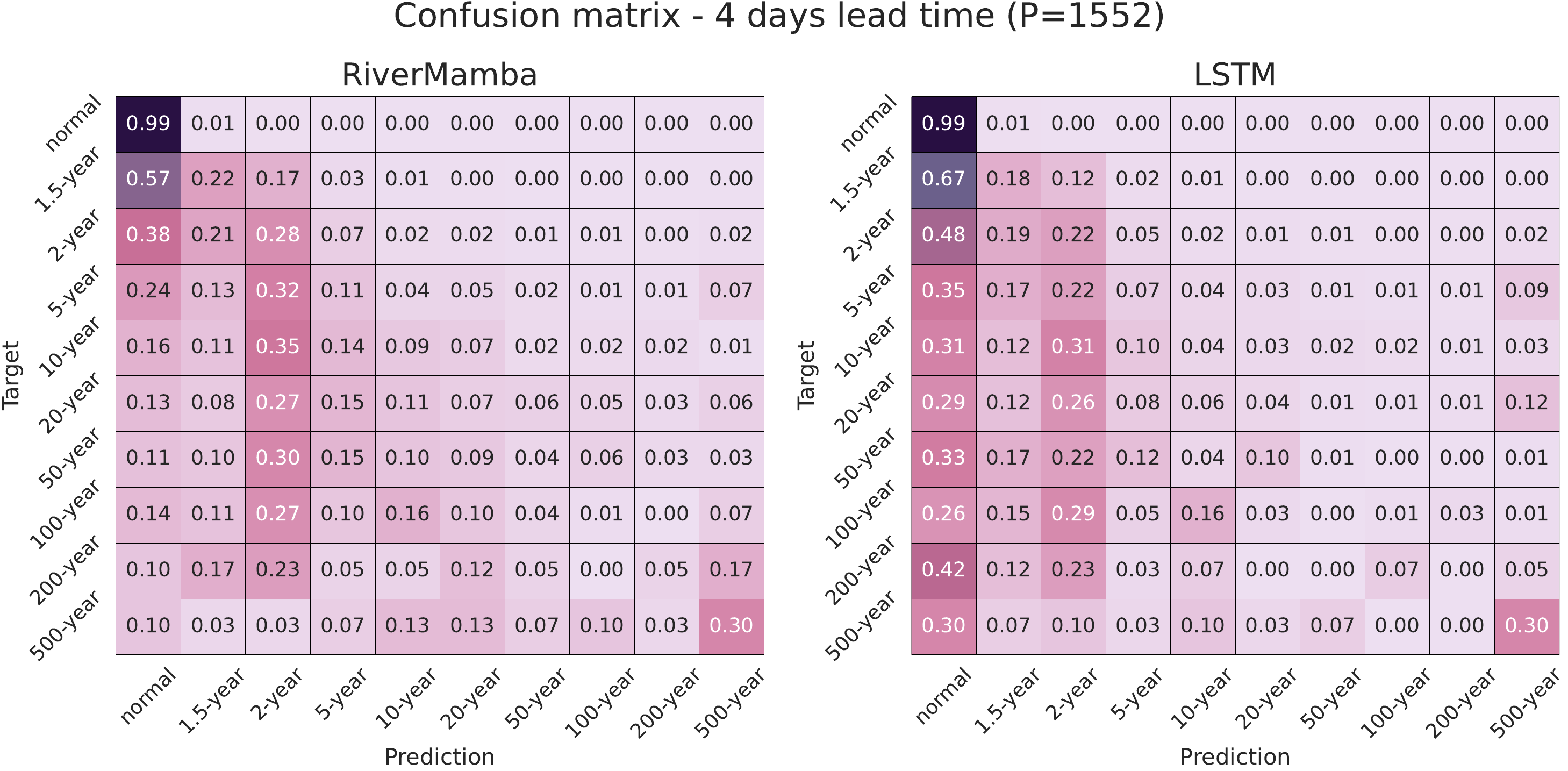}
  \caption{Comparison to LSTM on GRDC observations (test set 2021-2024 temporally out-of-sample).}\label{fig:obs_confusion_matrix_lstm}
\end{figure}

\begin{figure}[!h]
  \centering
  \includegraphics[draft=\draft, width=.98\textwidth]{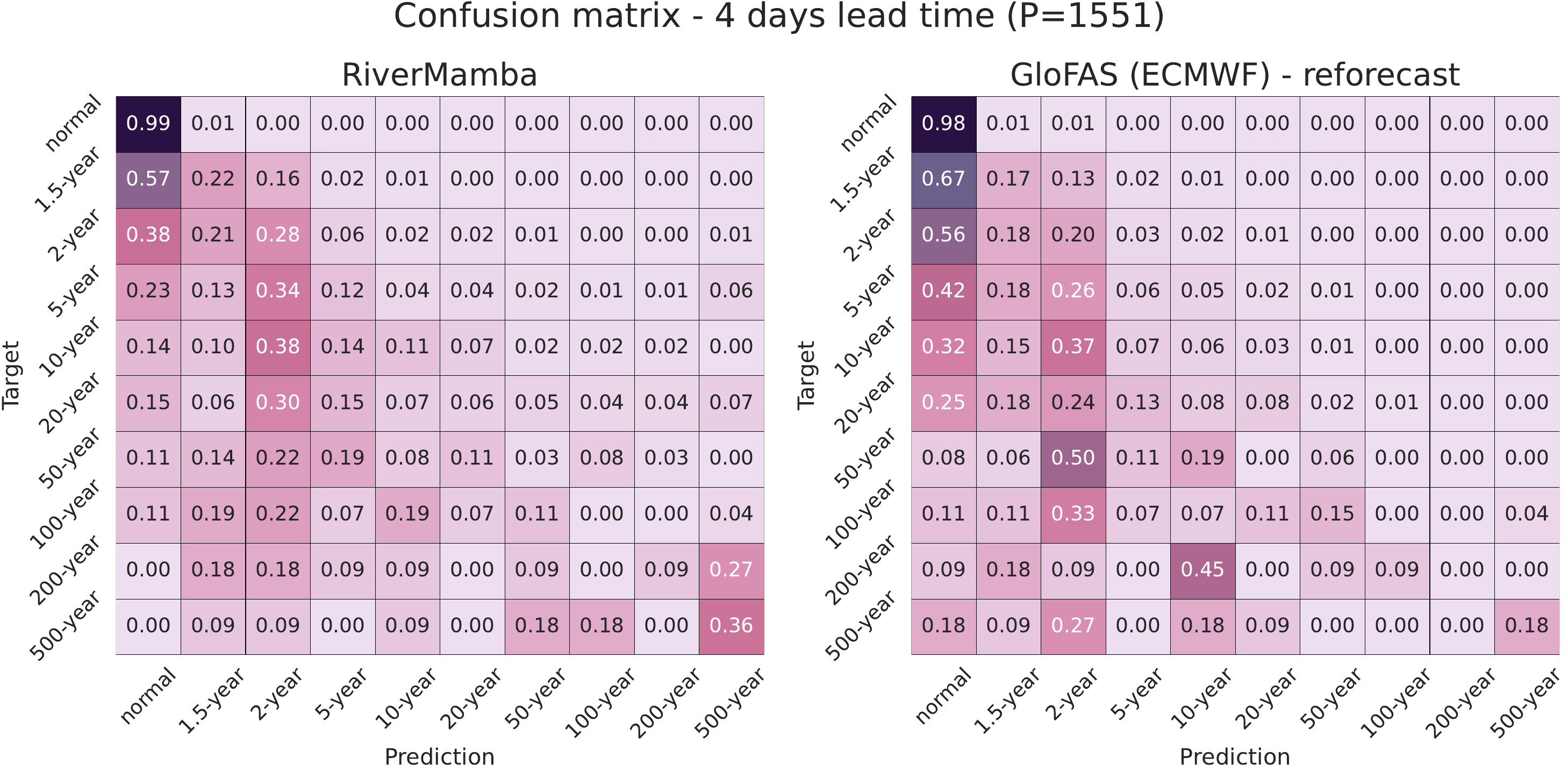}
  \caption{Comparison to GloFAS (ECMWF) - reforecast on GRDC observations (test set 2021-2023 temporally out-of-sample).}\label{fig:obs_confusion_matrix_ecmwf}
\end{figure}

\begin{figure}[!h]
  \centering
  \includegraphics[draft=\draft, width=.8\textwidth]{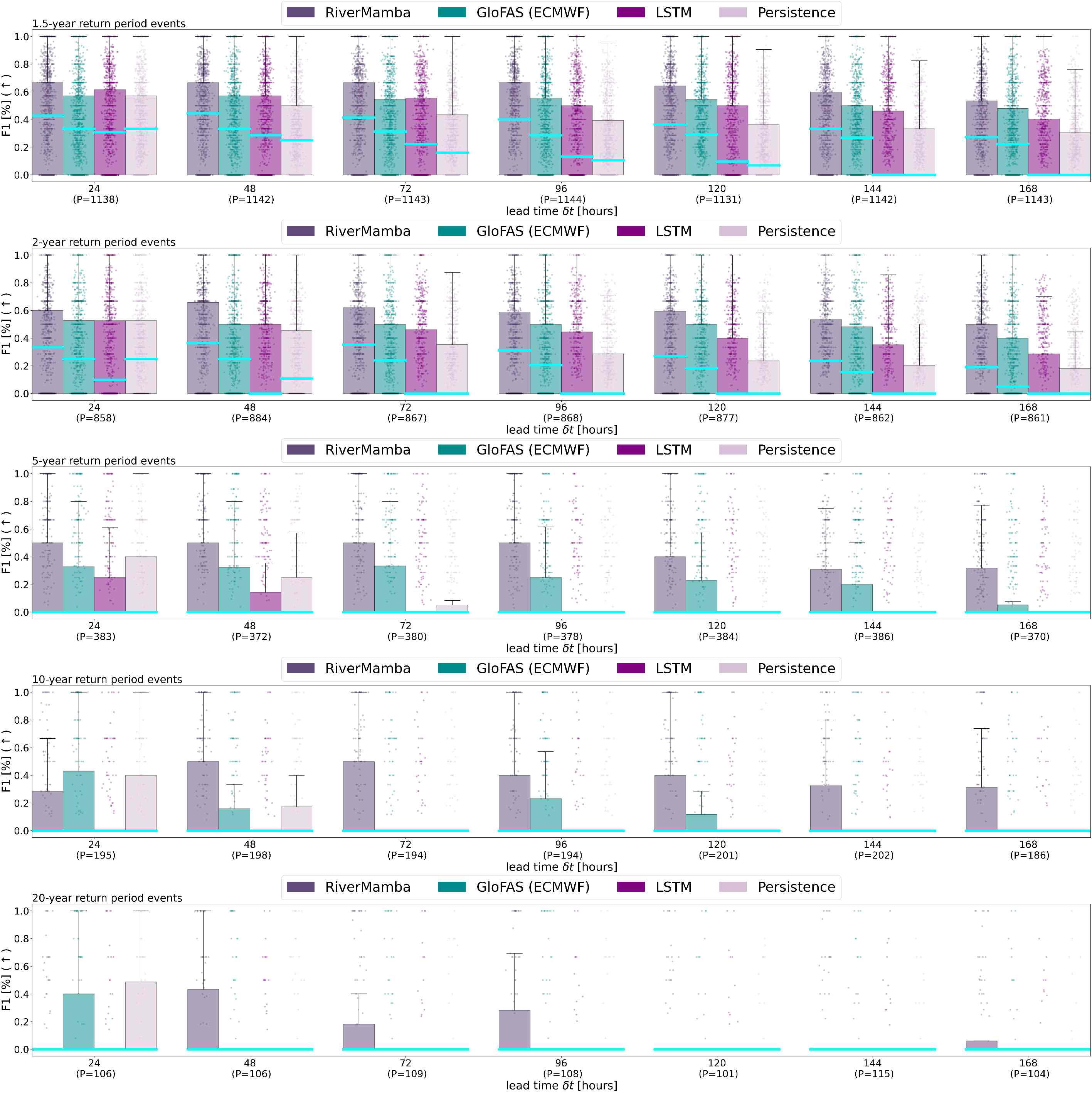}
  \caption{F1-score of flood forecasting for different return periods and lead time on GRDC observations (test set 2021-2023 temporally out-of-sample). Distribution quartiles are displayed in boxes, and the entire range excluding outliers is displayed in whiskers. The median score for the model is shown by the cyan line in the box.}\label{fig:obs_f1_cls}
\end{figure}

\begin{figure}[!h]
  \centering
  \includegraphics[width=.8\textwidth]{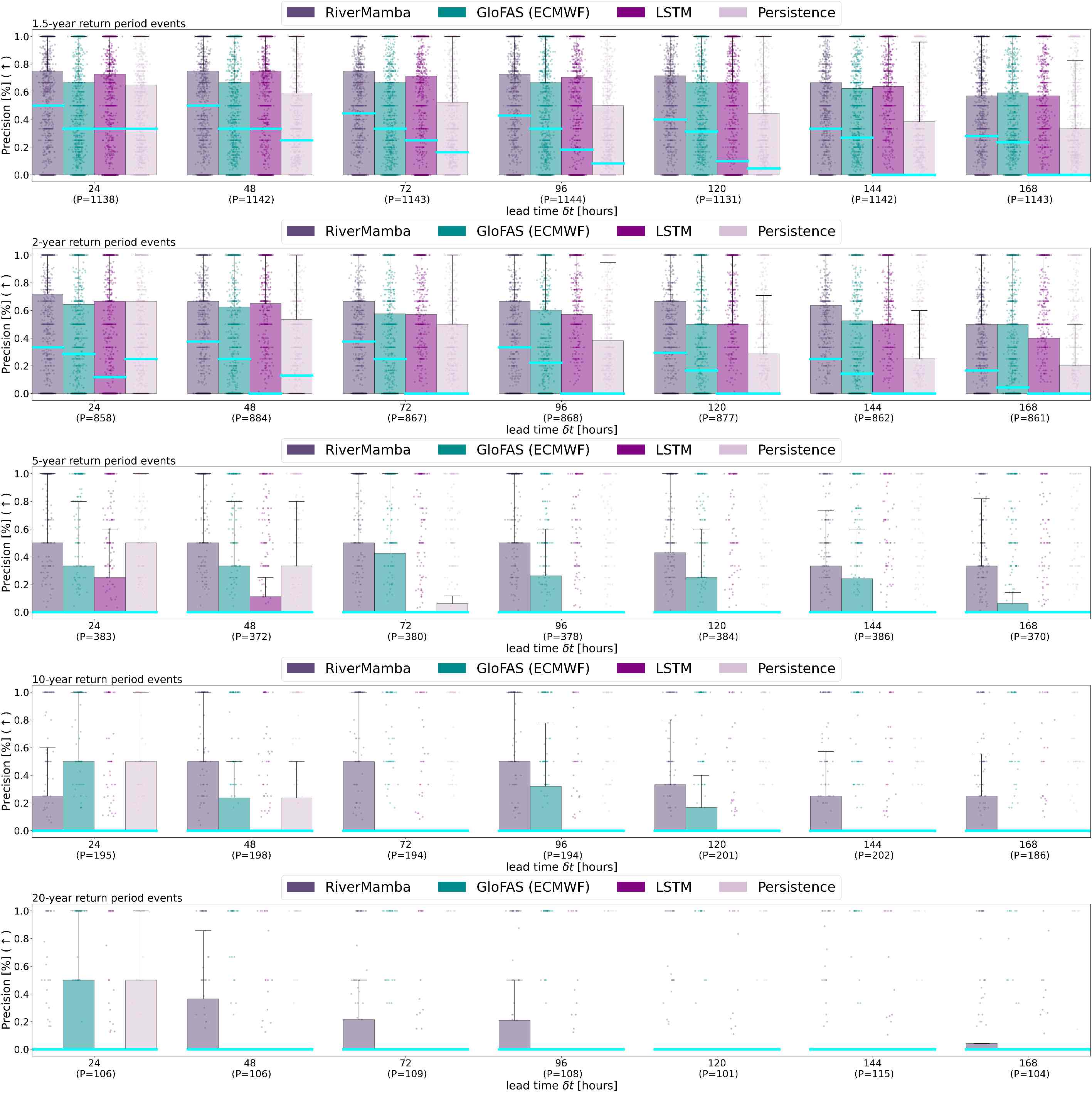}
  \caption{Precision of flood forecasting for different return periods and lead time on GRDC observations (test set 2021-2023 temporally out-of-sample). Distribution quartiles are displayed in boxes, and the entire range excluding outliers is displayed in whiskers. The median score for the model is shown by the cyan line in the box.}\label{fig:obs_precision_cls}
\end{figure}

\begin{figure}[!h]
  \centering
  \includegraphics[width=.8\textwidth]{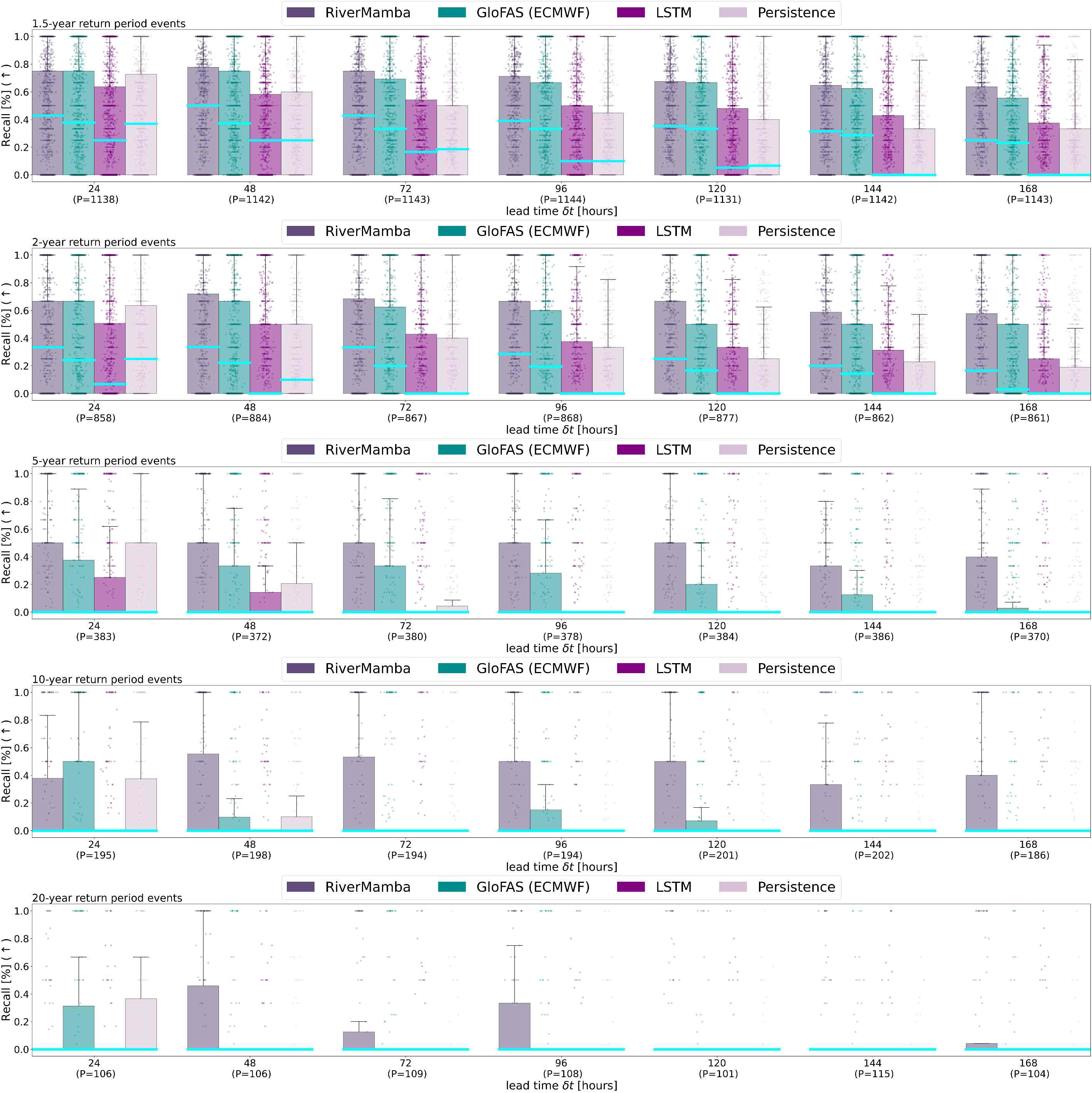}
  \caption{Recall of flood forecasting for different return periods and lead time on GRDC observations (test set 2021-2023 temporally out-of-sample). Distribution quartiles are displayed in boxes, and the entire range excluding outliers is displayed in whiskers. The median score for the model is shown by the cyan line in the box.}\label{fig:obs_recall_cls}
\end{figure}

\begin{figure}[!h]
  \centering
  \includegraphics[draft=\draft, width=.8\textwidth]{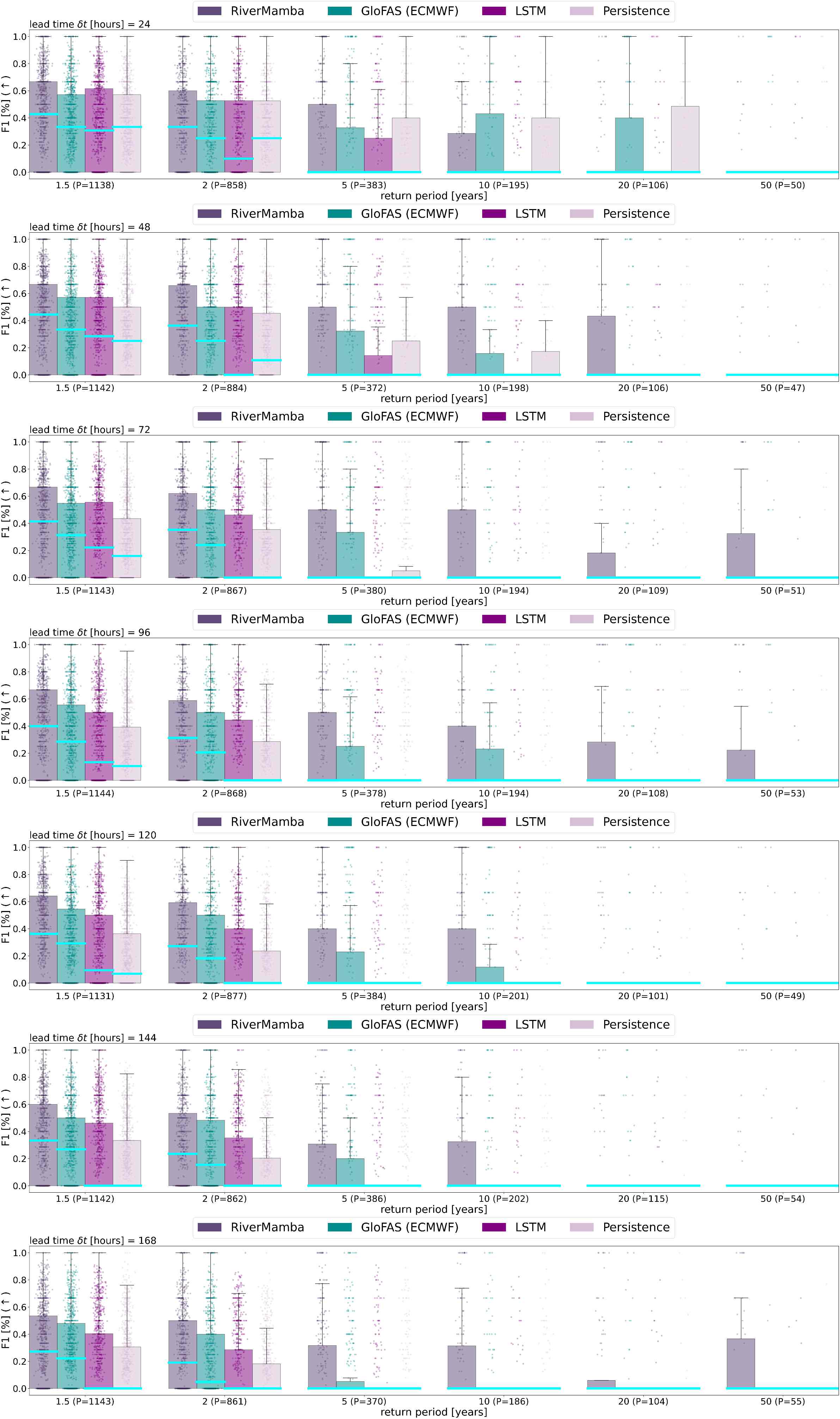}
  \caption{F1-score of flood forecasting for different lead time and return periods ($1.5$ - $50$ years) on GRDC observations (test set 2021-2023 temporally out-of-sample). Distribution quartiles are displayed in boxes, and the entire range excluding outliers is displayed in whiskers. The median score for the model is shown by the cyan line in the box.}\label{fig:obs_f1_time}
\end{figure}

\begin{figure}[!h]
  \centering
  \includegraphics[draft=\draft, width=.8\textwidth]{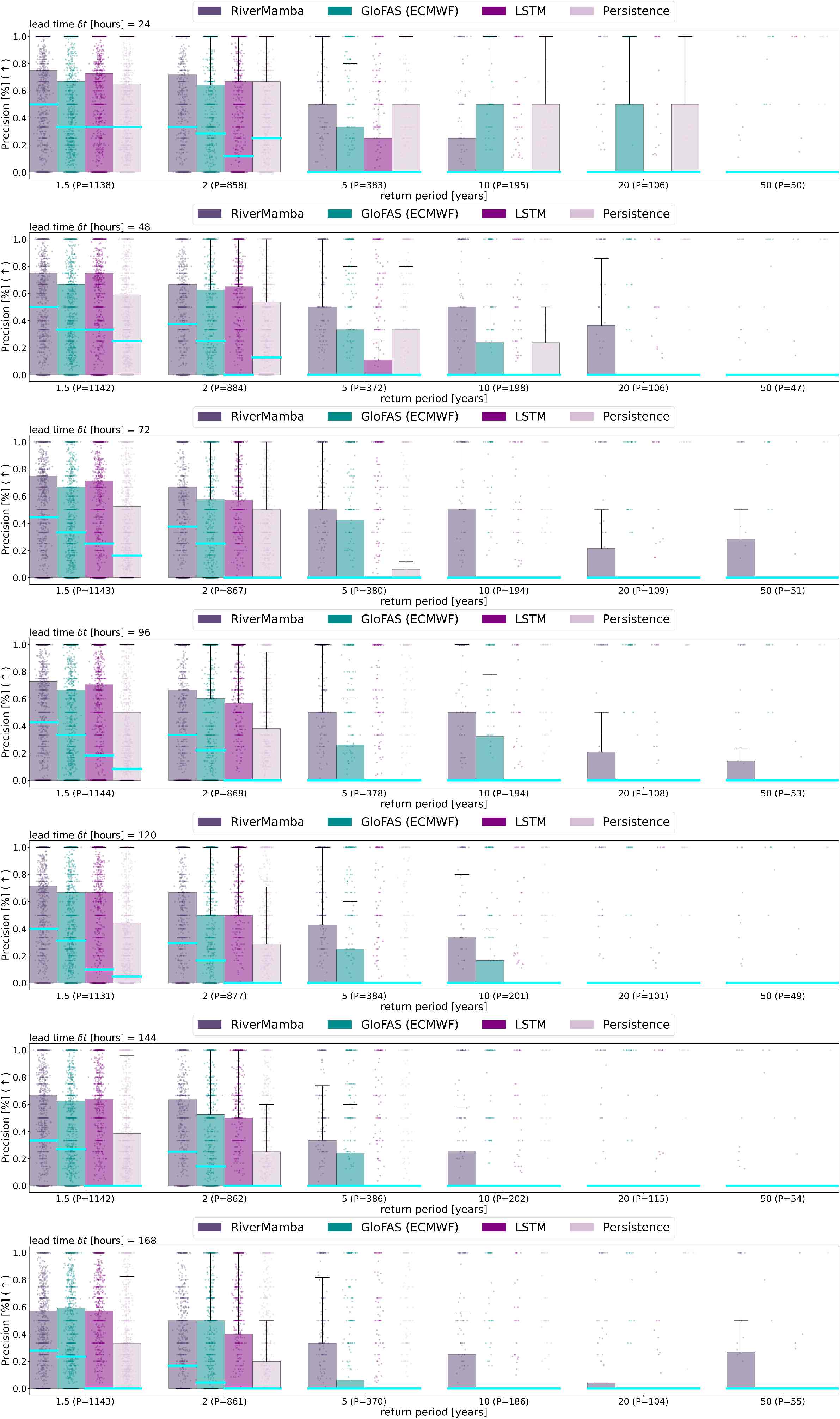}
  \caption{Precision of flood forecasting for different lead time and return periods ($1.5$ - $50$ years) on GRDC observations (test set 2021-2023 temporally out-of-sample). Distribution quartiles are displayed in boxes, and the entire range excluding outliers is displayed in whiskers. The median score for the model is shown by the cyan line in the box.}\label{fig:obs_precision_time}
\end{figure}

\begin{figure}[!h]
  \centering
  \includegraphics[width=.8\textwidth]{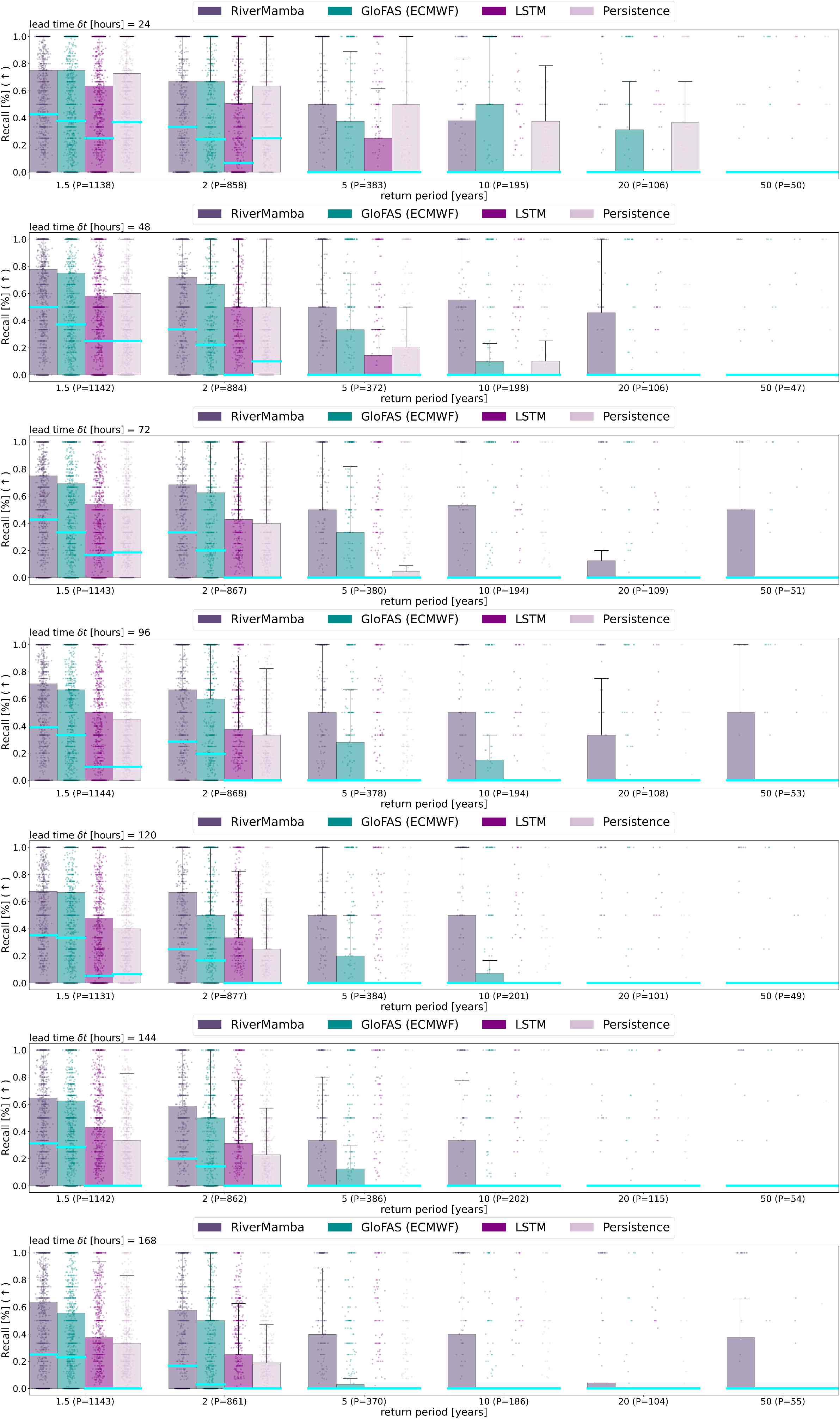}
  \caption{Recall of flood forecasting for different lead time and return periods ($1.5$ - $50$ years) on GRDC observations (test set 2021-2023 temporally out-of-sample). Distribution quartiles are displayed in boxes, and the entire range excluding outliers is displayed in whiskers. The median score for the model is shown by the cyan line in the box.}\label{fig:obs_recall_time}
\end{figure}

\begin{figure}[!h]
  \centering
  \includegraphics[width=.97\textwidth]{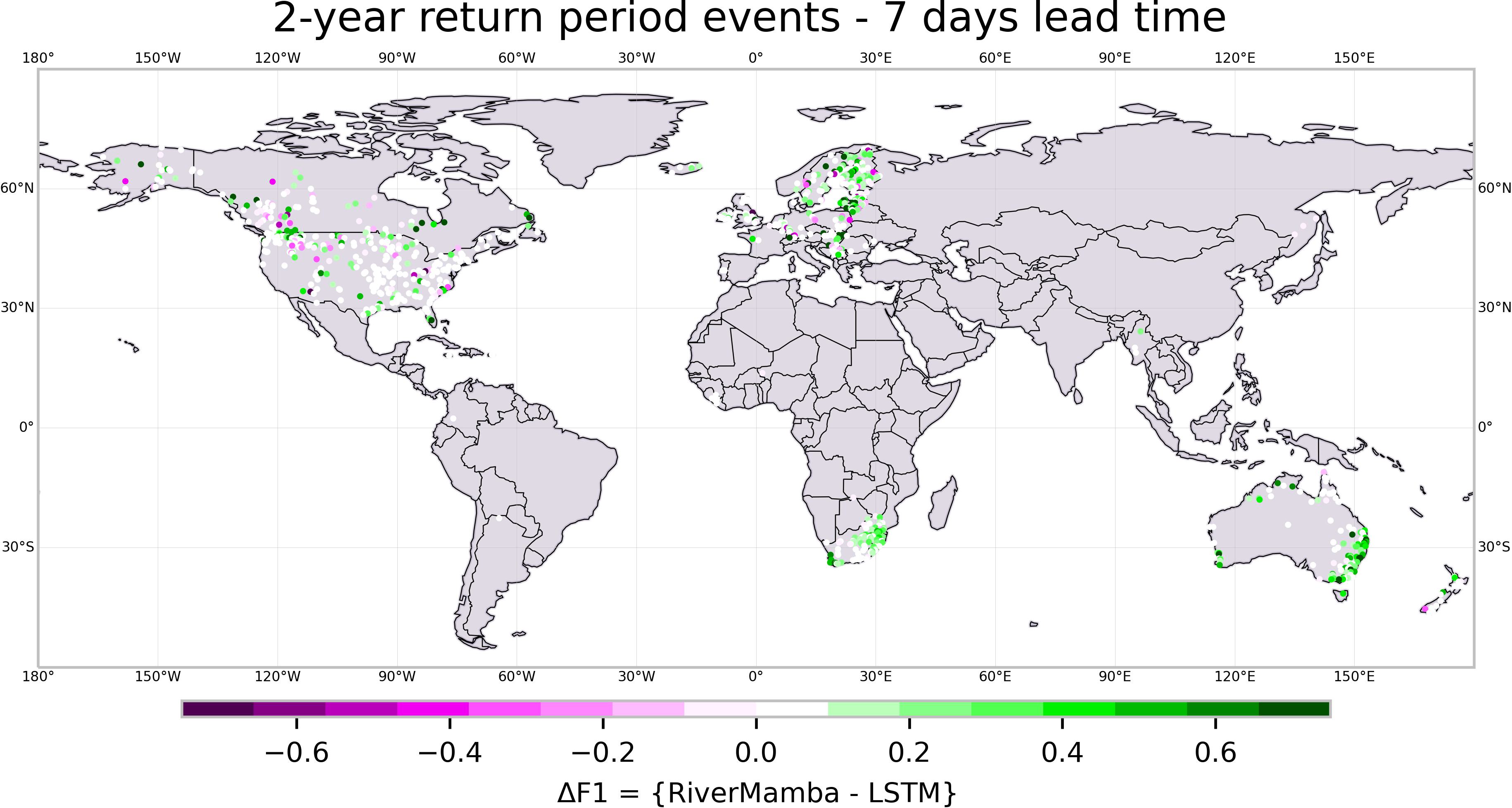}
  \caption{Comparison of F1-score between RiverMamba and LSTM on GRDC observations for the 2-year return period events (test set 2021-2023 temporally out-of-sample). RiverMamba improves over LSTM in 42\% of the stations (P=861) and is better or equally better in 86\% of the stations.}\label{fig:obs_delta_f1_lstm}
\end{figure}

\begin{figure}[!h]
  \centering
  \includegraphics[width=.97\textwidth]{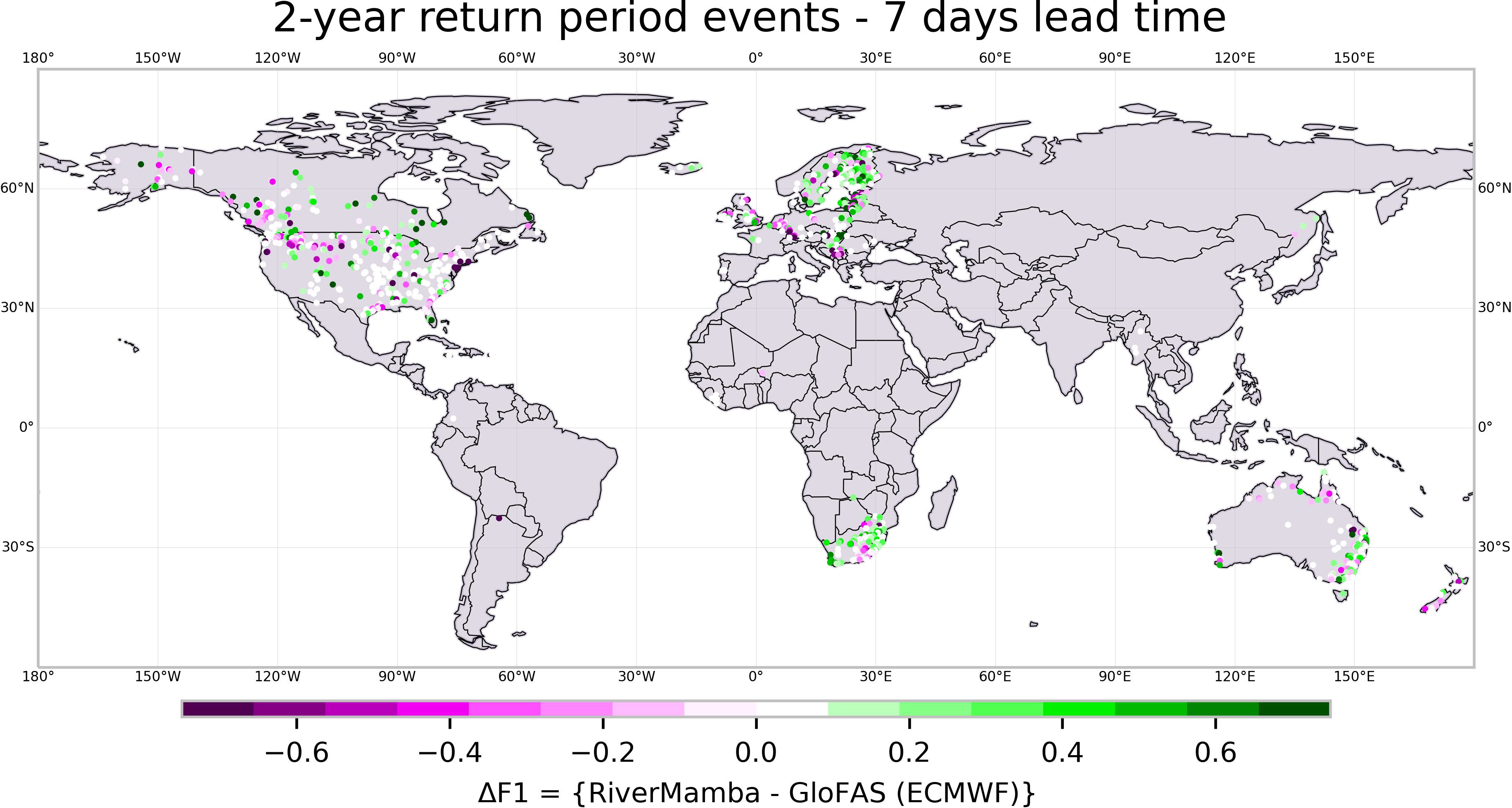}
  \caption{Comparison of F1-score between RiverMamba and GloFAS reforecast on GRDC observations for the 2-year return period events (test set 2021-2023 temporally out-of-sample). RiverMamba improves over GloFAS reforecast in 38\% of the stations (P=861) and is better or equally better in 73\% of the stations.}\label{fig:obs_delta_f1_glofas}
\end{figure}

\begin{figure}[!h]
  \centering
  \includegraphics[width=.97\textwidth]{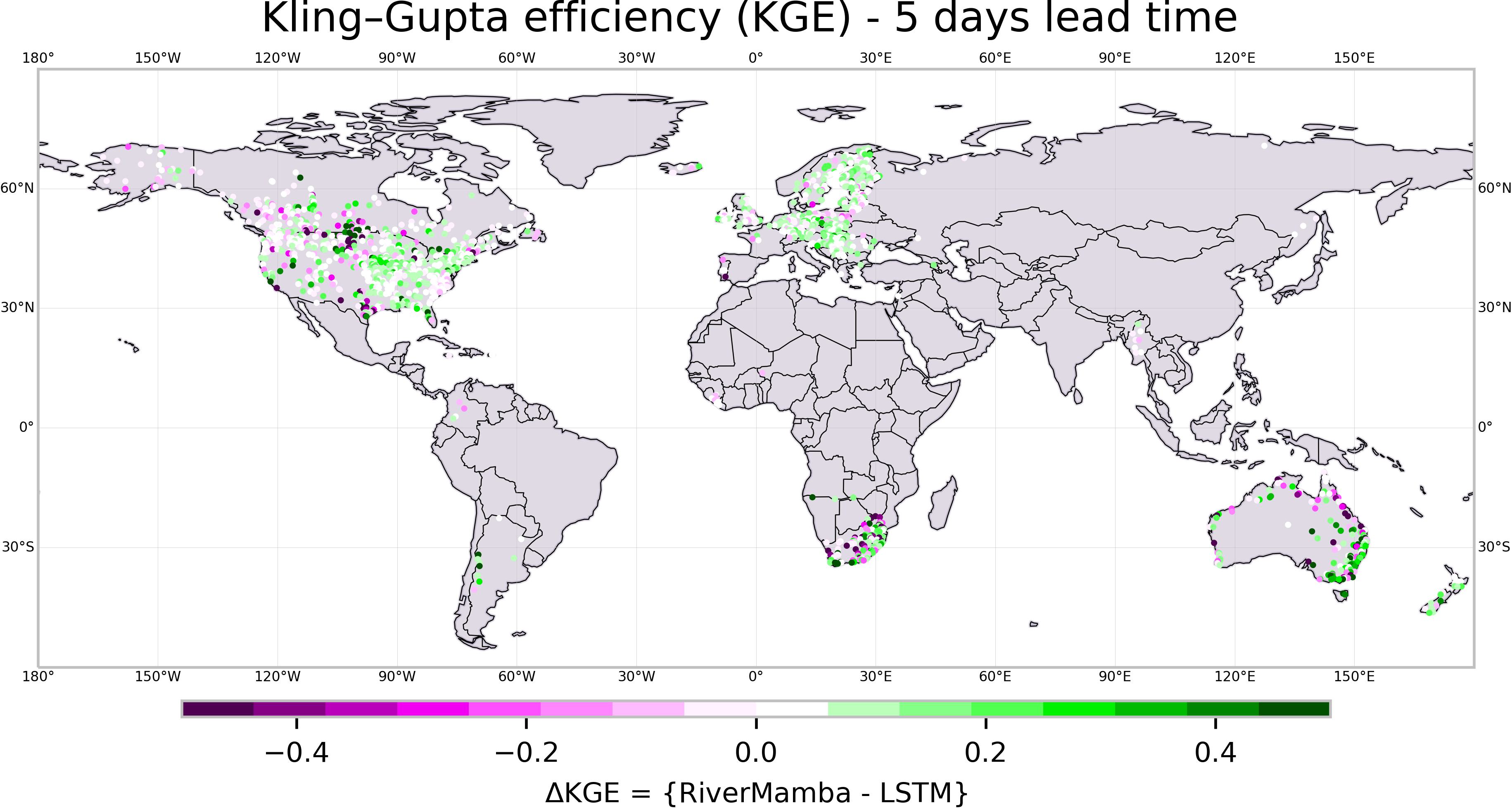}  \caption{Comparison of KGE between RiverMamba and LSTM on GloFAS reanalysis (test set 2021-2023 temporally out-of-sample). RiverMamba improves over LSTM in 68\% of the stations (P=1542).}\label{fig:obs_delta_kge_lstm}
\end{figure}

In this section, we plot additional results for the experiments on GRDC observational river discharge. In Figs.~\ref{fig:obs_MAE}-\ref{fig:obs_R}, we report the results for MAE, RMSE, R2, and R metrics with lead time. 
Figs.~\ref{fig:obs_confusion_matrix_lstm} and \ref{fig:obs_confusion_matrix_ecmwf} show the confusion matrix for RiverMamba, LSTM, and GloFAS.
In Figs.~\ref{fig:obs_f1_cls}-\ref{fig:obs_recall_time}, we report the results of F1-score, Precision, and Recall metrics for different return periods and lead times.
Finally, in Figs.~\ref{fig:obs_delta_f1_lstm}-\ref{fig:obs_delta_kge_lstm}, we compare the results between RiverMamba, LSTM, and GloFAS for F1-score and KGE metrics spatially.

\clearpage

\subsection{Comparison to operational GloFAS forecsating on gauged GRDC}
\label{sec:Appendix_comparison_glofas_forecast}

Here, we compare to the archival operational forecast from GloFAS ECMWF (Sec.~\ref{sec:Appendix_baselines_glofas_forecast}). This is different from the GloFAS reforecast and gives a more realistic assessment of the physics-based model.

\begin{table}[!h]
  \caption{Results on GRDC gauged stations. \gray{($\pm$)} denotes the standard deviation for 3 runs.}
  \label{table:Appendix_comparison_glofas_forecast}
  \centering
  \tabcolsep=3.5pt\relax
  \setlength\extrarowheight{0pt}
  \begin{tabular}{r *{5}l}
    \toprule
    & & \multicolumn{3}{c}{Test (2023-2024)} \\
    \midrule
    Model & & MAE ($\downarrow$) & R2 ($\uparrow$) & KGE ($\uparrow$) & F1-score ($\uparrow$) \\
   \midrule
     GloFAS$^*$ & & \colorm{65.35} &  \colorm{-0.0063} & \colorm{0.0439} & \colorf{0.1795}\\
     LSTM & & \colorm{53.07}{\small\gray{$\pm$0.39}} &  \colorm{-0.0005}{\small\gray{$\pm$0.0001}} &  \colorm{0.3147}{\small\gray{$\pm$0.0026}} & \colorf{0.1120}{\small\gray{$\pm$0.0065}} \\
   \midrule
    \bestc RiverMamba & \bestc & \bestc  \textbf{\colorm{49.32}}{\small\gray{$\pm$0.18}} & \bestc \textbf{\colorm{0.0001}}{\small\gray{$\pm$0.0001}} & \bestc \textbf{\colorm{0.3821}}{\small\gray{$\pm$0.0076}} & \bestc \textbf{\colorf{0.2358}}{\small\gray{$\pm$0.0211}} \\
    \bottomrule
    \multicolumn{5}{l}{\footnotesize{$^*$GloFAS operational forecast \cite{GloFAS_Forecast}}} \\
\end{tabular}
\end{table}

\clearpage

\section{Case studies of extreme flood events}
\label{sec:Appendix_case_study}

\subsection{2021 Western Europe flood}

\begin{figure}[!h]
  \centering
  \includegraphics[width=.99\textwidth]{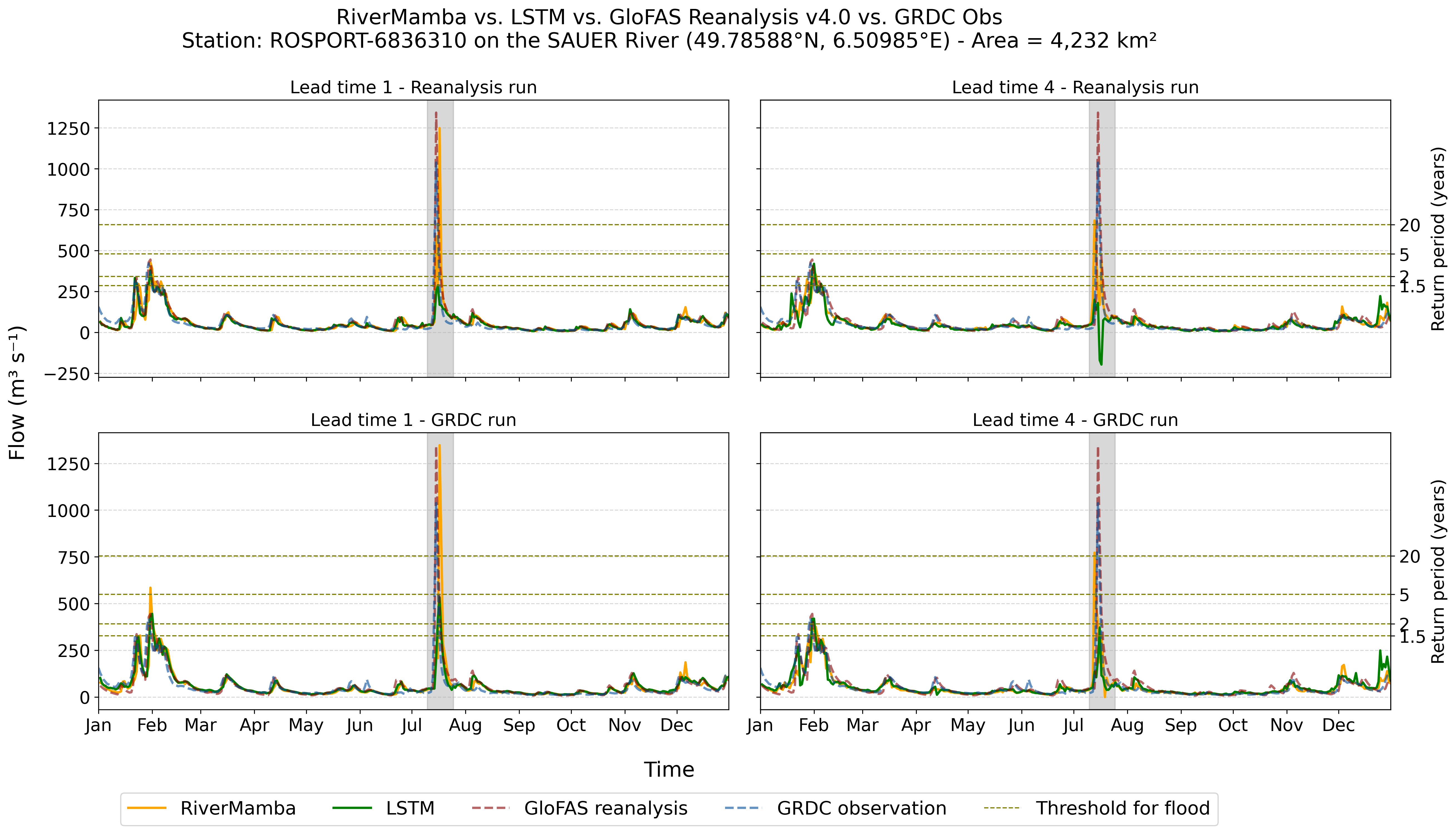}
  \caption{River discharge of the Sauer river in 2021 for RiverMamba (orange), LSTM (green), GloFAS reanalysis (dashed red), and GRDC observation (dashed green). The grey shaded area highlights the 2021 Germany flood between July 10 to July 20. The olive dashed lines represent the 1.5, 2, 5, and 20-year return periods calculated over reanalysis and GRDC observation data, respectively. }\label{fig:all_forecast_2021}
\end{figure}

\begin{figure}[!h]
  \centering
  \includegraphics[width=.95\textwidth]{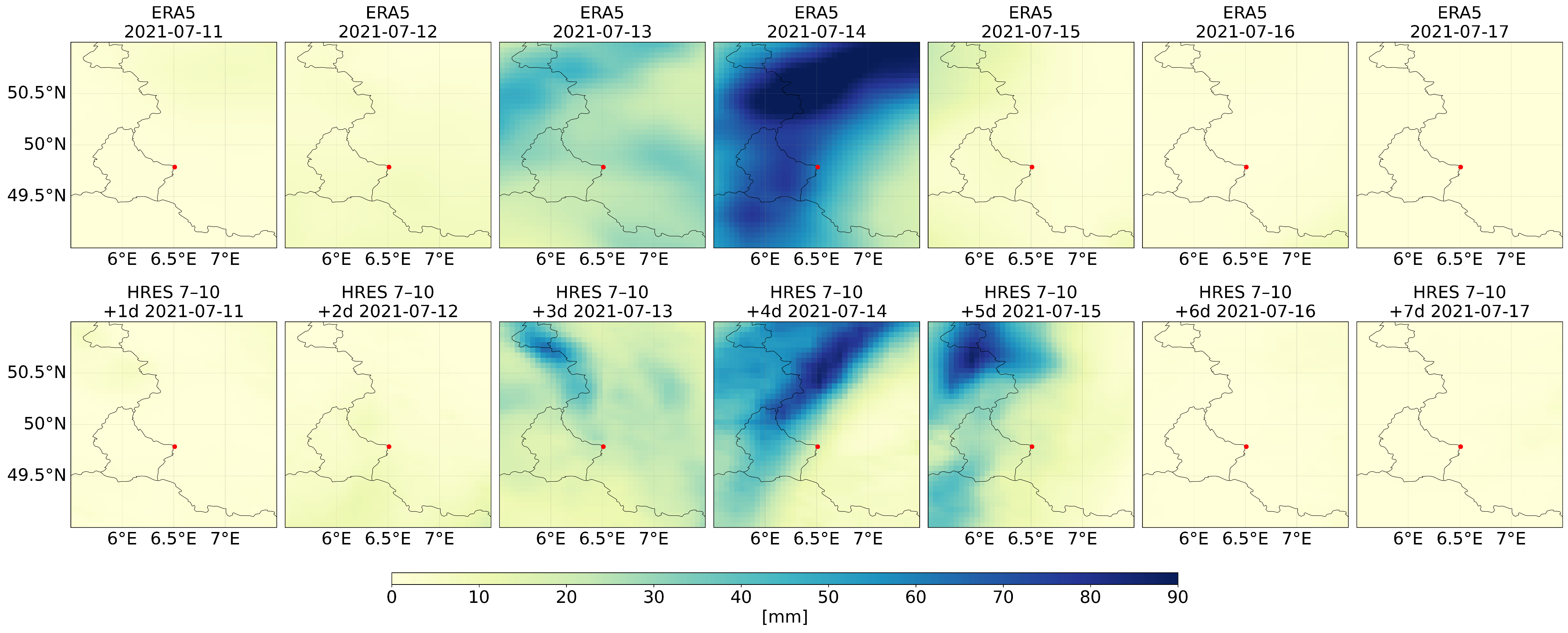}
  \caption{Daily total precipitation from 2021-07-11 to 2021-07-17 during the 2021 flood event in the target domain. First row shows precipitation as simulated by ERA5 Reanalysis and second row is the ECMWF HRES forecast issued at 2021-07-10 with 7-day lead time. The red dot is the location of the river discharge gauge station.}\label{fig:era5_hres_tp}
\end{figure}

In this section, we present the daily river discharge at a gauge station on the Sauer River—located at the border of Germany, France, and Luxembourg—for the year 2021. Shown is the river discharge signal as predicted by RiverMamba, LSTM, and GloFAS reanalysis, and compared against GRDC observations. Particular attention is given to the extreme flood event in July 2021 \cite{najafi2024high, Magdalena_2025}, highlighted by the grey-shaded area in Fig.~\ref{fig:all_forecast_2021}. To illustrate the meteorological drivers of this flood, we also show 7-day precipitation from ERA5 reanalysis and ECMWF HRES forecasts in Fig.~\ref{fig:era5_hres_tp}.

\clearpage

\subsection{2024 Southeast Europe floods}

In the following sections, we compare the daily flood severity map from GloFAS reanalysis as a ground truth and RiverMamba model at big flood events with different causes in 2024 and from different places around the Earth. These maps are usually used in the operational flood forecast service like GloFAS to provide a quick overview of the ongoing and upcoming flood events. As shown from these flood severity maps, RiverMamba can provide useful flood risk information at high spatial resolution to support decision-making. To the best of our knowledge, RiverMamba is the first AI model to demonstrate strong performance in predicting flood return periods globally under varying climate conditions at 5 km resolution across Europe, USA, Africa, and China. This further demonstrates its potential as a valuable component of operational flood early warning systems. It is important to note that the quality of the ECMWF HRES data driving the forecast is a key factor influencing RiverMamba performance.

\begin{figure}[!h]
  \centering
  \includegraphics[width=.97\textwidth]{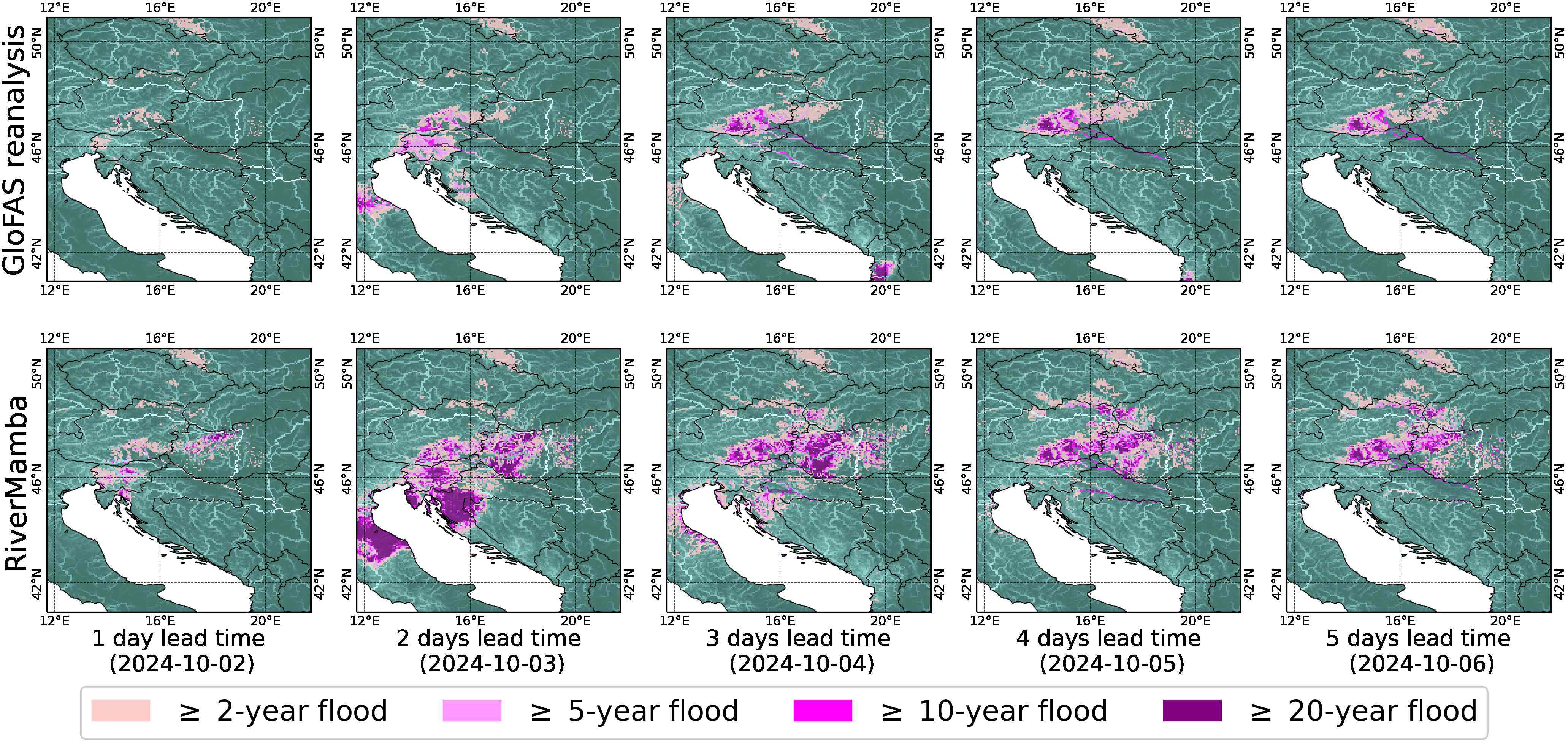}
  \caption{Comparison between GloFAS reanalysis (top, used as reference) and RiverMamba forecast (bottom) during the Southeast European flood in October 2024. Shown are flood severity maps at 5-km resolution where each panel shows flood extent at different lead times from 1 to 5 days.}\label{fig:2024_Southeast_Europe_floods}
\end{figure}

\newpage

\subsection{2024 Central European floods}


\begin{figure}[!h]
  \centering
  \includegraphics[width=.97\textwidth]{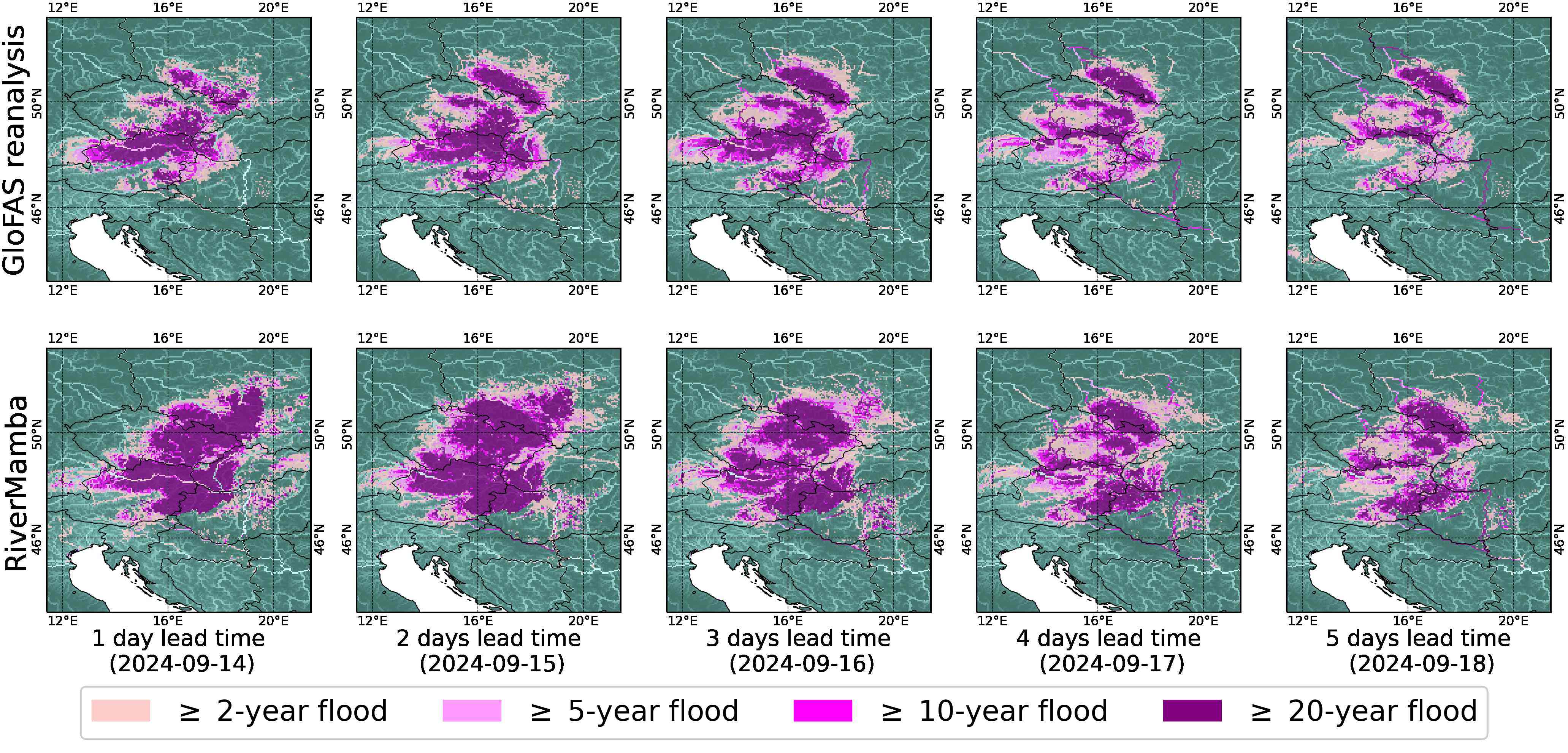}
  \caption{Comparison between GloFAS reanalysis (top, used as reference) and RiverMamba forecast (bottom) during the Central European floods event in 2024 \protect\cite{Austria_flood_2024}. In September 2024, Storm Boris brought record-breaking rainfall to Central Europe, causing devastating floods across Austria, the Czech Republic, Poland, Romania, Slovakia, Germany, and Hungary. Studies indicate that climate change doubled the likelihood and increased the intensity of such extreme rainfall events, highlighting the growing impact of global warming on severe weather patterns. Shown are flood severity maps at 5-km resolution where each panel shows flood extent at different lead times from 1 to 5 days.}\label{fig:2024_Central_European_floods}
\end{figure}

\subsection{2024 Spanish floods}


\begin{figure}[!h]
  \centering
  \includegraphics[width=.97\textwidth]{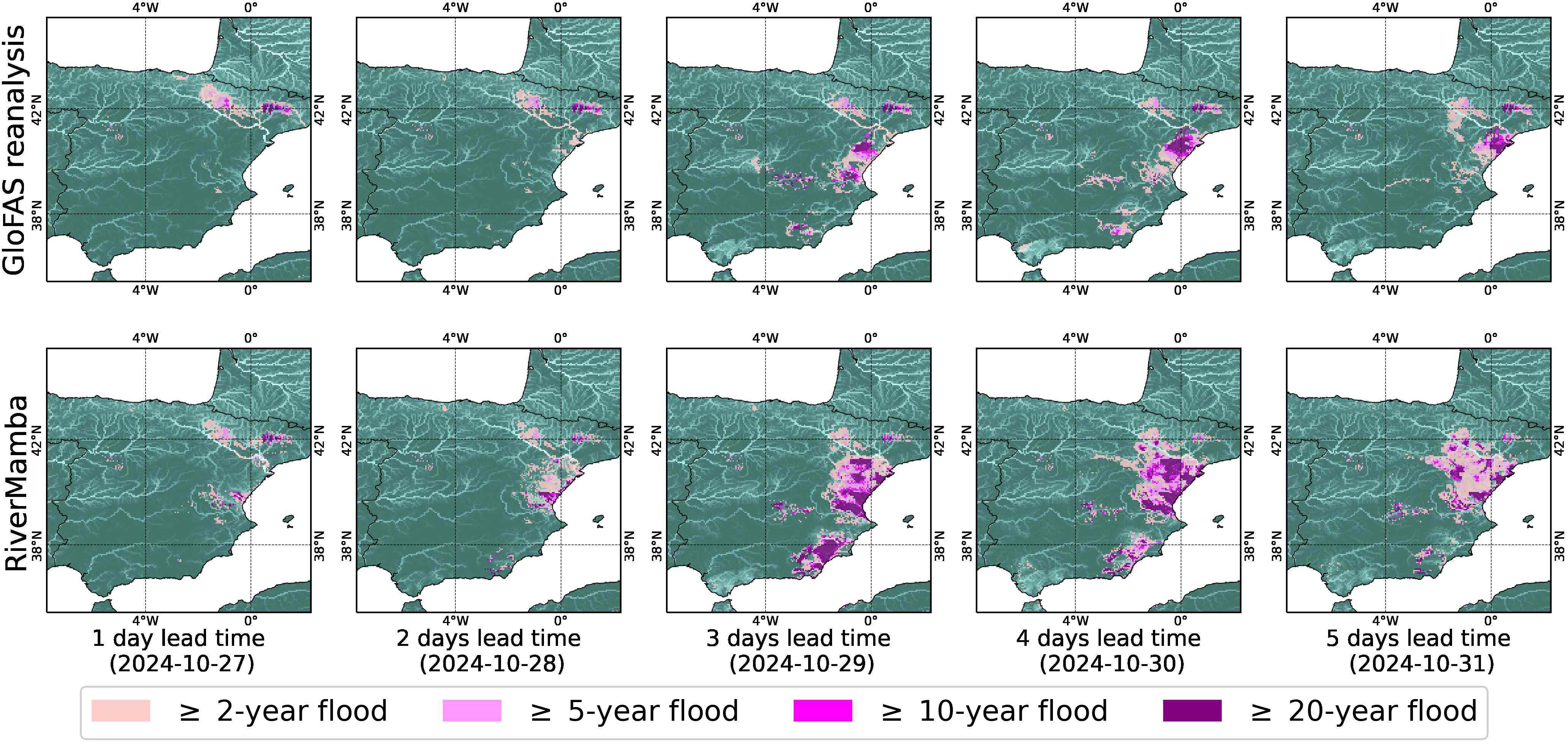}
  \caption{Comparison between GloFAS reanalysis (top, used as reference) and RiverMamba forecast (bottom) during the Spanish flood event in October 2024.
  The flood event primarily affecting the Valencia region, was caused by a cold drop (DANA) weather system intensified by climate change. Shown are flood severity maps at 5-km resolution where each panel shows flood extent at different lead times from 1 to 5 days.}\label{fig:2024_Spanish_floods}
\end{figure}

\newpage

\subsection{2024 Saarland Germany flood}


\begin{figure}[!h]
  \centering
  \includegraphics[width=.97\textwidth]{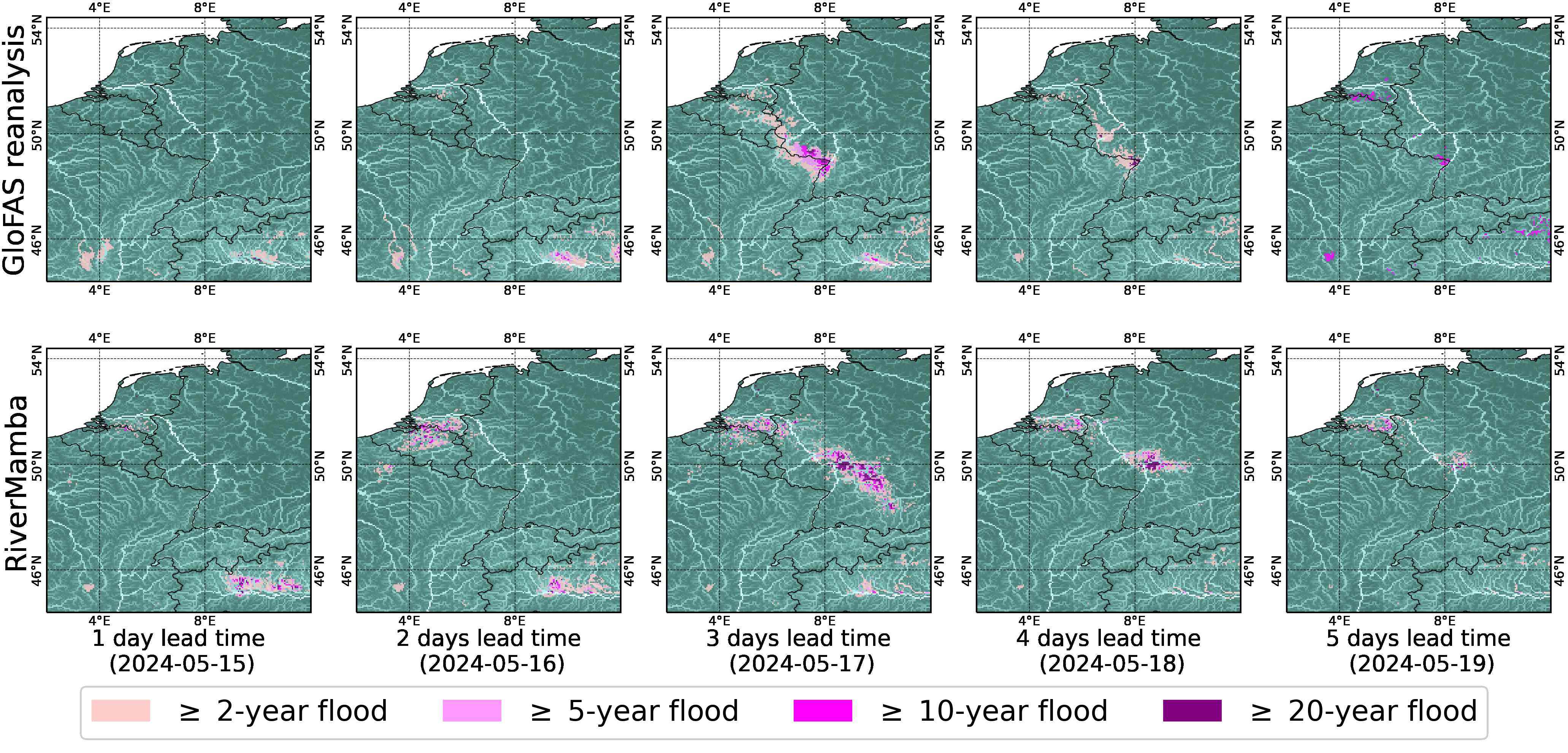}
  \caption{Comparison between GloFAS reanalysis (top, used as reference) and RiverMamba forecast (bottom) during the Saarland Germany flood event in May 2024. It was caused by thunderstorms and extreme rainfall, resulting in deadly floods and landslides across Saarland and Rheinland-Pfalz. However, RiverMamba predicted this flood event at a nearby location in Saarland, and the reason could be attributed to the inaccurate real-time ECMWF HRES forecast that drives the flood forecasting. Shown are flood severity maps at 5-km resolution where each panel shows flood extent at different lead times from 1 to 5 days.}\label{fig:2024_Saarland_Germany_floods}
\end{figure}

\subsection{2024 Kenya-Tanzania flood}


\begin{figure}[!h]
  \centering
  \includegraphics[width=.97\textwidth]{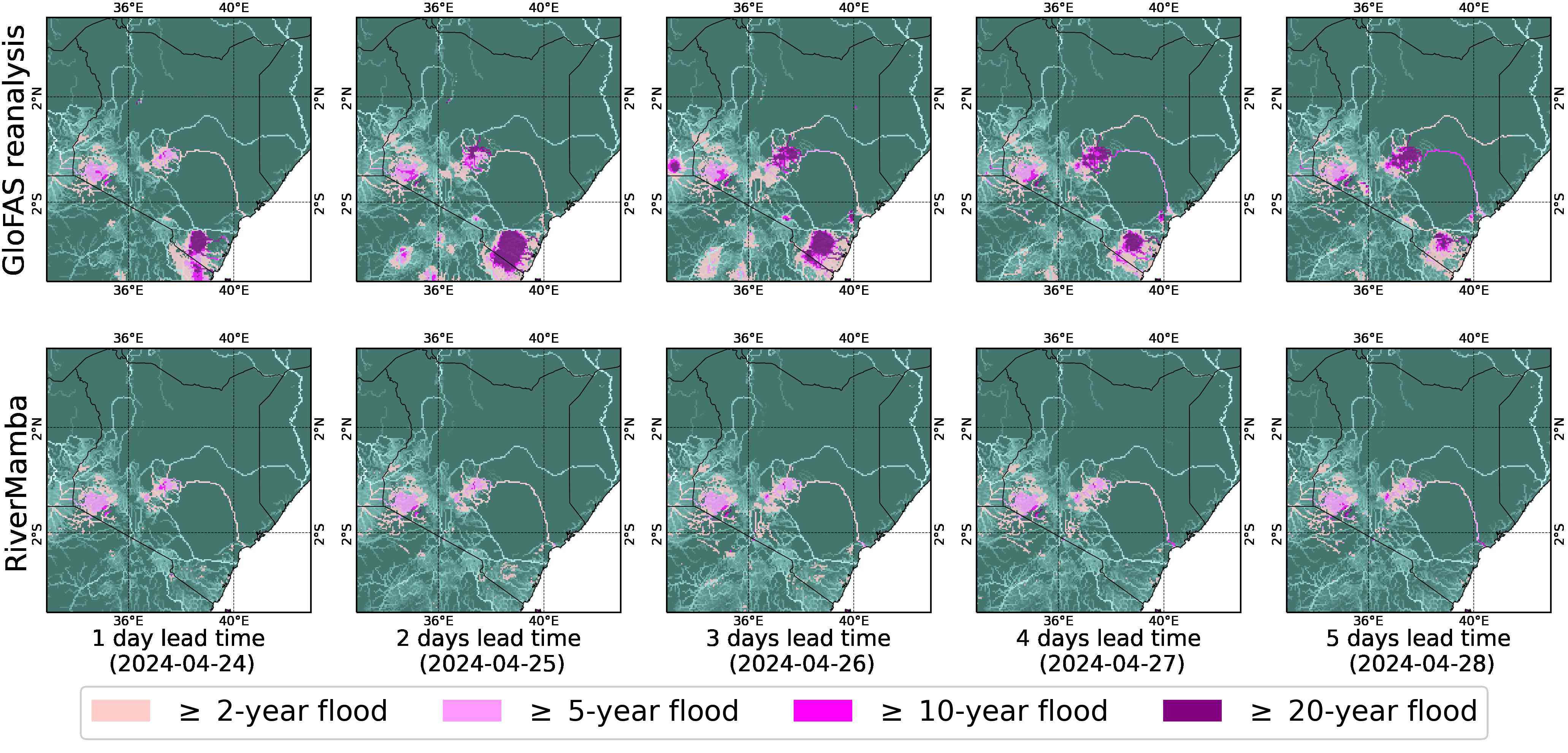}
  \caption{Comparison between GloFAS reanalysis (top, used as reference) and RiverMamba forecast (bottom) during the Kenya-Tanzania flood event in April 2024. This was the consequence of a combination of El Niño and a positive Indian Ocean Dipole, resulting in deaths, widespread displacement, and significant infrastructure damage.
  Shown are flood severity maps at 5-km resolution where each panel shows flood extent at different lead times from 1 to 5 days. 
  Here points with less than 1 m$^3$/s discharge have been removed to erase artifacts on desert grids.}\label{fig:2024_Kenya_Tanzania_floods}
\end{figure}

\newpage

\subsection{2024 California flood}

\begin{figure}[!h]
  \centering
  \includegraphics[width=.97\textwidth]{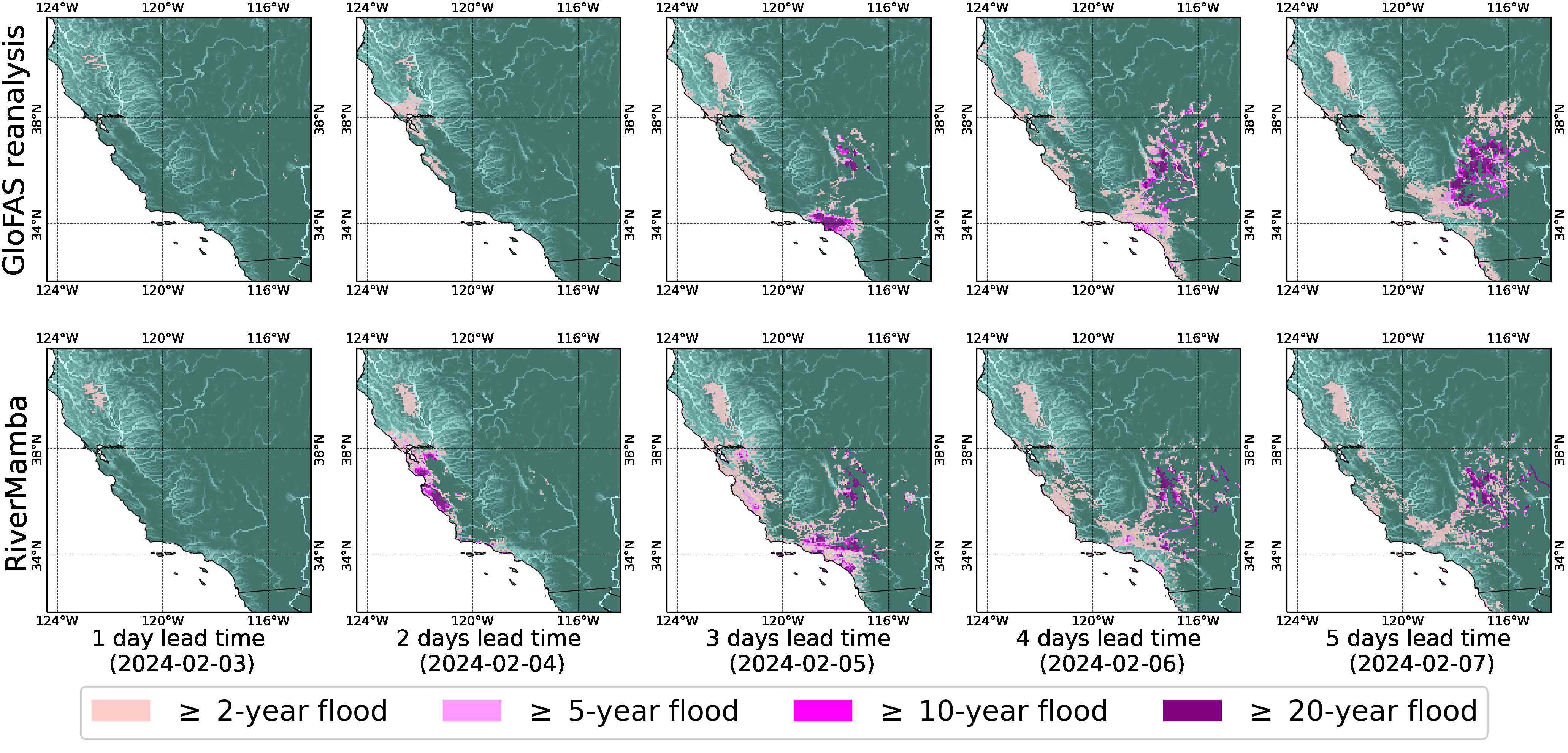}
  \caption{Comparison between GloFAS reanalysis (top, used as reference) and RiverMamba forecast (bottom) during the California, USA flood in February 2024. It was caused by two powerful atmospheric rivers.
  Shown are flood severity maps at 5-km resolution where each panel shows flood extent at different lead times from 1 to 5 days. 
  Here points with less than 0.01 m$^3$/s discharge have been removed to erase artifacts on desert grids.}\label{fig:2024_California_atmospheric_rivers}
\end{figure}

\subsection{2024 Central-South China floods}


\begin{figure}[!h]
  \centering
  \includegraphics[draft=\draft, width=.97\textwidth]{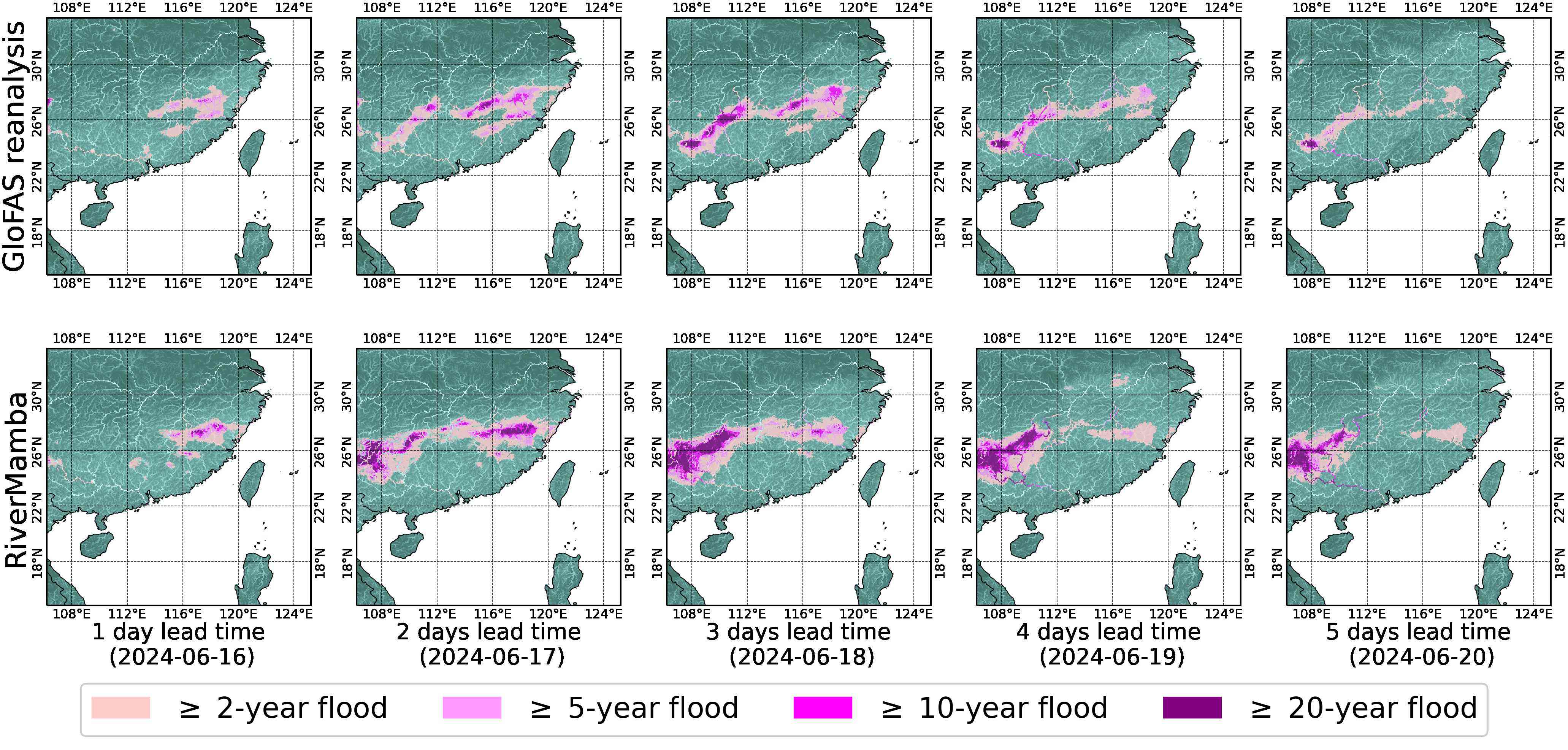}
\caption{Comparison between GloFAS reanalysis (top, used as reference) and RiverMamba forecast (bottom) during the Central-South China flood event in June 2024 due to unprecedented rainfall. Shown are flood severity maps at 5-km resolution where each panel shows flood extent at different lead times from 1 to 5 days.}\label{fig:2024_China_floods}
\end{figure}

\clearpage

\section{Code and data availability}
\label{sec:code_data_availability}


The code of RiverMamba and processing scripts are available on GitHub at \colorh{\url{https://github.com/HakamShams/RiverMamba_code}}.
The pre-processed data used in this study are available at  \colorh{\url{https://doi.org/10.60507/FK2/T8QYWE}} \cite{RiverMamba_dataset}.
GRDC data that has been used in this study is available for researchers after signing a license agreement with the owner of the data. 
Instructions on how the data can be obtained and used are provided in the source code.    

\section{Broader impacts}
\label{sec:broader_impacts}

Extreme flood events, characterized by longer return periods, are expected to become more frequent and intense due to climate change. Traditional hydrology models often struggle to accurately predict such events. To improve flood detection, it is crucial to develop computationally efficient and precise deep learning models capable of forecasting key hydrological variables, such as river discharge.
In this study, we demonstrated the potential of RiverMamba for predicting extreme riverine floods and in Appendix Sec.~\ref{sec:Appendix_case_study}, we presented case studies for extreme flood events.
We see this as the primary motivation for developing RiverMamba. However, it is important to acknowledge that early warning systems may sometimes fail, leading to inaccurate forecasts. This limitation should be considered when deploying early warning systems.



{
\small
\bibliographystyle{unsrt}
\bibliography{ref}

\begin{thebibliography}{100}

\bibitem{dottori2018increased}
Francesco Dottori, Wojciech Szewczyk, Juan-Carlos Ciscar, Fang Zhao, Lorenzo Alfieri, Yukiko Hirabayashi, Alessandra Bianchi, Ignazio Mongelli, Katja Frieler, Richard~A Betts, et~al.
\newblock Increased human and economic losses from river flooding with anthropogenic warming.
\newblock {\em Nature Climate Change}, 8(9):781--786, 2018.

\bibitem{merz2021causes}
Bruno Merz, G{\"u}nter Bl{\"o}schl, Sergiy Vorogushyn, Francesco Dottori, Jeroen C. J.~H. Aerts, Paul Bates, Miriam Bertola, Matthias Kemter, Heidi Kreibich, Upmanu Lall, et~al.
\newblock Causes, impacts and patterns of disastrous river floods.
\newblock {\em Nature Reviews Earth \& Environment}, 2(9):592--609, 2021.

\bibitem{kreibich2022challenge}
Heidi Kreibich, Anne~F Van~Loon, Kai Schr{\"o}ter, Philip~J Ward, Maurizio Mazzoleni, Nivedita Sairam, Guta~Wakbulcho Abeshu, Svetlana Agafonova, Amir AghaKouchak, Hafzullah Aksoy, et~al.
\newblock The challenge of unprecedented floods and droughts in risk management.
\newblock {\em Nature}, 608(7921):80--86, 2022.

\bibitem{rentschler2022flood}
Jun Rentschler, Melda Salhab, and Bramka~Arga Jafino.
\newblock Flood exposure and poverty in 188 countries.
\newblock {\em Nature communications}, 13(1):3527, 2022.

\bibitem{Myanna_2022}
Myanna Lahsen and Jesse Ribot.
\newblock Politics of attributing extreme events and disasters to climate change.
\newblock {\em WIREs Climate Change}, 13(1):e750, 2022.

\bibitem{Shijie_2022}
S.~Jiang, E.~Bevacqua, and J.~Zscheischler.
\newblock River flooding mechanisms and their changes in europe revealed by explainable machine learning.
\newblock {\em Hydrology and Earth System Sciences}, 26(24):6339--6359, 2022.

\bibitem{Shijie_2024}
Shijie Jiang, Larisa Tarasova, Guo Yu, and Jakob Zscheischler.
\newblock Compounding effects in flood drivers challenge estimates of extreme river floods.
\newblock {\em Science Advances}, 10(13):eadl4005, 2024.

\bibitem{PAPPENBERGER2015278}
Florian Pappenberger, Hannah~L. Cloke, Dennis~J. Parker, Fredrik Wetterhall, David~S. Richardson, and Jutta Thielen.
\newblock The monetary benefit of early flood warnings in europe.
\newblock {\em Environmental Science \& Policy}, 51:278--291, 2015.

\bibitem{camps2025artificial}
Gustau Camps-Valls, Miguel-{\'A}ngel Fern{\'a}ndez-Torres, Kai-Hendrik Cohrs, Adrian H{\"o}hl, Andrea Castelletti, Aytac Pacal, Claire Robin, Francesco Martinuzzi, Ioannis Papoutsis, Ioannis Prapas, et~al.
\newblock Artificial intelligence for modeling and understanding extreme weather and climate events.
\newblock {\em Nature Communications}, 16(1):1919, 2025.

\bibitem{Emerton_2016}
Rebecca~E. Emerton, Elisabeth~M. Stephens, Florian Pappenberger, Thomas~C. Pagano, Albrecht~H. Weerts, Andy~W. Wood, Peter Salamon, James~D. Brown, Niclas Hjerdt, Chantal Donnelly, Calum~A. Baugh, and Hannah~L. Cloke.
\newblock Continental and global scale flood forecasting systems.
\newblock {\em WIREs Water}, 3(3):391--418, 2016.

\bibitem{HyFS}
H.~A.~P. Hapuarachchi, M.~A. Bari, A.~Kabir, M.~M. Hasan, F.~M. Woldemeskel, N.~Gamage, P.~D. Sunter, X.~S. Zhang, D.~E. Robertson, J.~C. Bennett, and P.~M. Feikema.
\newblock Development of a national 7-day ensemble streamflow forecasting service for australia.
\newblock {\em Hydrology and Earth System Sciences}, 26(18):4801--4821, 2022.

\bibitem{EFAS}
F.~Dottori, M.~Kalas, P.~Salamon, A.~Bianchi, L.~Alfieri, and L.~Feyen.
\newblock An operational procedure for rapid flood risk assessment in europe.
\newblock {\em Natural Hazards and Earth System Sciences}, 17(7):1111--1126, 2017.

\bibitem{Nevo_2022}
S.~Nevo, E.~Morin, A.~Gerzi~Rosenthal, A.~Metzger, C.~Barshai, D.~Weitzner, D.~Voloshin, F.~Kratzert, G.~Elidan, G.~Dror, G.~Begelman, G.~Nearing, G.~Shalev, H.~Noga, I.~Shavitt, L.~Yuklea, M.~Royz, N.~Giladi, N.~Peled~Levi, O.~Reich, O.~Gilon, R.~Maor, S.~Timnat, T.~Shechter, V.~Anisimov, Y.~Gigi, Y.~Levin, Z.~Moshe, Z.~Ben-Haim, A.~Hassidim, and Y.~Matias.
\newblock Flood forecasting with machine learning models in an operational framework.
\newblock {\em Hydrology and Earth System Sciences}, 26(15):4013--4032, 2022.

\bibitem{najafi2024high}
Husain Najafi, Pallav~Kumar Shrestha, Oldrich Rakovec, Heiko Apel, Sergiy Vorogushyn, Rohini Kumar, Stephan Thober, Bruno Merz, and Luis Samaniego.
\newblock High-resolution impact-based early warning system for riverine flooding.
\newblock {\em Nature communications}, 15(1):3726, 2024.

\bibitem{GloFAS}
L.~Alfieri, P.~Burek, E.~Dutra, B.~Krzeminski, D.~Muraro, J.~Thielen, and F.~Pappenberger.
\newblock Glofas - global ensemble streamflow forecasting and flood early warning.
\newblock {\em Hydrology and Earth System Sciences}, 17(3):1161--1175, 2013.

\bibitem{GloFAS_Forecast}
S.~Harrigan, E.~Zsoter, H.~Cloke, P.~Salamon, and C.~Prudhomme.
\newblock Daily ensemble river discharge reforecasts and real-time forecasts from the operational global flood awareness system.
\newblock {\em Hydrology and Earth System Sciences}, 27(1):1--19, 2023.

\bibitem{jones2023ai}
Anne Jones, Julian Kuehnert, Paolo Fraccaro, Oph{\'e}lie Meuriot, Tatsuya Ishikawa, Blair Edwards, Nikola Stoyanov, Sekou~L Remy, Kommy Weldemariam, and Solomon Assefa.
\newblock Ai for climate impacts: applications in flood risk.
\newblock {\em npj Climate and Atmospheric Science}, 6(1):63, 2023.

\bibitem{reichstein2025early}
Markus Reichstein, Vitus Benson, Jan Blunk, Gustau Camps-Valls, Felix Creutzig, Carina~J Fearnley, Boran Han, Kai Kornhuber, Nasim Rahaman, Bernhard Sch{\"o}lkopf, et~al.
\newblock Early warning of complex climate risk with integrated artificial intelligence.
\newblock {\em Nature Communications}, 16(1):2564, 2025.

\bibitem{Nearing_2021}
Grey~S. Nearing, Frederik Kratzert, Alden~Keefe Sampson, Craig~S. Pelissier, Daniel Klotz, Jonathan~M. Frame, Cristina Prieto, and Hoshin~V. Gupta.
\newblock What role does hydrological science play in the age of machine learning?
\newblock {\em Water Resources Research}, 57(3):e2020WR028091, 2021.
\newblock e2020WR028091 10.1029/2020WR028091.

\bibitem{Frame_2022}
J.~M. Frame, F.~Kratzert, D.~Klotz, M.~Gauch, G.~Shalev, O.~Gilon, L.~M. Qualls, H.~V. Gupta, and G.~S. Nearing.
\newblock Deep learning rainfall--runoff predictions of extreme events.
\newblock {\em Hydrology and Earth System Sciences}, 26(13):3377--3392, 2022.

\bibitem{pathak2022fourcastnet}
Jaideep Pathak, Shashank Subramanian, Peter Harrington, Sanjeev Raja, Ashesh Chattopadhyay, Morteza Mardani, Thorsten Kurth, David Hall, Zongyi Li, Kamyar Azizzadenesheli, et~al.
\newblock Fourcastnet: A global data-driven high-resolution weather model using adaptive fourier neural operators.
\newblock {\em arXiv preprint arXiv:2202.11214}, 2022.

\bibitem{GenCast}
Ilan Price, Alvaro Sanchez-Gonzalez, Ferran Alet, Tom~R Andersson, Andrew El-Kadi, Dominic Masters, Timo Ewalds, Jacklynn Stott, Shakir Mohamed, Peter Battaglia, et~al.
\newblock Probabilistic weather forecasting with machine learning.
\newblock {\em Nature}, 637(8044):84--90, 2025.

\bibitem{PanguWeather}
Kaifeng Bi, Lingxi Xie, Hengheng Zhang, Xin Chen, Xiaotao Gu, and Qi~Tian.
\newblock Accurate medium-range global weather forecasting with 3d neural networks.
\newblock {\em Nature}, 619(7970):533--538, 2023.

\bibitem{Google_nature}
Grey Nearing, Deborah Cohen, Vusumuzi Dube, Martin Gauch, Oren Gilon, Shaun Harrigan, Avinatan Hassidim, Daniel Klotz, Frederik Kratzert, Asher Metzger, et~al.
\newblock Global prediction of extreme floods in ungauged watersheds.
\newblock {\em Nature}, 627(8004):559--563, 2024.

\bibitem{Linear_SSM}
Albert Gu, Isys Johnson, Karan Goel, Khaled Saab, Tri Dao, Atri Rudra, and Christopher R\'{e}.
\newblock Combining recurrent, convolutional, and continuous-time models with linear state space layers.
\newblock In {\em Advances in Neural Information Processing Systems}, volume~34, pages 572--585. Curran Associates, Inc., 2021.

\bibitem{S4}
Albert Gu, Karan Goel, and Christopher R\'e.
\newblock Efficiently modeling long sequences with structured state spaces.
\newblock In {\em The International Conference on Learning Representations ({ICLR})}, 2022.

\bibitem{S5}
Jimmy~T.H. Smith, Andrew Warrington, and Scott~W Linderman.
\newblock Simplified state space layers for sequence modeling.
\newblock In {\em The Eleventh International Conference on Learning Representations}, 2023.

\bibitem{mamba}
Albert Gu and Tri Dao.
\newblock Mamba: Linear-time sequence modeling with selective state spaces.
\newblock In {\em First Conference on Language Modeling}, 2024.

\bibitem{Bomers_2023}
Anouk Bomers and Suzanne J. M.~H. Hulscher.
\newblock Neural networks for fast fluvial flood predictions: Too good to be true?
\newblock {\em River Research and Applications}, 39(8):1652--1658, 2023.

\bibitem{nhess-25-747-2025}
J.~Green, I.~D. Haigh, N.~Quinn, J.~Neal, T.~Wahl, M.~Wood, D.~Eilander, M.~de~Ruiter, P.~Ward, and P.~Camus.
\newblock Review article: A comprehensive review of compound flooding literature with a focus on coastal and estuarine regions.
\newblock {\em Natural Hazards and Earth System Sciences}, 25(2):747--816, 2025.

\bibitem{hess-28-5443-2024}
T.~Cache, M.~S. Gomez, T.~Beucler, J.~Blagojevic, J.~P. Leitao, and N.~Peleg.
\newblock Enhancing generalizability of data-driven urban flood models by incorporating contextual information.
\newblock {\em Hydrology and Earth System Sciences}, 28(24):5443--5458, 2024.

\bibitem{w13162255}
Julian Hofmann and Holger Schüttrumpf.
\newblock floodgan: Using deep adversarial learning to predict pluvial flooding in real time.
\newblock {\em Water}, 13(16), 2021.

\bibitem{busker2025value}
Tim Busker, Bart van~den Hurk, Hans de~Moel, and Jeroen~CJH Aerts.
\newblock The value of precipitation forecasts to anticipate floods.
\newblock {\em Bulletin of the American Meteorological Society}, 2025.

\bibitem{mosavi2018flood}
Amir Mosavi, Pinar Ozturk, and Kwok-wing Chau.
\newblock Flood prediction using machine learning models: Literature review.
\newblock {\em Water}, 10(11):1536, 2018.

\bibitem{Bentivoglio_2022}
R.~Bentivoglio, E.~Isufi, S.~N. Jonkman, and R.~Taormina.
\newblock Deep learning methods for flood mapping: a review of existing applications and future research directions.
\newblock {\em Hydrology and Earth System Sciences}, 26(16):4345--4378, 2022.

\bibitem{Gau_2021}
Zifeng Guo, João~P. Leitão, Nuno~E. Simões, and Vahid Moosavi.
\newblock Data-driven flood emulation: Speeding up urban flood predictions by deep convolutional neural networks.
\newblock {\em Journal of Flood Risk Management}, 14(1):e12684, 2021.

\bibitem{10351020}
Alexander~Y. Sun, Zhi Li, Wonhyun Lee, Qixing Huang, Bridget~R. Scanlon, and Clint Dawson.
\newblock Rapid flood inundation forecast using fourier neural operator.
\newblock In {\em IEEE/CVF International Conference on Computer Vision Workshops (ICCVW)}, pages 3735--3741, 2023.

\bibitem{Seleem31122022}
Omar Seleem, Georgy Ayzel, Arthur Costa~Tomaz de~Souza, Axel Bronstert, and Maik~Heistermann and.
\newblock Towards urban flood susceptibility mapping using data-driven models in berlin, germany.
\newblock {\em Geomatics, Natural Hazards and Risk}, 13(1):1640--1662, 2022.

\bibitem{w15091760}
Benjamin Burrichter, Julian Hofmann, Juliana Koltermann~da Silva, Andre Niemann, and Markus Quirmbach.
\newblock A spatiotemporal deep learning approach for urban pluvial flood forecasting with multi-source data.
\newblock {\em Water}, 15(9), 2023.

\bibitem{CHU2020104587}
Haibo Chu, Wenyan Wu, Q.J. Wang, Rory Nathan, and Jiahua Wei.
\newblock An ann-based emulation modelling framework for flood inundation modelling: Application, challenges and future directions.
\newblock {\em Environmental Modelling \& Software}, 124:104587, 2020.

\bibitem{w15030566}
Fazlul Karim, Mohammed~Ali Armin, David Ahmedt-Aristizabal, Lachlan Tychsen-Smith, and Lars Petersson.
\newblock A review of hydrodynamic and machine learning approaches for flood inundation modeling.
\newblock {\em Water}, 15(3), 2023.

\bibitem{Floodsformer_2025}
Matteo Pianforini, Susanna Dazzi, Andrea Pilzer, and Renato Vacondio.
\newblock Floodsformer: A transformer-based data-driven model for predicting the 2-d dynamics of fluvial floods.
\newblock {\em Environmental Modelling \& Software}, 193:106599, 2025.

\bibitem{10758300}
Björn Lütjens, Brandon Leshchinskiy, Océane Boulais, Farrukh Chishtie, Natalia Díaz-Rodríguez, Margaux Masson-Forsythe, Ana Mata-Payerro, Christian Requena-Mesa, Aruna Sankaranarayanan, Aaron Piña, Yarin Gal, Chedy Raïssi, Alexander Lavin, and Dava Newman.
\newblock Generating physically-consistent satellite imagery for climate visualizations.
\newblock {\em IEEE Transactions on Geoscience and Remote Sensing}, 62:1--11, 2024.

\bibitem{JANIZADEH2021113551}
Saeid Janizadeh, Subodh {Chandra Pal}, Asish Saha, Indrajit Chowdhuri, Kourosh Ahmadi, Sajjad Mirzaei, Amir~Hossein Mosavi, and John~P. Tiefenbacher.
\newblock Mapping the spatial and temporal variability of flood hazard affected by climate and land-use changes in the future.
\newblock {\em Journal of Environmental Management}, 298:113551, 2021.

\bibitem{Dasgupta_2021}
Antara Dasgupta, Renaud Hostache, RAAJ Ramsankaran, Guy J.-P. Schumann, Stefania Grimaldi, Valentijn R.~N. Pauwels, and Jeffrey~P. Walker.
\newblock A mutual information-based likelihood function for particle filter flood extent assimilation.
\newblock {\em Water Resources Research}, 57(2):e2020WR027859, 2021.
\newblock e2020WR027859 2020WR027859.

\bibitem{NEURIPS2024_43612b06}
Nikolaos~Ioannis Bountos, Maria Sdraka, Angelos Zavras, Ilektra Karasante, Andreas Karavias, Themistocles Herekakis, Angeliki Thanasou, Dimitrios Michail, and Ioannis Papoutsis.
\newblock Kuro siwo: 33 billion m\^{}2 under the water. a global multi-temporal satellite dataset for rapid flood mapping.
\newblock In {\em Advances in Neural Information Processing Systems}, volume~37, pages 38105--38121. Curran Associates, Inc., 2024.

\bibitem{palash2024data}
Wahid Palash, Ali~S Akanda, and Shafiqul Islam.
\newblock A data-driven global flood forecasting system for medium to large rivers.
\newblock {\em Scientific Reports}, 14(1):8979, 2024.

\bibitem{ZHONG2024132165}
Liangjin Zhong, Huimin Lei, Zhiyuan Li, and Shijie Jiang.
\newblock Advancing streamflow prediction in data-scarce regions through vegetation-constrained distributed hybrid ecohydrological models.
\newblock {\em Journal of Hydrology}, 645:132165, 2024.

\bibitem{hess-28-4187-2024}
F.~Kratzert, M.~Gauch, D.~Klotz, and G.~Nearing.
\newblock Hess opinions: Never train a long short-term memory (lstm) network on a single basin.
\newblock {\em Hydrology and Earth System Sciences}, 28(17):4187--4201, 2024.

\bibitem{LSTM}
Sepp Hochreiter and Jürgen Schmidhuber.
\newblock Long short-term memory.
\newblock {\em Neural Computation}, 9(8):1735--1780, 1997.

\bibitem{EA-LSTM}
F.~Kratzert, D.~Klotz, G.~Shalev, G.~Klambauer, S.~Hochreiter, and G.~Nearing.
\newblock Towards learning universal, regional, and local hydrological behaviors via machine learning applied to large-sample datasets.
\newblock {\em Hydrology and Earth System Sciences}, 23(12):5089--5110, 2019.

\bibitem{Heudorder_2025}
Benedikt Heudorfer, Hoshin~V. Gupta, and Ralf Loritz.
\newblock Are deep learning models in hydrology entity aware?
\newblock {\em Geophysical Research Letters}, 52(6):e2024GL113036, 2025.
\newblock e2024GL113036 2024GL113036.

\bibitem{ZHANG2022128577}
Yikui Zhang, Silvan Ragettli, Peter Molnar, Olga Fink, and Nadav Peleg.
\newblock Generalization of an encoder-decoder lstm model for flood prediction in ungauged catchments.
\newblock {\em Journal of Hydrology}, 614:128577, 2022.

\bibitem{ZHANG2024100617}
Binlan Zhang, Chaojun Ouyang, Peng Cui, Qingsong Xu, Dongpo Wang, Fei Zhang, Zhong Li, Linfeng Fan, Marco Lovati, Yanling Liu, and Qianqian Zhang.
\newblock Deep learning for cross-region streamflow and flood forecasting at a global scale.
\newblock {\em The Innovation}, 5(3):100617, 2024.

\bibitem{ruparell2024hydra}
Karan Ruparell, Robert~J. Marks, Andy Wood, Kieran M.~R. Hunt, Hannah~L. Cloke, Christel Prudhomme, Florian Pappenberger, and Matthew Chantry.
\newblock Hydra-lstm: A semi-shared machine learning architecture for prediction across watersheds.
\newblock {\em Artificial Intelligence for the Earth Systems}, 4(3):240103, 2025.

\bibitem{MC-LSTM}
Yihan Wang, Lujun Zhang, N.~Benjamin Erichson, and Tiantian Yang.
\newblock Investigating the streamflow simulation capability of a new mass-conserving long short-term memory (mc-lstm) model across the contiguous united states.
\newblock {\em Journal of Hydrology}, 658:133161, 2025.

\bibitem{MF-LSTM}
E.~Acu\~na Espinoza, F.~Kratzert, D.~Klotz, M.~Gauch, M.~\'Alvarez~Chaves, R.~Loritz, and U.~Ehret.
\newblock Technical note: An approach for handling multiple temporal frequencies with different input dimensions using a single lstm cell.
\newblock {\em Hydrology and Earth System Sciences}, 29(6):1749--1758, 2025.

\bibitem{DRUM_2025}
Zhigang Ou, Congyi Nai, Baoxiang Pan, Yi~Zheng, Chaopeng Shen, Peishi Jiang, Xingcai Liu, Qiuhong Tang, Wenqing Li, and Ming Pan.
\newblock Probabilistic diffusion models advance extreme flood forecasting.
\newblock {\em Geophysical Research Letters}, 52(15):e2025GL115705, 2025.
\newblock e2025GL115705 2025GL115705.

\bibitem{stein2025causalrivers}
Gideon Stein, Maha Shadaydeh, Jan Blunk, Niklas Penzel, and Joachim Denzler.
\newblock Causalrivers--scaling up benchmarking of causal discovery for real-world time-series.
\newblock In {\em The Thirteenth International Conference on Learning Representations}, 2025.

\bibitem{kirschstein2024merit}
Nikolas Kirschstein and Yixuan Sun.
\newblock The merit of river network topology for neural flood forecasting.
\newblock In {\em International Conference on Machine Learning}, pages 24713--24725. PMLR, 2024.

\bibitem{roudbari2024data}
Naghmeh~Shafiee Roudbari, Shubham~Rajeev Punekar, Zachary Patterson, Ursula Eicker, and Charalambos Poullis.
\newblock From data to action in flood forecasting leveraging graph neural networks and digital twin visualization.
\newblock {\em Scientific reports}, 14(1):18571, 2024.

\bibitem{FloodGNN}
Arnold Kazadi, James Doss-Gollin, Antonia Sebastian, and Arlei Silva.
\newblock Floodgnn-gru: a spatio-temporal graph neural network for flood prediction.
\newblock {\em Environmental Data Science}, 3:e21, 2024.

\bibitem{yang2025global}
Yuan Yang, Dapeng Feng, Hylke~E. Beck, Weiming Hu, Ather Abbas, Agniv Sengupta, Luca Delle~Monache, Robert Hartman, Peirong Lin, Chaopeng Shen, and Ming Pan.
\newblock Global daily discharge estimation based on grid long short-term memory (lstm) model and river routing.
\newblock {\em Water Resources Research}, 61(6):e2024WR039764, 2025.
\newblock e2024WR039764 2024WR039764.

\bibitem{Vischer_2025}
M.~A. Vischer, N.~Otero, and J.~Ma.
\newblock Spatially resolved rainfall streamflow modeling in central europe.
\newblock {\em EGUsphere}, 2025:1--26, 2025.

\bibitem{Bindas_2024}
Tadd Bindas, Wen-Ping Tsai, Jiangtao Liu, Farshid Rahmani, Dapeng Feng, Yuchen Bian, Kathryn Lawson, and Chaopeng Shen.
\newblock Improving river routing using a differentiable muskingum-cunge model and physics-informed machine learning.
\newblock {\em Water Resources Research}, 60(1):e2023WR035337, 2024.
\newblock e2023WR035337 2023WR035337.

\bibitem{Song_2025}
Yalan Song, Tadd Bindas, Chaopeng Shen, Haoyu Ji, Wouter J.~M. Knoben, Leo Lonzarich, Martyn~P. Clark, Jiangtao Liu, Katie van Werkhoven, Sam Lamont, Matthew Denno, Ming Pan, Yuan Yang, Jeremy Rapp, Mukesh Kumar, Farshid Rahmani, Cyril Thébault, Richard Adkins, James Halgren, Trupesh Patel, Arpita Patel, Kamlesh~Arun Sawadekar, and Kathryn Lawson.
\newblock High-resolution national-scale water modeling is enhanced by multiscale differentiable physics-informed machine learning.
\newblock {\em Water Resources Research}, 61(4):e2024WR038928, 2025.
\newblock e2024WR038928 2024WR038928.

\bibitem{Wang_2024}
Chao Wang, Shijie Jiang, Yi~Zheng, Feng Han, Rohini Kumar, Oldrich Rakovec, and Siqi Li.
\newblock Distributed hydrological modeling with physics-encoded deep learning: A general framework and its application in the amazon.
\newblock {\em Water Resources Research}, 60(4):e2023WR036170, 2024.
\newblock e2023WR036170 2023WR036170.

\bibitem{vmamba}
Yue Liu, Yunjie Tian, Yuzhong Zhao, Hongtian Yu, Lingxi Xie, Yaowei Wang, Qixiang Ye, Jianbin Jiao, and Yunfan Liu.
\newblock Vmamba: Visual state space model.
\newblock In {\em Advances in Neural Information Processing Systems}, volume~37, pages 103031--103063. Curran Associates, Inc., 2024.

\bibitem{vimmamba}
Lianghui Zhu, Bencheng Liao, Qian Zhang, Xinlong Wang, Wenyu Liu, and Xinggang Wang.
\newblock Vision mamba: Efficient visual representation learning with bidirectional state space model.
\newblock In {\em Forty-first International Conference on Machine Learning}, 2024.

\bibitem{ViT}
Alexey Dosovitskiy, Lucas Beyer, Alexander Kolesnikov, Dirk Weissenborn, Xiaohua Zhai, Thomas Unterthiner, Mostafa Dehghani, Matthias Minderer, Georg Heigold, Sylvain Gelly, et~al.
\newblock An image is worth 16x16 words: Transformers for image recognition at scale.
\newblock {\em arXiv preprint arXiv:2010.11929}, 2020.

\bibitem{DiMSUM}
Hao Phung, Quan Dao, Trung Dao, Hoang Phan, Dimitris~N. Metaxas, and Anh Tran.
\newblock Dimsum: Diffusion mamba - a scalable and unified spatial-frequency method for image generation.
\newblock In {\em Advances in Neural Information Processing Systems}, volume~37, pages 32947--32979. Curran Associates, Inc., 2024.

\bibitem{zigmamba}
Vincent~Tao Hu, Stefan~Andreas Baumann, Ming Gui, Olga Grebenkova, Pingchuan Ma, Johannes Fischer, and Bj{\"o}rn Ommer.
\newblock Zigma: A dit-style zigzag mamba diffusion model.
\newblock In {\em European Conference on Computer Vision (ECCV)}, pages 148--166, Cham, 2025. Springer Nature Switzerland.

\bibitem{NEURIPS2024_89b89c04}
Yicheng Xiao, Lin Song, Shaoli Huang, Jiangshan Wang, Siyu Song, Yixiao Ge, Xiu Li, and Ying Shan.
\newblock Mambatree: Tree topology is all you need in state space model.
\newblock In {\em Advances in Neural Information Processing Systems}, volume~37, pages 75329--75354. Curran Associates, Inc., 2024.

\bibitem{GroupMamba}
Abdelrahman Shaker, Syed~Talal Wasim, Salman Khan, Juergen Gall, and Fahad~Shahbaz Khan.
\newblock Groupmamba: Efficient group-based visual state space model.
\newblock In {\em Proceedings of the IEEE/CVF Conference on Computer Vision and Pattern Recognition (CVPR)}, pages 14912--14922, June 2025.

\bibitem{videomamba}
Kunchang Li, Xinhao Li, Yi~Wang, Yinan He, Yali Wang, Limin Wang, and Yu~Qiao.
\newblock Videomamba: State space model for efficient video understanding.
\newblock In {\em European Conference on Computer Vision (ECCV)}, pages 237--255, Cham, 2025. Springer Nature Switzerland.

\bibitem{chen2024video}
Guo Chen, Yifei Huang, Jilan Xu, Baoqi Pei, Zhe Chen, Zhiqi Li, Jiahao Wang, Kunchang Li, Tong Lu, and Limin Wang.
\newblock Video mamba suite: State space model as a versatile alternative for video understanding.
\newblock {\em arXiv preprint arXiv:2403.09626}, 2024.

\bibitem{motionmamba}
Zeyu Zhang, Akide Liu, Ian Reid, Richard Hartley, Bohan Zhuang, and Hao Tang.
\newblock Motion mamba: Efficient and long sequence motion generation.
\newblock In {\em European Conference on Computer Vision (ECCV)}, pages 265--282, Cham, 2025. Springer Nature Switzerland.

\bibitem{zatsarynna2025manta}
Olga Zatsarynna, Emad Bahrami, Yazan~Abu Farha, Gianpiero Francesca, and Juergen Gall.
\newblock Manta: Diffusion mamba for efficient and effective stochastic long-term dense action anticipation.
\newblock In {\em IEEE Conference on Computer Vision and Pattern Recognition (CVPR)}, 2025.

\bibitem{pointmamba}
Dingkang Liang, Xin Zhou, Wei Xu, Xingkui Zhu, Zhikang Zou, Xiaoqing Ye, Xiao Tan, and Xiang Bai.
\newblock Pointmamba: A simple state space model for point cloud analysis.
\newblock In {\em Advances in Neural Information Processing Systems}, volume~37, pages 32653--32677. Curran Associates, Inc., 2024.

\bibitem{NEURIPS2024_947b6383}
Guowen Zhang, Lue Fan, Chenhang He, Zhen Lei, Zhaoxiang Zhang, and Lei Zhang.
\newblock Voxel mamba: Group-free state space models for point cloud based 3d object detection.
\newblock In {\em Advances in Neural Information Processing Systems}, volume~37, pages 81489--81509. Curran Associates, Inc., 2024.

\bibitem{ERA5-Land}
J.~Mu\~noz Sabater, E.~Dutra, A.~Agust{\'i}-Panareda, C.~Albergel, G.~Arduini, G.~Balsamo, S.~Boussetta, M.~Choulga, S.~Harrigan, H.~Hersbach, B.~Martens, D.~G. Miralles, M.~Piles, N.~J. Rodr{\'i}guez-Fern\'andez, E.~Zsoter, C.~Buontempo, and J.-N. Th\'epaut.
\newblock Era5-land: a state-of-the-art global reanalysis dataset for land applications.
\newblock {\em Earth System Science Data}, 13(9):4349--4383, 2021.

\bibitem{GloFAS_Reanalysis}
S.~Harrigan, E.~Zsoter, L.~Alfieri, C.~Prudhomme, P.~Salamon, F.~Wetterhall, C.~Barnard, H.~Cloke, and F.~Pappenberger.
\newblock Glofas-era5 operational global river discharge reanalysis 1979--present.
\newblock {\em Earth System Science Data}, 12(3):2043--2060, 2020.

\bibitem{xie2010cpc}
Pingping Xie, M~Chen, and W~Shi.
\newblock Cpc unified gauge-based analysis of global daily precipitation.
\newblock In {\em Preprints, 24th Conf. on Hydrology, Atlanta, GA, Amer. Meteor. Soc}, volume~2, 2010.

\bibitem{xie2007gauge}
Pingping Xie, Mingyue Chen, Song Yang, Akiyo Yatagai, Tadahiro Hayasaka, Yoshihiro Fukushima, and Changming Liu.
\newblock A gauge-based analysis of daily precipitation over east asia.
\newblock {\em Journal of Hydrometeorology}, 8(3):607--626, 2007.

\bibitem{chen2008assessing}
Mingyue Chen, Wei Shi, Pingping Xie, Viviane~BS Silva, Vernon~E Kousky, R~Wayne~Higgins, and John~E Janowiak.
\newblock Assessing objective techniques for gauge-based analyses of global daily precipitation.
\newblock {\em Journal of Geophysical Research: Atmospheres}, 113(D4), 2008.

\bibitem{LISFLOOD_Static}
M.~Choulga, F.~Moschini, C.~Mazzetti, S.~Grimaldi, J.~Disperati, H.~Beck, P.~Salamon, and C.~Prudhomme.
\newblock Technical note: Surface fields for global environmental modelling.
\newblock {\em Hydrology and Earth System Sciences}, 28(13):2991--3036, 2024.

\bibitem{dao2022flashattention}
Tri Dao, Dan Fu, Stefano Ermon, Atri Rudra, and Christopher R{\'e}.
\newblock Flashattention: Fast and memory-efficient exact attention with io-awareness.
\newblock {\em Advances in neural information processing systems}, 35:16344--16359, 2022.

\bibitem{dao2023flashattention2}
Tri Dao.
\newblock Flash{A}ttention-2: Faster attention with better parallelism and work partitioning.
\newblock In {\em International Conference on Learning Representations (ICLR)}, 2024.

\bibitem{peano1990courbe}
Giuseppe Peano and G~Peano.
\newblock {\em Sur une courbe, qui remplit toute une aire plane}.
\newblock Springer, 1990.

\bibitem{hilbert1935stetige}
David Hilbert and David Hilbert.
\newblock {\"U}ber die stetige abbildung einer linie auf ein fl{\"a}chenst{\"u}ck.
\newblock {\em Dritter Band: Analysis{\textperiodcentered} Grundlagen der Mathematik{\textperiodcentered} Physik Verschiedenes: Nebst Einer Lebensgeschichte}, pages 1--2, 1935.

\bibitem{LOAN}
Mohamad~Hakam Shams~Eddin, Ribana Roscher, and Juergen Gall.
\newblock Location-aware adaptive normalization: A deep learning approach for wildfire danger forecasting.
\newblock {\em IEEE Transactions on Geoscience and Remote Sensing}, 61:1--18, 2023.

\bibitem{Transformer}
Ashish Vaswani, Noam Shazeer, Niki Parmar, Jakob Uszkoreit, Llion Jones, Aidan~N Gomez, \L~ukasz Kaiser, and Illia Polosukhin.
\newblock Attention is all you need.
\newblock In {\em Advances in Neural Information Processing Systems}, volume~30. Curran Associates, Inc., 2017.

\bibitem{LISFLOOD}
J.~M. Van~Der Knijff, J.~Younis, and A.~P. J. De~Roo and.
\newblock Lisflood: a gis‐based distributed model for river basin scale water balance and flood simulation.
\newblock {\em International Journal of Geographical Information Science}, 24(2):189--212, 2010.

\bibitem{ERA5}
Hans Hersbach, Bill Bell, Paul Berrisford, Shoji Hirahara, András Horányi, Joaquín Muñoz-Sabater, Julien Nicolas, Carole Peubey, Raluca Radu, Dinand Schepers, Adrian Simmons, Cornel Soci, Saleh Abdalla, Xavier Abellan, Gianpaolo Balsamo, Peter Bechtold, Gionata Biavati, Jean Bidlot, Massimo Bonavita, Giovanna De~Chiara, Per Dahlgren, Dick Dee, Michail Diamantakis, Rossana Dragani, Johannes Flemming, Richard Forbes, Manuel Fuentes, Alan Geer, Leo Haimberger, Sean Healy, Robin~J. Hogan, Elías Hólm, Marta Janisková, Sarah Keeley, Patrick Laloyaux, Philippe Lopez, Cristina Lupu, Gabor Radnoti, Patricia de~Rosnay, Iryna Rozum, Freja Vamborg, Sebastien Villaume, and Jean-Noël Th{\'e}paut.
\newblock The era5 global reanalysis.
\newblock {\em Quarterly Journal of the Royal Meteorological Society}, 146(730):1999--2049, 2020.

\bibitem{ZAJAC2017}
Zuzanna Zajac, Beatriz Revilla-Romero, Peter Salamon, Peter Burek, Feyera~A. Hirpa, and Hylke Beck.
\newblock The impact of lake and reservoir parameterization on global streamflow simulation.
\newblock {\em Journal of Hydrology}, 548:552--568, 2017.

\bibitem{HTESSEL}
Gianpaolo Balsamo, Anton Beljaars, Klaus Scipal, Pedro Viterbo, Bart van~den Hurk, Martin Hirschi, and Alan~K. Betts.
\newblock A revised hydrology for the ecmwf model: Verification from field site to terrestrial water storage and impact in the integrated forecast system.
\newblock {\em Journal of Hydrometeorology}, 10(3):623 -- 643, 2009.

\bibitem{xESMF}
Jiawei Zhuang, raphael dussin, André Jüling, and Stephan Rasp.
\newblock {JiaweiZhuang/xESMF: v0.3.0 Adding ESMF.LocStream capabilities}, March 2020.

\bibitem{HydroRIVERS}
Bernhard Lehner and Günther Grill.
\newblock Global river hydrography and network routing: baseline data and new approaches to study the world's large river systems.
\newblock {\em Hydrological Processes}, 27(15):2171--2186, 2013.

\bibitem{HydroATLAS}
Simon Linke, Bernhard Lehner, Camille Ouellet~Dallaire, Joseph Ariwi, G{\"u}nther Grill, Mira Anand, Penny Beames, Vicente Burchard-Levine, Sally Maxwell, Hana Moidu, et~al.
\newblock Global hydro-environmental sub-basin and river reach characteristics at high spatial resolution.
\newblock {\em Scientific data}, 6(1):283, 2019.

\bibitem{Gauch_2023}
Martin Gauch, Frederik Kratzert, Oren Gilon, Hoshin Gupta, Juliane Mai, Grey Nearing, Bryan Tolson, Sepp Hochreiter, and Daniel Klotz.
\newblock In defense of metrics: Metrics sufficiently encode typical human preferences regarding hydrological model performance.
\newblock {\em Water Resources Research}, 59(6):e2022WR033918, 2023.
\newblock e2022WR033918 2022WR033918.

\bibitem{mamba2}
Tri Dao and Albert Gu.
\newblock Transformers are {SSM}s: Generalized models and efficient algorithms through structured state space duality.
\newblock In {\em International Conference on Machine Learning (ICML)}, 2024.

\bibitem{Wu_2024_CVPR}
Xiaoyang Wu, Li~Jiang, Peng-Shuai Wang, Zhijian Liu, Xihui Liu, Yu~Qiao, Wanli Ouyang, Tong He, and Hengshuang Zhao.
\newblock Point transformer v3: Simpler faster stronger.
\newblock In {\em Proceedings of the IEEE/CVF Conference on Computer Vision and Pattern Recognition (CVPR)}, pages 4840--4851, June 2024.

\bibitem{GELU}
Dan Hendrycks and Kevin Gimpel.
\newblock Gaussian error linear units (gelus).
\newblock {\em arXiv preprint arXiv:1606.08415}, 2016.

\bibitem{pseudo-hilbert}
Jian Zhang, Sei-ichiro Kamata, and Yoshifumi Ueshige.
\newblock A pseudo-hilbert scan algorithm for arbitrarily-sized rectangle region.
\newblock In {\em Advances in Machine Vision, Image Processing, and Pattern Analysis}, pages 290--299, Berlin, Heidelberg, 2006. Springer Berlin Heidelberg.

\bibitem{grey_nearing_2023_10397664}
Grey Nearing.
\newblock Global prediction of extreme floods in ungauged watersheds, December 2023.

\bibitem{Magdalena_2025}
Magdalena Kracheletz, Ziyu Liu, Anne Springer, Jürgen Kusche, and Petra Friederichs.
\newblock Would the 2021 western europe flood event be visible in satellite gravimetry?
\newblock {\em Journal of Geophysical Research: Atmospheres}, 130(3):e2024JD042190, 2025.
\newblock e2024JD042190 2024JD042190.

\bibitem{Austria_flood_2024}
C~Hauer, M~Paster, U~Pulg, T~Ofenb{\"o}ck, and H~Habersack.
\newblock Critical flows at the wien river during the 1000-years event in september 2024--causes, consequences and possible management options for urban river flood management.
\newblock {\em Natural Hazards}, pages 1--13, 2025.

\bibitem{RiverMamba_dataset}
Mohamad~Hakam Shams~Eddin, Yikui Zhang, Stefan Kollet, and Juergen Gall.
\newblock {RiverMamba: A State Space Model for Global River Discharge and Flood Forecasting [data set]}, 2025.

\end{thebibliography}
}

\end{document}